\documentclass[acmlarge]{acmart}

\usepackage{multirow}
\usepackage[utf8]{inputenc} % allow utf-8 input
\usepackage{url}            % simple URL typesetting
\usepackage{booktabs}       % professional-quality tables
\usepackage{amsfonts}       % blackboard math symbols
\usepackage{nicefrac}       % compact symbols for 1/2, etc.
\usepackage{xcolor}         % colors
\usepackage{colortbl}       % table colors
\usepackage{float}
\usepackage{amsmath}
\usepackage{amsthm}
\usepackage{latexsym}
\usepackage{thmtools}
\usepackage{thm-restate}
\usepackage{etoolbox}
\usepackage{array}
\usepackage{booktabs}
\usepackage{makecell}
\usepackage{tabularx}
\usepackage{enumitem}
\usepackage{titlesec}
\usepackage{caption}
\usepackage{subcaption}
\usepackage{comment}
\usepackage{algorithm}
\usepackage{algorithmicx}
\usepackage[noend]{algpseudocode}
\usepackage{pifont}
\usepackage[framemethod=TikZ]{mdframed}
\usepackage{bbding}
\usepackage{verbatim,ifthen,alltt}
\usepackage{microtype}      % microtypography
\usepackage{stmaryrd}
\usepackage{xfrac}
\usepackage{tikz}
\usepackage{pgf}
\usepackage{forest}

\usetikzlibrary{fit,calc,tikzmark}
\usetikzlibrary{arrows,decorations.pathmorphing,decorations.pathreplacing,decorations.footprints,fadings,calc,trees,mindmap,shadows,decorations.text,patterns,positioning,shapes}
\usetikzlibrary{arrows,shadows,backgrounds}
\usetikzlibrary{arrows.meta}
\usetikzlibrary{positioning,fit,automata,matrix}
\usetikzlibrary{shapes.symbols,shapes.misc,shapes.arrows}

\definecolor{darkred}{rgb}{0.7,0.1,0.1}
\definecolor{medred}{rgb}{0.5,0.1,0.1}
\definecolor{midred}{rgb}{0.7,0.2,0.2}
\definecolor{vdarkred}{rgb}{0.4,0.1,0.1}
\definecolor{darkslategray}{rgb}{0.18, 0.31, 0.31} %#2F4F4F
\definecolor{platinum}{rgb}{0.9, 0.89, 0.89} %#E5E4E2
\definecolor{gray}{rgb}{.4,.4,.4}
\definecolor{midgrey}{rgb}{0.5,0.5,0.5}
\definecolor{middarkgrey}{rgb}{0.35,0.35,0.35}
\definecolor{darkgrey}{rgb}{0.3,0.3,0.3}
\definecolor{darkred}{rgb}{0.7,0.1,0.1}
\definecolor{midblue}{rgb}{0.2,0.2,0.7}
\definecolor{darkblue}{rgb}{0.1,0.1,0.5}
\definecolor{darkgreen}{rgb}{0.1,0.5,0.1}
\definecolor{defseagreen}{cmyk}{0.69,0,0.50,0}
\definecolor{purple3}{RGB}{125,38,205}          % purple3

\definecolor{tyellow1}{HTML}{FCE94F}
\definecolor{tyellow2}{HTML}{EDD400}
\definecolor{tyellow3}{HTML}{C4A000}
\definecolor{torange1}{HTML}{FCAF3E}
\definecolor{torange2}{HTML}{F57900}
\definecolor{torange3}{HTML}{C35C00}
\definecolor{tbrown1}{HTML}{E9B96E}
\definecolor{tbrown2}{HTML}{C17D11}
\definecolor{tbrown3}{HTML}{8F5902}
\definecolor{tgreen1}{HTML}{8AE234}
\definecolor{tgreen2}{HTML}{73D216}
\definecolor{tgreen3}{HTML}{4E9A06}
\definecolor{tblue1}{HTML}{729FCF}
\definecolor{tblue2}{HTML}{3465A4}
\definecolor{tblue3}{HTML}{204A87}
\definecolor{tpurple1}{HTML}{AD7FA8}
\definecolor{tpurple2}{HTML}{75507B}
\definecolor{tpurple3}{HTML}{5C3566}
\definecolor{tred1}{HTML}{EF2929}
\definecolor{tred2}{HTML}{CC0000}
\definecolor{tred3}{HTML}{A40000}
\definecolor{tlgray1}{HTML}{EEEEEC}
\definecolor{tlgray2}{HTML}{D3D7CF}
\definecolor{tlgray3}{HTML}{BABDB6}
\definecolor{tdgray1}{HTML}{888A85}
\definecolor{tdgray2}{HTML}{555753}
\definecolor{tdgray3}{HTML}{2E3436}

\newcommand{\hlight}[1]{{\color{darkred}#1}}
\newcommand{\rhlight}[1]{\hlight{#1}}

\newcommand{\dbhlight}[1]{{\color{darkblue}#1}}

\newcommand{\dghlight}[1]{{\color[RGB]{0,120,0}#1}}
\newcommand{\nohlight}[1]{{#1}}

\usepackage[noabbrev,nameinlink,capitalise]{cleveref}
\usepackage{xurl}
\usepackage{breakcites}
\usepackage[toc,page, title, titletoc]{appendix}

\AtBeginDocument{%
  
  \hypersetup{breaklinks=true}
}

\setcopyright{none}
\copyrightyear{2023}
\acmYear{2023}
\acmDOI{XXXXXXX.XXXXXXX}

\acmConference[arXiv preprint]{}{September 2023}{IRIT, Toulouse, France} %
\makeatletter
\def\thm@space@setup{\thm@preskip=5.0pt
\thm@postskip=0pt}
\makeatother

\newtheoremstyle{newstyle}      
{5.5pt} %Aboveskip 
{-2.0pt} %Below skip
{\mdseries} %Body font e.g.\mdseries,\bfseries,\scshape,\itshape
{} %Indent
{\bfseries} %Head font e.g.\bfseries,\scshape,\itshape
{.} %Punctuation afer theorem header
{ } %Space after theorem header
{} %Heading
\theoremstyle{newstyle}

\newtheorem{proposition}{Proposition}

\newtheorem{corollary}{Corollary}

\newtheorem{remark}{Remark}

\crefname{theorem}{Theorem}{Theorems}
\crefname{lemma}{Lemma}{Lemmas}
\crefname{proposition}{Proposition}{Propositions}
\crefname{definition}{Definition}{Definitions}
\crefname{corollary}{Corollary}{Corollaries}
\crefname{example}{Example}{Examples}
\crefname{claim}{Claim}{Claims}
\crefname{assumption}{Assumption}{Assumptions}
\crefname{enumi}{}{}



\newcommand{\nfrac}{\nicefrac}

\newcommand{\fml}[1]{{\mathcal{#1}}}

\newcommand{\tn}[1]{\textnormal{#1}}
\newcommand{\tbf}[1]{\textbf{#1}}
\newcommand{\mbf}[1]{\ensuremath\mathbf{#1}}
\newcommand{\msf}[1]{\ensuremath\mathsf{#1}}
\newcommand{\mbb}[1]{\ensuremath\mathbb{#1}}

\newcommand{\waxp}{\ensuremath\mathsf{WAXp}}
\newcommand{\wcxp}{\ensuremath\mathsf{WCXp}}
\newcommand{\axp}{\ensuremath\mathsf{AXp}}
\newcommand{\cxp}{\ensuremath\mathsf{CXp}}

\newcommand{\sv}{\ensuremath\msf{Sv}}

\newcommand{\rows}{\ensuremath\msf{rows}}

\newcommand{\oper}[1]{\ensuremath\mathsf{#1}}

\newcommand{\bigland}{\ensuremath\bigwedge}

\newcommand{\relevant}{\oper{Relevant}}
\newcommand{\irrelevant}{\oper{Irrelevant}}

\newcommand{\yesmark}{{\small\Checkmark}}
\newcommand{\nomark}{{\small\XSolidBrush}}

\newcounter{tableeqn}[table]

\DeclareMathOperator*{\limply}{\rightarrow}

\newcommand{\jnoteF}[1]{}

\newcolumntype{L}[1]{>{\raggedright\let\newline\\\arraybackslash\hspace{0pt}}m{#1}}
\newcolumntype{C}[1]{>{\centering\let\newline\\\arraybackslash\hspace{0pt}}m{#1}}
\newcolumntype{R}[1]{>{\raggedleft\let\newline\\\arraybackslash\hspace{0pt}}m{#1}}

\makeatletter
\def\xscriptsize{\@setfontsize\scriptsize{8pt}{9pt}}
\def\xtiny{\@setfontsize\tiny{7pt}{8pt}}
\def\xlarge{\@setfontsize\large{11pt}{12pt}}
\def\xLarge{\@setfontsize\Large{12pt}{14pt}}
\def\xLARGE{\@setfontsize\LARGE{13pt}{15pt}}
\def\xhuge{\@setfontsize\huge{20pt}{20pt}}
\def\xHuge{\@setfontsize\Huge{28pt}{28pt}}
\makeatother

\tikzset{
  0 my edge/.style={densely dashed, my edge, draw=midblue},
  my edge/.style={-{Stealth[]}, draw=midblue},
}

\def\HiLi{\leavevmode\rlap{\hbox to \linewidth{\color{platinum}\leaders\hrule height .8\baselineskip depth .5ex\hfill}}}

\titlespacing{\section}{0pt}{*2.15}{*1.0}
\titlespacing{\subsection}{0pt}{*1.25}{*0.75}
\titlespacing{\subsubsection}{0pt}{*0.35}{*0.5}
\titlespacing{\paragraph}{0pt}{*0.1}{*0.575}
\titleformat{\paragraph}[runin]
{\bfseries}{\theparagraph}{1em}{}

\makeatletter
\newcommand\nparagraph{%
  \@startsection{paragraph}
    {4}
    {\z@}
    {0.225ex \@plus0.225ex \@minus.125ex}
    {-1em}
    {\normalfont\normalsize\bfseries}%
}
\makeatother

\setitemize{noitemsep,topsep=0pt,parsep=0pt,partopsep=0pt}
\setenumerate{noitemsep,topsep=0pt,parsep=0pt,partopsep=0pt}
\setlist{nosep,leftmargin=0.45cm}

\providetoggle{long}
\settoggle{long}{false}

\makeatletter
\algnewcommand{\LineComment}[1]{\Statex \hskip\ALG@thistlm \(\triangleright\) #1}
\makeatother

\newboolean{addval}
\setboolean{addval}{false}

\newcommand{\condaddval}[1]{
  \ifthenelse{\boolean{addval}}{#1}{}
}

\newboolean{addapp}
\setboolean{addapp}{false}

\newcommand{\condaddapp}[1]{
  \ifthenelse{\boolean{addapp}}{#1}{}
}

\hypersetup{breaklinks=true}

\begin{document}

\let\oldaddcontentsline\addcontentsline% Store \addcontentsline
\renewcommand{\addcontentsline}[3]{}% Make \addcontentsline a no-op
%\title{Additional Evidence on Refuting Shapley Values for XAI}
\title{Refutation of Shapley Values for XAI -- Additional Evidence}

\author{Xuanxiang Huang}
\authornotemark[1]
\email{xuanxiang.huang@univ-toulouse.fr}
\orcid{0000-0002-3722-7191}
\affiliation{%
  \institution{University of Toulouse}
  %\streetaddress{No Street}
  \city{Toulouse}
  %\state{No State}
  \country{France}
  \postcode{31400}
}

\author{Joao Marques-Silva}
\authornote{Both authors contributed equally to this research.}
\email{joao.marques-silva@irit.fr}
\orcid{0000-0002-6632-3086}
\affiliation{%
  \institution{IRIT, CNRS}
  %\streetaddress{No Street}
  \city{Toulouse}
  %\state{No State}
  \country{France}
  \postcode{31400}
}

%% \author{Lars Th{\o}rv{\"a}ld}
%% \affiliation{%
%%   \institution{The Th{\o}rv{\"a}ld Group}
%%   \streetaddress{1 Th{\o}rv{\"a}ld Circle}
%%   \city{Hekla}
%%   \country{Iceland}}
%% \email{larst@affiliation.org}

%% \author{Valerie B\'eranger}
%% \affiliation{%
%%   \institution{Inria Paris-Rocquencourt}
%%   \city{Rocquencourt}
%%   \country{France}
%% }

%% \author{Aparna Patel}
%% \affiliation{%
%%  \institution{Rajiv Gandhi University}
%%  \streetaddress{Rono-Hills}
%%  \city{Doimukh}
%%  \state{Arunachal Pradesh}
%%  \country{India}}

%% \author{Huifen Chan}
%% \affiliation{%
%%   \institution{Tsinghua University}
%%   \streetaddress{30 Shuangqing Rd}
%%   \city{Haidian Qu}
%%   \state{Beijing Shi}
%%   \country{China}}

%% \author{Charles Palmer}
%% \affiliation{%
%%   \institution{Palmer Research Laboratories}
%%   \streetaddress{8600 Datapoint Drive}
%%   \city{San Antonio}
%%   \state{Texas}
%%   \country{USA}
%%   \postcode{78229}}
%% \email{cpalmer@prl.com}

%% \author{John Smith}
%% \affiliation{%
%%   \institution{The Th{\o}rv{\"a}ld Group}
%%   \streetaddress{1 Th{\o}rv{\"a}ld Circle}
%%   \city{Hekla}
%%   \country{Iceland}}
%% \email{jsmith@affiliation.org}

%% \author{Julius P. Kumquat}
%% \affiliation{%
%%   \institution{The Kumquat Consortium}
%%   \city{New York}
%%   \country{USA}}
%% \email{jpkumquat@consortium.net}

\renewcommand{\shortauthors}{Marques-Silva and Huang}

\begin{abstract}
  Recent work demonstrated the inadequacy of Shapley values for
  explainable artificial intelligence (XAI). Although to disprove a
  theory a single counterexample suffices, a possible criticism of
  earlier work is that the focus was solely on Boolean classifiers.
  To address such possible criticism, this paper demonstrates the
  inadequacy of Shapley values for families of classifiers where
  features are not boolean, but also for families of classifiers for
  which multiple classes can be picked.
  Furthermore, the paper shows that the features changed in any
  minimal $l_0$ distance adversarial examples do not include
  irrelevant features, thus offering further arguments regarding the 
  inadequacy of Shapley values for XAI.
\end{abstract}




%\received{31 July 2023}
%\received[revised]{12 March 2023}
%\received[accepted]{5 June 2023}

\maketitle

\renewcommand\appendixtocname{Appendix}
\renewcommand{\contentsname}{Overview}
\setcounter{tocdepth}{1}
\tableofcontents\clearpage
\let\addcontentsline\oldaddcontentsline% Restore \addcontentsline

\section{Introduction} \label{sec:intro}

A number of recent reports~\cite{hms-corr23a,msh-corr23,hms-corr23b}
has demonstrated that, for some classifiers, Shapley values for
XAI~\cite{conklin-asmbi01,kononenko-jmlr10,kononenko-kis14,lundberg-nips17,barcelo-aaai21,vandenbroeck-aaai21,vandenbroeck-jair22,barcelo-jmlr23}
produce measures of relative feature importance that are uncorrelated
with measures of feature relevancy, as proposed in the context of
abductive reasoning~\cite{gottlob-jacm95,hiims-kr21,hcmpms-tacas23}.

Methods of XAI can be broadly characterized as based on \emph{feature
attribution}, as exemplified by the use of Shapley 
values~\cite{lundberg-nips17}, or based on \emph{feature selection}.
Methods of feature selection include informal
approaches~\cite{guestrin-aaai18}, but also formal logic-based
approaches~\cite{inms-aaai19}.
Whereas feature attribution assigns a score to each feature as a
measure of its effective importance to a prediction, feature selection
identifies a subset of features which are deemed sufficient for a
prediction. 
Abductive explanations~\cite{inms-aaai19} provide a rigorous,
model-accurate, method for computing explanations based on feature 
selection. Abductive explanations are grounded on logic-based
abduction~\cite{gottlob-jacm95}, which can be traced to the seminal 
work of Peirce on abduction~\cite{peirce-works31}.

The results mentioned above~\cite{hms-corr23a,msh-corr23,hms-corr23b}
can be restated as follows: for some classifiers, formal definitions
of feature importance based on feature selection are uncorrelated with
axiomatic definitions of feature importance based on feature
attribution as exemplified by Shapley values for XAI.
More concretely, it has been shown
that~\cite{hms-corr23a,msh-corr23,hms-corr23b}: i) features that are
\emph{irrelevant} for a prediction can be assigned feature importance
of greater absolute value than features that are \emph{relevant} for
that prediction, and ii) features that are \emph{relevant} for a
prediction can be assigned \emph{no} importance even when
\emph{irrelevant} features assigned \emph{some} importance.
(Recall that a feature is relevant if it occurs in some abductive
explanation; otherwise it is
irrelevant~\cite{gottlob-jacm95,hiims-kr21,hcmpms-tacas23}.)

An immediate corollary of these recent results is that relative
measures of feature importance based on feature selection (and defined
using the concept of feature relevancy in abductive
reasoning~\cite{gottlob-jacm95}) cannot in general be related with
relative measures of feature importance based on feature attribution
(as obtained with Shapley values for XAI).

One might contend that the fact that the two measures of relative
feature importance cannot be compared is not a major issue per se.
However, earlier work~\cite{hms-corr23a,msh-corr23,hms-corr23b} argued
that irrelevant features should be deemed as having no feature
importance, and that relevant features should have some sort of
feature importance.
This report provides additional insights on how to make this argument
more intuitive.
Furthermore, this report evaluates the role of features in finding
minimal-distance adversarial examples, and shows that irrelevant
features need never be changed for finding adversarial examples,
i.e.\ those features do not occur in minimal Hamming ($l_0$) distance
adversarial examples\footnote{%
In this paper, features are assumed not to be real-valued. Other
distances could be considered for real-valued features. Furthermore,
and similarly to earlier
work~\cite{goodfellow-corr16a,goodfellow-corr16b,carlini-woot17,carlini-corr17,kwiatkowska-ijcai19,yeung-iccv21}
we will seek adversarial examples offering guarantees of minimality,
either cardinality or subset-minimality.}.
These observations further support earlier arguments about Shapley
values for XAI providing misleading information about relative feature
importance \cite{hms-corr23a,msh-corr23,hms-corr23b}.
A conclusion of the arguments proposed in earlier work and in this
paper is that Shapley values for XAI can offer human decision-makers
misleading information regarding relative feature importance.

Finally, one possible drawback of earlier results is that the obtained
counterexamples consist of (arbitrary many) boolean classifiers.
Hence, a natural question is whether earlier
results~\cite{hms-corr23a,msh-corr23,hms-corr23b} extend beyond
boolean classifiers. This paper studies a number of non-boolean 
classifiers, and shows that the conclusions of earlier work also
apply to those non-boolean classifiers. 

The paper is organized as follows. \cref{sec:prelim} overviews the 
notation and definitions introduced in earlier
work~\cite{hms-corr23a,msh-corr23,hms-corr23b}.
\cref{sec:tt:exs} analyzes example classifiers defined with
generalized tabular representations.
\cref{sec:dt:exs} summarizes results obtained on publicly available
decision trees.
Moreover, \cref{sec:mdd:exs} summarizes results in the case of
OMDD (ordered multi-valued decidion diagram) classifiers.
Moreover, \cref{sec:threat} discusses a number of suggested threats to
validity of the results in this report.
\cref{sec:conc} concludes the paper.

%
%(...)\\
%

\section{Preliminaries} \label{sec:prelim}

We consider the notation and definitions used in earlier
work~\cite{inms-aaai19,barcelo-aaai21,msi-aaai22,ms-rw22,barcelo-jmlr23,hms-corr23a,hms-corr23b,ms-iceccs23,msh-corr23}.
These are briefly overviewed next, and borrow extensively
from~\cite{msh-corr23}.

\subsection{Classification Problems} \label{ssec:clf}

A classification problem is defined on a set of features
$\fml{F}=\{1,\ldots,m\}$, and a set of classes
$\fml{K}=\{c_1,\ldots,c_K\}$.
Each feature $i\in\fml{F}$ takes values from a domain $\mbb{D}_i$.
Domains can be ordinal (e.g.\ real- or integer-valued) or
categorical. 
Feature space is defined by the cartesian product of the domains of
the features: $\mbb{F}=\mbb{D}_1\times\cdots\times\mbb{D}_m$.
A classifier $\fml{M}$ computes a (non-constant) classification
function: $\kappa:\mbb{F}\to\fml{K}$\footnote{%
A classifier that computes a constant function, i.e.\ the same
prediction for all points in feature space, is of course
uninteresting, and so it is explicitly disallowed.}.
A classifier $\fml{M}$ is associated with a tuple
$(\fml{F},\mbb{F},\fml{K},\kappa)$.
For the purposes of this paper, we restrict $\kappa$ to be a
non-constant boolean function. This restriction does not in any way
impact the validity of our results.

Given a classifier $\fml{M}$, and a point $\mbf{v}\in\mbb{F}$, with
$c=\kappa(\mbf{v})$ and $c\in\fml{K}$, $(\mbf{v},c)$ is referred to as
an \emph{instance} (or sample). An explanation problem $\fml{E}$ is
associated with a tuple $(\fml{M},(\mbf{v},c))$.
As a result, $\mbf{v}$ represents a concrete point in feature space,
whereas $\mbf{x}\in\mbb{F}$ represents an arbitrary point in feature
space.

\subsection{Formal Explanations} \label{ssec:xps}

The presentation of formal explanations follows recent
accounts \cite{ms-rw22}.
In the context of XAI, abductive explanations (AXp's) have been
studied since 2018~\cite{darwiche-ijcai18,inms-aaai19}\footnote{%
Initial work considered prime implicants of boolean
classifiers~\cite{darwiche-ijcai18}. Later work~\cite{inms-aaai19}
formulated explanations in terms of abductive reasoning, and
considered a much wider range of classifiers.}. Similar to
other heuristic approaches, e.g.\ Anchors~\cite{guestrin-aaai18},
abductive explanations are an example of explainability by feature
selection, i.e.\ a subset of features is selected as the explanation.
AXp's represent a rigorous example of explainability by feature
selection, and can be viewed as the answer to a ``\emph{\rhlight{Why
  (the prediction, given $\mbf{v}$)?}}'' question. 
An AXp is defined as a subset-minimal (or irreducible) set of features
$\fml{X}\subseteq\fml{F}$ such that the features in $\fml{X}$ are
sufficient for the prediction, given $\mbf{v}$.
This is to say that, if the features in $\fml{X}$ are fixed to the
values determined by $\mbf{v}$, then the prediction is guaranteed to
be $c=\kappa(\mbf{v})$.
The sufficiency for the prediction can be stated formally:
\begin{equation} \label{eq:waxp}
  \forall(\mbf{x}\in\mbb{F}).\left[\bigland\nolimits_{i\in\fml{X}}(x_i=v_i)\right]\limply(\kappa(\mbf{x})=\kappa(\mbf{v}))
\end{equation}
For simplicity, we associate a predicate $\waxp$ with~\eqref{eq:waxp}, 
such that $\waxp(\fml{X})$ holds if and only if~\eqref{eq:waxp} holds.

Observe that \eqref{eq:waxp} is monotone on $\fml{X}$, and so the two
conditions for a set $\fml{X}\subseteq\fml{F}$ to be an AXp
(i.e.\ sufficiency for prediction and subset-minimality), can be
stated as follows:
\begin{align} \label{eq:axp}
  &\forall(\mbf{x}\in\mbb{F}).\left[\bigland\nolimits_{i\in\fml{X}}(x_i=v_i)\right]\limply(\kappa(\mbf{x})=\kappa(\mbf{v}))\land\\
  &
  \forall(t\in\fml{X}).\exists(\mbf{x}\in\mbb{F}).\left[\bigland\nolimits_{i\in\fml{X}\setminus\{t\}}(x_i=v_i)\right]\land(\kappa(\mbf{x})\not=\kappa(\mbf{v}))
  \nonumber
  %\\
\end{align}
Moreover, a predicate $\axp:2^{\fml{F}}\to\{0,1\}$ is associated with
\eqref{eq:axp}, such that $\axp(\fml{X};\fml{E})$ holds true if and
only if \eqref{eq:axp} holds true\footnote{% 
When defining concepts, we will show the necessary parameterizations.
However, in later uses, those parameterizations will be omitted, for
simplicity.}.

An AXp can be interpreted as a logic rule of the form:
\begin{equation}
  \tn{IF} \quad \left[\bigland\nolimits_{i\in\fml{X}}(x_i=v_i)\right] \quad
  \tn{THEN} \quad (\kappa(\mbf{x})=c) %%\kappa(\mbf{v})
\end{equation}
where $c=\kappa(\mbf{v})$.
It should be noted that informal XAI methods have also proposed the
use of IF-THEN rules~\cite{guestrin-aaai18} which, in the case of
Anchors~\cite{guestrin-aaai18} may or may not be
sound~\cite{inms-aaai19,ignatiev-ijcai20}. In contrast, rules obtained
from AXp's are logically sound.

Alternatively, contrastive explanations (CXp's) represent a type of
explanation that differs from AXp's, in that CXp's answer a
``\emph{\rhlight{Why Not (some other prediction, given $\mbf{v}$)?}}''
question~\cite{miller-aij19,inams-aiia20}, again given $\mbf{v}$.
Given a set $\fml{Y}\subseteq\fml{F}$, sufficiency for changing the
prediction can be stated formally:
\begin{equation} \label{eq:wcxp}
  \exists(\mbf{x}\in\mbb{F}).\left[\bigland\nolimits_{i\in\fml{F}\setminus\fml{Y}}(x_i=v_i)\right]\land(\kappa(\mbf{x})\not=\kappa(\mbf{v}))
\end{equation}
For simplicity, we associate a predicate $\wcxp$ with~\eqref{eq:wcxp}, 
such that $\wcxp(\fml{Y})$ holds if and only if~\eqref{eq:wcxp} holds.

A CXp is a subset-minimal set of features which, if allowed to take a
value other than the value determined by $\mbf{v}$, then the
prediction can be changed by choosing suitable values to those
features.

Similarly to the case of AXp's, for CXp's \eqref{eq:wcxp} is monotone
on $\fml{Y}$, and so the two conditions (sufficiency for changing the
prediction and subset-minimality) can be stated formally as follows:
\begin{align} \label{eq:cxp}
  &\exists(\mbf{x}\in\mbb{F}).\left[\bigland\nolimits_{i\in\fml{F}\setminus\fml{Y}}(x_i=v_i)\right]\land(\kappa(\mbf{x})\not=\kappa(\mbf{v}))\land\\
  &
  \forall(t\in\fml{Y}).\forall(\mbf{x}\in\mbb{F}).\left[\bigland\nolimits_{i\in\fml{F}\setminus(\fml{Y}\setminus\{t\})}(x_i=v_i)\right]\limply(\kappa(\mbf{x})=\kappa(\mbf{v}))
  \nonumber
  %\\
\end{align}
A predicate $\cxp:2^{\fml{F}}\to\{0,1\}$ is associated with
\eqref{eq:cxp}, such that $\cxp(\fml{Y};\fml{E})$ holds true if and
only if \eqref{eq:cxp} holds true.

Algorithms for computing AXp's and CXp's for different families of
classifiers have been proposed in recent years (\cite{msi-aaai22}
provides a recent account of the progress observed in computing formal
explanations). These algorithms include the use of automated reasoners
(e.g.\ SAT, SMT or MILP solvers), or dedicated algorithms for families
of classifiers for which computing one explanation is tractable.

Given an explanation problem $\fml{E}$, the sets of AXp's and CXp's
are represented by:
\begin{align}
  \mbb{A}(\fml{E}) = \{\fml{X}\subseteq\fml{F}\,|\,\axp(\fml{X};\fml{E})\}
  \label{eq:seta}
  \\
  \mbb{C}(\fml{E}) = \{\fml{Y}\subseteq\fml{F}\,|\,\cxp(\fml{Y};\fml{E})\}%\\
  \label{eq:setc}
\end{align}
For example, $\mbb{A}(\fml{E})$ represents the set of \dbhlight{all}
logic rules that predict $c=\kappa(\mbf{v})$, which are consistent
with $\mbf{v}$, and which are irreducible (i.e.\ no literal $x_i=v_i$
can be discarded).

Furthermore, it has been proved~\cite{inams-aiia20} that (i) a set
$\fml{X}\subseteq\fml{F}$ is an AXp if and only if it is a minimal
hitting set (MHS) of the set of CXp's; and (ii) a set
$\fml{Y}\subseteq\fml{F}$ is a CXp if and only if it is an MHS of the
set of AXp's.
This property is referred to as MHS duality, and can be traced back to
the seminal work of R.~Reiter~\cite{reiter-aij87} in model-based
diagnosis.
Moreover, MHS duality has been shown to be instrumental for the
enumeration of AXp's and CXp's, but also for answering other
explainability queries~\cite{ms-rw22}.

\subsection{Shapley Values for XAI} \label{ssec:svs}

Shapley values were proposed in the 1950s, in the context of game
theory~\cite{shapley-ctg53}, and find a wealth of
uses~\cite{roth-bk88}.
More recently, starting in 2001, Shapley values have been extensively
used for explaining the predictions of ML models,
e.g.~\cite{conklin-asmbi01,kononenko-jmlr10,kononenko-kis14,zick-sp16,lundberg-nips17,jordan-iclr19,taly-cdmake20,lakkaraju-nips21,watson-facct22},
among a vast number of recent examples (see~\cite{hms-corr23a} for a
more comprehensive list of references).
Shapley values represent one example of explainability by feature
attribution, i.e.\ some score is assigned to each feature as a form of
explanation.
The complexity of computing Shapley values (as proposed in
SHAP~\cite{lundberg-nips17}) has been studied in recent
years~\cite{barcelo-aaai21,vandenbroeck-aaai21,vandenbroeck-jair22,barcelo-jmlr23}.
This section provides a brief overview of how Shapley values for
explainability are computed.
Throughout, we build on the notation used in recent
work~\cite{barcelo-aaai21,barcelo-jmlr23}, which builds on the work 
of~\cite{lundberg-nips17}.

Let $\Upsilon:2^{\fml{F}}\to2^{\mbb{F}}$ be defined by,
%
%\footnote{%
%When defining concepts, we will show the necessary parameterizations.
%However, in later uses, those parameterizations will be omitted, for
%simplicity.},
%
\begin{equation} \label{eq:upsilon}
  \Upsilon(\fml{S};\mbf{v})=\{\mbf{x}\in\mbb{F}\,|\,\land_{i\in\fml{S}}x_i=v_i\}
\end{equation}
i.e.\ for a given set $\fml{S}$ of features, and parameterized by
the point $\mbf{v}$ in feature space, $\Upsilon(\fml{S};\mbf{v})$
denotes all the points in feature space that have in common with
$\mbf{v}$ the values of the features specified by $\fml{S}$.
%
%Observe that $\Upsilon$ is also used (implicitly) for picking the set
%of rows where are interested in when computing explanations
%(see~\cref{tab:ex01}).

Also, let $\phi:2^{\fml{F}}\to\mbb{R}$ be defined by,
\begin{equation} \label{eq:phi}
  \phi(\fml{S};\fml{M},\mbf{v})=\frac{1}{2^{|\fml{F}\setminus\fml{S}|}}\sum\nolimits_{\mbf{x}\in\Upsilon(\fml{S};\mbf{v})}\kappa(\mbf{x})
\end{equation}
Thus, given a set $\fml{S}$ of features,
$\phi(\fml{S};\fml{M},\mbf{v})$ represents the average value of the
classifier over the points of feature space represented by
$\Upsilon(\fml{S};\mbf{v})$.
%
%Clearly, $\phi$ gives the average (expected) value of the classifier
%in the point given by $\Upsilon(\fml{S}$.
%
The formulation presented in earlier
work~\cite{barcelo-aaai21,barcelo-corr21} allows for different input
distributions when computing the average values. For the purposes of
this paper, it suffices to consider solely a uniform input
distribution, and so the dependency on the input distribution is not
accounted for. 
%However, assuming a uniform distribution suffices for the purposes of
%this paper.
%

To simplify the notation, the following definitions are used
throu\-ghout,
\begin{align}
  \Delta(i, \fml{S}; \fml{M},\mbf{v}) & =
  \left( \phi(\fml{S}\cup\{i\};\fml{M},\mbf{v}) -
  \phi(\fml{S};\fml{M},\mbf{v}) \right) \\  
  \varsigma(\fml{S};\fml{M},\mbf{v}) & =
  \sfrac{|\fml{S}|!(|\fml{F}|-|\fml{S}|-1)!}{|\fml{F}|!}
\end{align}

Finally, let $\sv:\fml{F}\to\mbb{R}$, i.e.\ the Shapley value for
feature $i$, be defined by,
%\footnote{%
%We distinguish $\shap(\cdot;\cdot,\cdot)$ from $\sv(\cdot;\cdot,\cdot)$. Whereas
%$\shap(\cdot;\cdot,\cdot)$ represents the value computed by the tool
%SHAP~\cite{lundberg-nips17}, $\sv(\cdot;\cdot,\cdot)$ represents the Shapley
%value in the context of (feature attribution based) explainability, as
%studied in a number of
%works~\cite{kononenko-jmlr10,kononenko-kis14,lundberg-nips17,barcelo-aaai21,vandenbroeck-aaai21,vandenbroeck-jair22}. Thus, 
%$\shap(\cdot;\cdot,\cdot)$ is a heuristic approximation of $\sv(\cdot;\cdot,\cdot)$.},
%
\begin{equation} \label{eq:sv}
  %\sv(i;\fml{M},\mbf{v})=\sum_{\fml{S}\subseteq(\fml{F}\setminus\{i\})}\frac{|\fml{S}|!(|\fml{F}|-|\fml{S}|-1)!}{|\fml{F}|!}\left(\phi(\fml{S}\cup\{i\};\fml{M},\mbf{v})-\phi(\fml{S};\fml{M},\mbf{v})\right)
  \sv(i;\fml{M},\mbf{v})=\sum\nolimits_{\fml{S}\subseteq(\fml{F}\setminus\{i\})}\varsigma(\fml{S};\fml{M},\mbf{v})\times\Delta(i,\fml{S};\fml{M},\mbf{v})
\end{equation}
Given an instance $(\mbf{v},c)$, the Shapley value assigned to each
feature measures the \emph{contribution} of that feature with respect
to the prediction. 

Throughout this paper, we use the term \emph{Shapley values} to refer
to the SHAP scores studied in earlier
work~\cite{lundberg-nips17,barcelo-aaai21,vandenbroeck-aaai21,vandenbroeck-jair22,barcelo-jmlr23}.
This is to emphasize the different between Shapley values for XAI (or
SHAP scores in earlier work), and the values computed by the tool
SHAP~\cite{lundberg-nips17}. As demonstrated in recent
work~\cite{hms-corr23a}, there can exist significant differences
between Shapley values and the results produced by the tool SHAP.

\subsection{Feature (Ir)relevancy \& Necessity}
\label{ssec:xpr}

Given \eqref{eq:seta} and \eqref{eq:setc}, we can aggregate the
features that occur in AXp's and CXp's:
\begin{align}
  \fml{F}_{\mbb{A}(\fml{E})}=\bigcup\nolimits_{\fml{X}\in\mbb{A}(\fml{E})}\fml{X}
  \\
  \fml{F}_{\mbb{C}(\fml{E})}=\bigcup\nolimits_{\fml{Y}\in\mbb{C}(\fml{E})}\fml{Y}
\end{align}
Moreover, MHS duality between the sets of AXp's and CXp's allows
proving that: $\fml{F}_{\mbb{A}(\fml{E})}=\fml{F}_{\mbb{C}(\fml{E})}$.
Hence, we just refer to $\fml{F}_{\mbb{A}(\fml{E})}$ as the set of
features that are contained in some AXp (or CXp).

A feature $i\in\fml{F}$ is relevant if it is contained in some AXp,
i.e.\ $i\in\fml{F}_{\mbb{A}(\fml{E})}=\fml{F}_{\mbb{C}(\fml{E})}$;
otherwise it is irrelevant,
i.e.\ $i\not\in\fml{F}_{\mbb{A}(\fml{E})}$.
A feature is necessary if it is contained in all AXp's%
\footnote{%
It should be noted that feature relevancy and necessity mirror the
concepts of relevancy and necessity studied in logic-based 
abduction~\cite{gottlob-jacm95}.}.
We will use the predicate $\relevant(i)$ to denote that feature $i$ is
relevant, and predicate $\irrelevant(i)$ to denote that feature $i$ is
irrelevant.

Relevant and irrelevant features provide a fine-grained
characterization of feature importance, in that irrelevant features
play no role whatsoever in prediction sufficiency.
In fact, if $p\in\fml{F}$ is an irrelevant feature, then we can write: 
\begin{align}
  \forall(&\fml{X}\in\mbb{A}(\fml{E})).
  \forall(u_p\in\mbb{D}_p).
  \forall(\mbf{x}\in\mbb{F}).\nonumber\\
  &\left[\bigland\nolimits_{i\in\fml{X}}(x_i=v_i)\land
  %%\nonumber\\
  (x_p=u_p)\right]\limply(\kappa(\mbf{x})=\kappa(\mbf{v}))
  %%\kappa(v_1,\ldots,v_{p-1},u_p,v_{p+1},\ldots,v_m))
\end{align}
The logic statement above clearly states that, if we fix the values of
the features identified by any AXp then, no matter the value picked for
feature $p$, the prediction is guaranteed to be $c=\kappa(\mbf{v})$.
The bottom line is that an irrelevant feature $p$ is absolutely
unimportant for the prediction, and so there is no reason to include 
it in a logic rule consistent with the instance.

As argued in earlier work~\cite{hms-corr23a,msh-corr23,hms-corr23b},
the fact that irrelevant features are not considered in explanations
means their value is absolutely unimportant for either keeping or
changing the prediction.

\condaddval{
  \paragraph{Validating the results.}
  %~\\
  To validate the results, the following must
  hold~\cite{kononenko-kis14,vandenbroeck-jair22}:
  \begin{equation} \label{eq:vval}
    \sum_{i}\sv(i)+\phi(\emptyset) = \kappa(\mbf{v})
  \end{equation}
  %%
  %%\jnote{Where was this proved?}
}

%\clearpage
%\section{Adversarial Examples, Explanations \& Shapley Values}
\section{Adversarial Examples vs (Ir)relevant Features}
\label{sec:aex}

This section develops a number of results regarding
the non-importance of irrelevant features for adversarial
examples. These results offer further support to the claims of
inadequacy of Shapley values in this and earlier
reports~\cite{hms-corr23a,msh-corr23,hms-corr23b}.

As indicated earlier, we consider categorical or discrete features and
Hamming distance as a measure of distance between points in feature
space. The Hamming distance is also referred to as the $l_0$ measure
of distance, and it is defined as follows:
\begin{equation} \label{eq:l0}
  ||\mbf{x}-\mbf{y}||_{0}\triangleq \sum_{i=1}^m \tn{ITE}(x_i\not=y_i,1,0)
\end{equation}

Given a point $\mbf{v}$ in feature space, an adversarial example (AE)
is some other point $\mbf{x}$ in feature space that changes the
prediction and such that the measure of distance $l_p$ between the two
points is small enough:
\begin{equation} \label{eq:ae}
  \exists(\mbf{x}\in\mbb{F}).||\mbf{x}-\mbf{v}||_{p}\le\epsilon\land(\kappa(\mbf{x})\not=\kappa(\mbf{v}))
\end{equation}
(in our case, we consider solely $p=0$.)
Although we could consider specific values of $\epsilon$, as proposed
in~\cite{hms-corr23c}, we will opt in this paper for allowing
$\epsilon=+\infty$, and then asking for adversarial examples
respecting some criterion of minimality.

The features that are changed for a given AE in~\eqref{eq:ae} are
denoted by $\fml{A}\subseteq\fml{F}$. Thus, if we say that $\fml{A}$
is an adversarial example, then~\eqref{eq:ae} holds true for some
$\mbf{x}$ such that $\mbf{x}$ and $\mbf{v}$ differ in the values of
the features included in $\fml{A}$.

Since we can represent AEs as sets (of the features that change their
value), we will consider subset-minimal AEs, i.e.\ sets of features
that represent adversarial examples, and no proper subset represents
an adversarial example.

%% \[
%% \mbb{X}=\{\}
%% \]

%% Given the above, the following propositions can be proved.

%% \begin{proposition}
%%   The following statements hold true:
%%   \begin{enumerate}
%%   \item There exists an AE of s ...
%%   \end{enumerate}
%% \end{proposition}

\begin{proposition}
  Given an instance $(\mbf{v},c)$, if $\fml{A}$ is an AE and
  $j\in\fml{A}$ is an irrelevant feature, then there exists another AE
  $\fml{B}$ with $\fml{B}\subsetneq\fml{A}$, with $j\not\in\fml{B}$.
\end{proposition}

\begin{corollary}
  Subset- or cardinality-minimal AEs do not contain irrelevant
  features.
\end{corollary}

We can strengthen the above results, by analyzing instead feature
relevancy.

\begin{proposition}
  A feature $j\in\fml{F}$ is included in some (minimal) adversarial
  example iff feature $j$ is relevant.
\end{proposition}

\begin{remark}
  Thus, we conclude that there is a tight relationship between
  adversarial examples and feature relevancy, and so with abductive
  and contrastive explanations, all of which relate with either
  keeping or changing the prediction.
  In contrast, the relative order of importance provided by Shapley
  values is not related neither with abductive explanations, nor with 
  contrastive explanations. Finally, given the results in this and
  earlier reports~\cite{hms-corr23a,msh-corr23,hms-corr23b}, Shapley
  values for XAI are also not related with $l_0$-minimal adversarial
  examples.
\end{remark}

%\section{Truth Table Based Example Classifiers}
%\section{Classifiers Defined by Truth Tables}
\section{Classifiers Defined by Tabular Representations}
\label{sec:tt:exs}

Building on earlier work~\cite{hms-corr23a,msh-corr23,hms-corr23b},
this section analyzes several additional examples, further extending
the earlier results on the inadequacy of Shapley values for XAI.

\subsection{Example of Multi-Valued Classifier} \label{ssec:ex:01}

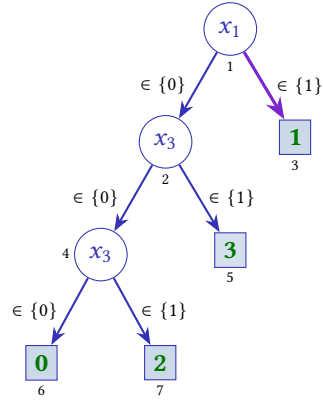
\begin{figure}[t]
  \begin{subtable}[b]{0.45\textwidth} %[ht]
    \begin{center}
      \begin{tabular}{ccccc} \toprule
        row \# & $x_1$ & $x_2$ & $x_3$ & $\kappa_1(\mbf{x})$ \\ \toprule
        1 & 0 & 0 & 0 & 0 \\
        2 & 0 & 0 & 1 & 3 \\
        3 & 0 & 1 & 0 & 2 \\
        4 & 0 & 1 & 1 & 3 \\
        5 & 1 & 0 & 0 & 1 \\
        6 & 1 & 0 & 1 & 1 \\
        7 & 1 & 1 & 0 & 1 \\
        8 & 1 & 1 & 1 & 1 \\
        \bottomrule
      \end{tabular}
    \end{center}

    \begin{tabular}{c} ~\\[2pt] ~\\[3pt] \end{tabular}
    \caption{Tabular representation for $\kappa_1$} \label{tab:01:tt}
  \end{subtable}
  \hfill
  \begin{subfigure}[b]{0.45\textwidth}
    \begin{center}
      % Concocted example
%%
%\tikzset{every label/.style={xshift=-0.35ex,
%  yshift=-5.25ex,
%  text width=1ex,
%  align=right, inner sep=1pt, font=\tiny, text=midblue}}
%%
%\tikzset{tlabel/.style={xshift=0.25ex, yshift=1.75ex, text width=1ex,
%    align=right, inner sep=1pt, font=\tiny, text=midblue}}
%%%\tikzset{every node/.style={---rectangle---}}
%
\forestset{
  BDT/.style={
    for tree={
      l=1.5cm,s sep=1.15cm,
      if n children=0{}{circle}, %rectangle
      %if n children=0{}{draw},
      draw=midblue,%draw=black,%
      text=midblue,%text=black,%
      edge={
        my edge
      },
      %if n=1{
      %  edge+={0 my edge},
      %}{},
      edge=thick,
    }
  },
}
\begin{forest}
  BDT
  [{$x_1$}, label={[yshift=-6.875ex]{{\tiny1}}} 
    [{$x_3$}, label={[yshift=-6.875ex]{{\tiny2}}}, %edge={very thick}, %top-left=x
      edge label={node[midway,left,xshift=-0.5pt] {{\scriptsize$\in\{0\}$}}}
      [{$x_3$}, label={[xshift=-3.075ex,yshift=-3.5ex]{{\tiny4}}}, %yshift=-6.875ex
        edge label={node[midway,left,xshift=-1.5pt] {{\scriptsize$\in\{0\}$}}}
        [\dghlight{\textbf{0}}, label={[yshift=-5.25ex]{{\tiny6}}},
          edge label={node[midway,left,xshift=-0.5pt] {{\scriptsize$\in\{0\}$}}}, rectangle, fill={tblue2!25} ]
        [\dghlight{\textbf{2}}, label={[yshift=-5.25ex]{{\tiny7}}},
          edge label={node[midway,right,xshift=-0.575pt] {{\scriptsize$\in\{1\}$}}}, rectangle, fill={tblue2!25} ]
      ]
      [\dghlight{\textbf{3}}, label={[yshift=-5.25ex]{{\tiny5}}},
        edge label={node[midway,right,xshift=-0.5pt] {{\scriptsize$\in\{1\}$}}},
        rectangle, fill={tblue2!20} ]
    ]
    [\dghlight{\textbf{1}}, label={[yshift=-5.25ex]{{\tiny3}}},
      edge={very thick, draw=purple3}, edge label={node[midway,right,xshift=0.5pt] {{\scriptsize$\in\{1\}$}}},
      rectangle, fill={tblue2!25} ]
  ]
\end{forest}
    \end{center}
    \caption{Decision tree for classifier $\kappa_1$} \label{fig:01:dt}
  \end{subfigure}
  \caption{Multi-valued classifier. The DT path $\langle1,3\rangle$,
    which is consistent with instance $((1,0,0),1)$, is highlighted.}
  \label{fig:01:clssf}
\end{figure}

%%\begin{example} \label{ex:01a}
%
\paragraph{Classifier.}
%~\\
We consider the following multi-valued classifier, defined on
boolean features, with $\mbb{D}_i=\mbb{B}=\{0,1\}$, with
$1\le{i}\le{m}$:
\[
\kappa_1(x_1,x_2,\dots,x_m)=
\left\{
\begin{array}{lcl}
  1 & \quad & \tn{if $x_1=1$} \\[5pt]
  \max\{i\,|\,x_i>0\land1<{i}\le{m}\} & \quad & \tn{otherwise}
\end{array}
\right.
\]
Although the classifier is defined on $m$ features, throu\-gh\-out we
will consider $m=3$, to facilitate the analysis of the main claims.
Thus, $\mbb{F}=\mbb{B}^{3}$.
Also, we consider the instance $((1,0,0),1)$.
The classifier is depicted in~\cref{fig:01:clssf}, with a
(multi-valued) tabular representation (TR) is shown
in~\cref{tab:01:tt}, and a decision tree (DT) is shown
in~\cref{fig:01:dt}.
Given the TR/DT, we set $\fml{K}=\{0,1,2,3\}$.

\paragraph{Feature influence in predicted class.}
%~\\
Recall that the instance is $((1,0,0),1)$, and so the predicted
class is 1.
By inspection of the DT, it is simple to conclude that, for any
point in feature space, the predicted class is 1 if and only of
$x_1=1$, and that the predicted class is other than 1 if and only if
$x_1=0$. These statements hold true \emph{independently} of the
values assigned to features 2 and 3. Thus, to keep the predicted
class only the value of feature 1 matters. Similarly, to change the
predicted class, only the value of feature 1 matters.

\paragraph{Formal explanations \& feature relevancy.}
%~\\
\cref{tab:01:xps} (see~\cpageref{tab:01:xps}) illustrates the
role of each set of features in terms of explanation sufficiency
and irredundancy.
The computed explanations also serve for deciding feature
(ir)relevancy.
Unsurprisingly, feature 1 is shown to be relevant (and necessary), and
features 2 and 3 are shown to be irrelevant.
%
%%\end{example}
%
As can be concluded, feature 1 is sufficient for ensuring that the
predicted class is 1. In contrast, when feature 1 takes value 1, the
other features can be assigned \emph{any} value from their domain,
since that does not change the predicted class.
By subset-minimality (and so invoking Occam's razor), features 2 and 3
are never included in formal explanations.

\begin{table*}[t]
  \begin{mdframed}[linewidth=1.5pt,linecolor=darkblue,roundcorner=5pt]
    \begin{center}
      \scalebox{0.925}{
        \begin{tabular}{ccccc} \toprule[1.0pt]
          Xp set $\fml{S}$ &
          $\fml{S}$ sufficient? &
          $\fml{S}$ irreducible? &
          \makecell{Pick $\fml{X}\subseteq\fml{S}$,\\ $\fml{X}$ sufficient
            \& irreducible} &
          %\lor\fml{X}=\fml{S}
          %\makecell{
          Meaning of $\fml{X}$ %\\ ... }
          relative to $\fml{S}$ \\ \toprule[1.0pt]
          $\emptyset$ & \nomark & -- & -- & -- \\ \midrule[0.25pt]
          $\{1\}$ & \yesmark & \yesmark & $\{1\}$ & $\fml{X}=\fml{S}$ is
          an AXp \\ \midrule[0.25pt]
          $\{2\}$ & \nomark & -- & -- & -- \\ \midrule[0.25pt]
          $\{3\}$ & \nomark & -- & -- & -- \\  \midrule[0.25pt]
          $\{1,2\}$ & \yesmark & \nomark & $\{1\}$ &
          $\begin{aligned}[c]\forall&(u_2\in\mbb{D}_2).\forall(\mbf{x}\in\mbb{F}).\\&[(x_1=1)\land(x_2=u_2)]\limply(\kappa_1(\mbf{x})=1)\end{aligned}$\\ \midrule[0.25pt]
          $\{1,3\}$ & \yesmark & \nomark & $\{1\}$ &
          $\begin{aligned}[c]\forall&(u_3\in\mbb{D}_2).\forall(\mbf{x}\in\mbb{F}).\\&[(x_1=1)\land(x_3=u_3)]\limply(\kappa_1(\mbf{x})=1)\end{aligned}$\\ \midrule[0.25pt]
          $\{2,3\}$ & \nomark & -- & -- & -- \\ \midrule[0.25pt]
          $\{1,2,3\}$ & \yesmark & \nomark & $\{1\}$ &
          $\begin{aligned}[c]\forall&(u_2\in\mbb{D}_2).\forall(u_3\in\mbb{D}_3).\forall(\mbf{x}\in\mbb{F}).\\&[(x_1=1)\land(x_2=u_2)\land(x_3=u_3)]\limply(\kappa_1(\mbf{x})=1)\end{aligned}$\\
          %\begin{array}{l}\end{array}
          \bottomrule[0.8pt]
        \end{tabular}
      }
    \end{center}
    \caption{Detailed analysis of the classifiers
      in~\cref{fig:01:clssf} (and in~\cref{fig:02:clssf,fig:04:clssf}).
      The analysis is exactly the same for all these example
      classifiers.
      %For~\cref{ex:01a,ex:04a}, $\mbb{D}_T=\mbb{B}$. For~\cref{ex:02a},
      %$\mbb{D}_T=\{0,1,2\}$.
    }
    \label{tab:01:xps}
  \end{mdframed}
\end{table*}

%\jnote{Include DT for this classifier, and proceed to analyze operation
%  of classifier.}
%
%\jnote{Analyze influence of features and represent classifier as a set
%  of logic rules. The bottom line: prediction of class 1 is oblivious
%  to values of features 2 and 3.}

%\begin{claim}
%  For~\cref{ex:01a} (recall that the instance is
%  $(\mbf{v},c)=((1,0,0),1)$), feature 1 is relevant, and that all the
%  other features are irrelevant. This is easy to prove, as shown
%  in~\cref{tab:01:xps}, in the case of AXps.
%  %
%  Thus, in terms of features used in explanations, features 2 and 3
%  (for $m=3$) play no role whatsoever either in prediction sufficiency
%  or in changing the prediction.
%\end{claim}

\paragraph{Features in adversarial examples.}
%~\\
\cref{tab:01:aes} summarizes the possible adversarial examples for the
classifier given instance $((1,0,0),1)$. An adversarial example is a
point $\mbf{y}\in\mbb{F}$ that causes the prediction to change, and
for which the Hamming ($l_0$) distance between the two points is
minimized. (As noted earlier, we opt for subset-minimality.) As can be
observed, it suffices to change the value of $x_1$ to ensure that the
prediction changes.

\begin{table}[t] %[ht]
  \begin{tabular}{ccccccc} \toprule
    $x_1$ & $x_2$ & $x_3$ & $\kappa_1(\mbf{x})$ &
    $\kappa_1(\mbf{x})\not=\kappa_1(\mbf{v})$? & $l_0$ distance 
    & AE? \\ \toprule
    0 & 0 & 0 & 0 & \yesmark & 1 & \yesmark \\
    0 & 0 & 1 & 3 & \yesmark & 2 & \nomark \\
    0 & 1 & 0 & 2 & \yesmark & 2 & \nomark \\
    0 & 1 & 1 & 3 & \yesmark & 3 & \nomark \\
    1 & 0 & 0 & 1 & \nomark & -- & -- \\
    1 & 0 & 1 & 1 & \nomark & -- & -- \\
    1 & 1 & 0 & 1 & \nomark & -- & -- \\
    1 & 1 & 1 & 1 & \nomark & -- & -- \\
    \bottomrule
  \end{tabular}
  \caption{AE for $\kappa_1$ on instance $((1,0,0),1)$ for
    classifier from~\cref{tab:01:tt}}
  \label{tab:01:aes}
\end{table}

\paragraph{Shapley values \& feature importance.}
%~\\
\cref{tab:01:svs} summarizes the computation of Shapley values (for
XAI)~\cite{barcelo-aaai21,barcelo-jmlr23,lundberg-nips17} for the
classifier of~\cref{fig:01:clssf} and for the instance $((1,0,0),1)$.
As can be concluded, the relative order of feature importance is
3, 2, 1.

\begin{table}[t]
  \begin{subtable}[b]{0.425\textwidth}
    \begin{tabular}{ccc} \toprule
      $\fml{S}$ & rows picked by $\fml{S}$ & $\phi(\fml{S})$
      \\ \toprule
      $\emptyset$ & 1..8 & $\nfrac{12}{8}=\nfrac{3}{2}$ \\
      $\{1\}$ & 5..8     & $1$ \\
      $\{2\}$ & 1,2,5,6  & $\nfrac{5}{4}$ \\
      $\{3\}$ & 1,3,5,7  & $1$ \\
      $\{1,2\}$ & 5,6    & $1$ \\
      $\{1,3\}$ & 5,7    & $1$ \\
      $\{2,3\}$ & 1,5    & $\nfrac{1}{2}$ \\
      $\{1,2,3\}$ & 5    & $1$ \\
      \bottomrule
    \end{tabular}
    
    \begin{tabular}{c} ~\\[10pt] ~\\[10pt] ~\\[10pt] \end{tabular}
    \caption{Computing average values for the possible sets $\fml{S}$}
    \label{tab:01:svs:sets}
  \end{subtable}
  \begin{subtable}[b]{0.56125\textwidth}
      \renewcommand{\tabcolsep}{0.45em}
      \begin{tabular}{cccccc} \toprule
        \multicolumn{6}{c}{$i=1$} \\
        \midrule
        $\fml{S}_i$ & $\phi(\fml{S}_i)$ & $\phi(\fml{S}_i\cup\{i\})$ &
        $\Delta(i,\fml{S}_i)$ & $\varsigma(\fml{S}_i)$ & $\sv(i)$ \\
        \toprule
        $\emptyset$ & $\nfrac{3}{2}$ & $1$ & $-\nfrac{1}{2}$ &
        $\nfrac{1}{3}$ & -- \\
        $\{2\}$ & $\nfrac{5}{4}$ & $1$ & $-\nfrac{1}{4}$ & $\nfrac{1}{6}$
        & -- \\
        $\{3\}$ & $1$ & $1$ & $0$ & $\nfrac{1}{6}$ & -- \\
        $\{2,3\}$ & $\nfrac{1}{2}$ & $1$ & $\nfrac{1}{2}$ &
        $\nfrac{1}{3}$ & -- \\\midrule[0.25pt]
        & & & & & $-\nfrac{1}{24}=-0.0417$ \\
        \midrule[0.7pt]
        \multicolumn{6}{c}{$i=2$} \\
        \midrule
        $\emptyset$ & $\nfrac{3}{2}$ & $\nfrac{5}{4}$ & $-\nfrac{1}{4}$ &
        $\nfrac{1}{3}$ & -- \\
        $\{1\}$ & $1$ & $1$ & $0$ & $\nfrac{1}{6}$ & -- \\
        $\{3\}$ & $1$ & $\nfrac{1}{2}$ & $-\nfrac{1}{2}$ & $\nfrac{1}{6}$ & -- \\
        $\{1,3\}$ & $1$ & $1$ & $0$ & $\nfrac{1}{3}$ & -- \\\midrule[0.25pt]
        & & & & & $-\nfrac{1}{6}=-0.1667$ \\
        \midrule[0.7pt]
        \multicolumn{6}{c}{$i=3$} \\
        \midrule
        $\emptyset$ & $\nfrac{3}{2}$ & $1$ & $-\nfrac{1}{2}$ &
        $\nfrac{1}{3}$ & -- \\
        $\{1\}$ & $1$ & $1$ & $0$ & $\nfrac{1}{6}$ & -- \\
        $\{2\}$ & $\nfrac{5}{4}$ & $\nfrac{1}{2}$ & $-\nfrac{3}{4}$ & $\nfrac{1}{6}$ & -- \\
        $\{1,2\}$ & $1$ & $1$ & $0$ & $\nfrac{1}{3}$ & -- \\\midrule[0.25pt]
        & & & & & $-\nfrac{7}{24}=-0.2917$ \\
        %\midrule
        \bottomrule
      \end{tabular}
      \caption{Shapley values computed using the intermediate values
        from~\cref{tab:01:svs:sets}}
      \label{tab:01:svs:tab}
  \end{subtable}
  \caption{Computation of Shapley values for the classifier
        $\kappa_1$ of~\cref{fig:01:clssf}} \label{tab:01:svs}
\end{table}

%\begin{claim} 
%  For~\cref{ex:01a}, the relative order of feature importance obtained
%  with absolute Shapley values for
%  XAI~\cite{barcelo-aaai21,barcelo-jmlr23,lundberg-nips17} is:
%  3, 2, 1. This is shown~\cref{tab:01:svs}, where~\cref{tab:01:svs}
%  shows the values of $\phi(\fml{S})$ for the different sets
%  $\fml{S}$; these values are then used in~\cref{tab:01:svs:tab} for
%  computing the Shapley values.
%  Given the computed values, the interpretation that can
%  be made is: feature 3 is more important for the prediction than is
%  feature 2, and feature 3 is more important for the prediction than
%  is feature 1.
%\end{claim}

%\begin{remark}
%  As~\cref{tab:01a} clarifies, irrelevant features, even if included
%  in some proposed explanation would be allowed to be assigned
%  \emph{any} value from their domain. Since such features could be
%  assigned \emph{any} value from their domain, then there is no reason
%  whatsoever to include those features in the explanation to start
%  with. (Also, by Occam's razor these features ought not be included
%  in explanations.)
%\end{remark}

\paragraph{Assessment.}
%~\\
The following observations substantiate our claim that assigning
importance to feature 2 or 3 is misleading for the classifier
of~\cref{fig:01:clssf}:
\begin{enumerate}[nosep]
\item As shown in~\cref{tab:01:xps}, any subset- (or cardinality-)
  minimal set of features that is sufficient for the prediction does
  not contain either feature 2 or feature 3.
\item Motivated by the duality between abductive and contrastive
  explanations~\cite{inams-aiia20}, any subset- (or cardinality-)
  minimal subset of features sufficient for changing the prediction
  does not include either feature 2 or feature 3.
\item A related observation, that offers a somewhat different
  perspective, is that given the relationship between
  (distance-restricted) abductive explanations and adversarial
  examples~\cite{inms-nips19,hms-corr23c}, it is simple 
  to prove that any (subset- or cardinality-) minimal $l_0$ distance
  adversarial example will not include either feature 2 or feature
  3.
\end{enumerate}

%\begin{remark}
%  The reason for why Shapley values can provide misleading information
%  is that Shapley values aggregate information for different subsets
%  of fixed features, of the function's average value, and some of
%  those sets serve no purpose whatsoever in terms of guaranteeing or
%  changing the prediction.
%\end{remark}

\subsection{Example of Discrete Classifier} \label{ssec:ex:02}

\paragraph{Classifier.}
%~\\
%
%\begin{example} \label{ex:02a}
We consider the following discrete classifier, defined on discrete
features, with $\mbb{D}_1=\mbb{B}=\{0,1\}$,
$\mbb{D}_i=\{0,1,2\},i=2,3$.
\[
\kappa_2(x_1,x_2,\dots,x_m)=
\left\{
\begin{array}{lcl}
  1 & \quad & \tn{if $x_1=1$} \\[5pt]
  %%\max\{i\times{x_i}\,|\,2\le{i}\le{m}\} & \quad & \tn{otherwise}
  2  & \quad & \tn{if $(x_1=0)\land(x_2=2)\land(x_3=2)$} \\[5pt]
  0  & \quad & \tn{otherwise}
\end{array}
\right.
\]
%%As before, and although the classifier is defined on $m$ features,
%throu\-gh\-out we will consider $m=3$, to facilitate the analysis of
%the main claims. %%Similarly, we set $T=2$.
%
Given the domains of the features, we have
$\mbb{F}=\mbb{B}\times\mbb{D}_{2}\times\mbb{D}_{3}$.
Furthermore, we consider the instance $((1,2,2),1)$.
The classifier is shown in~\cref{fig:02:clssf}, consisting of a
tabular representation (see~\cref{tab:02:tt}) and a decision
tree~(see~\cref{fig:02:dt})
Given the classifier's description, we set $\fml{K}=\{0,1,2\}$.
Furthermore, \cref{tab:01:xps} (see~\cpageref{tab:01:xps}) illustrates
the role of each set of features in terms of explanations sufficiency
and irredundancy.
%\end{example}

\begin{figure}[t]
  \begin{subtable}[c]{0.375\textwidth} %[ht]
    \begin{center}
      \begin{tabular}{ccccc} \toprule
        row \# & $x_1$ & $x_2$ & $x_3$ & $\kappa_2(\mbf{x})$ \\ \toprule
        1 & 0 & 0 & 0 & 0 \\
        2 & 0 & 0 & 1 & 0 \\
        3 & 0 & 0 & 2 & 0 \\
        4 & 0 & 1 & 0 & 0 \\
        5 & 0 & 1 & 1 & 0 \\
        6 & 0 & 1 & 2 & 0 \\
        7 & 0 & 2 & 0 & 0 \\
        8 & 0 & 2 & 1 & 0 \\
        9 & 0 & 2 & 2 & 2 \\
        10 & 1 & 0 & 0 & 1 \\
        11 & 1 & 0 & 1 & 1 \\
        12 & 1 & 0 & 2 & 1 \\
        13 & 1 & 1 & 0 & 1 \\
        14 & 1 & 1 & 1 & 1 \\
        15 & 1 & 1 & 2 & 1 \\
        16 & 1 & 2 & 0 & 1 \\
        17 & 1 & 2 & 1 & 1 \\
        18 & 1 & 2 & 2 & 1 \\
        \bottomrule
      \end{tabular}
    \end{center}
    \caption{Tabular representation for $\kappa_2$} \label{tab:02:tt}
  \end{subtable}
  \begin{minipage}{0.6\textwidth}
    \begin{subfigure}[c]{1.0\textwidth}%[t]
      \begin{center}
        % Concocted example
%%
%\tikzset{every label/.style={xshift=-0.35ex,
%  yshift=-5.25ex,
%  text width=1ex,
%  align=right, inner sep=1pt, font=\tiny, text=midblue}}
%%
%\tikzset{tlabel/.style={xshift=0.25ex, yshift=1.75ex, text width=1ex,
%    align=right, inner sep=1pt, font=\tiny, text=midblue}}
%%%\tikzset{every node/.style={---rectangle---}}
%
\forestset{
  BDT/.style={
    for tree={
      l=1.5cm,s sep=1.15cm,
      if n children=0{}{circle}, %rectangle
      %if n children=0{}{draw},
      draw=midblue,%draw=black,%
      text=midblue,%text=black,%
      edge={
        my edge
      },
      %if n=1{
      %  edge+={0 my edge},
      %}{},
      edge=thick,
    }
  },
}
\begin{forest}
  BDT
  [{$x_1$}, label={[yshift=-6.875ex]{{\tiny1}}} 
    [{$x_2$}, label={[yshift=-6.875ex]{{\tiny2}}}, %edge={very thick}, %top-left=x
      edge label={node[midway,left,xshift=-0.5pt] {{\scriptsize$\in\{0\}$}}}
      [{$x_3$}, label={[yshift=-6.875ex]{{\tiny4}}}, %xshift=-3.075ex,yshift=-3.5ex
        edge label={node[midway,left,xshift=-1.5pt] {{\scriptsize$\in\{2\}$}}}
        [\dghlight{\textbf{2}}, label={[yshift=-5.25ex]{{\tiny6}}},
          edge label={node[midway,left,xshift=-0.5pt] {{\scriptsize$\in\{2\}$}}}, rectangle, fill={tblue2!25} ]
        [\dghlight{\textbf{0}}, label={[yshift=-5.25ex]{{\tiny7}}},
          edge label={node[midway,right,xshift=-0.575pt] {{\scriptsize$\in\{0,1\}$}}}, rectangle, fill={tblue2!25} ]
      ]
      [\dghlight{\textbf{0}}, label={[yshift=-5.25ex]{{\tiny5}}},
        edge label={node[midway,right,xshift=-0.5pt] {{\scriptsize$\in\{0,1\}$}}},
        rectangle, fill={tblue2!20} ]
    ]
    [\dghlight{\textbf{1}}, label={[yshift=-5.25ex]{{\tiny3}}},
      edge={very thick,draw=purple3}, edge label={node[midway,right,xshift=0.5pt] {{\scriptsize$\in\{1\}$}}},
      rectangle, fill={tblue2!25} ]
  ]
\end{forest}
      \end{center}
      \caption{DT for classifier $\kappa_2$} \label{fig:02:dt}
    \end{subfigure}
    
    \begin{subfigure}[c]{1.0\textwidth}
        \setlength{\fboxrule}{0.875pt}
        \setlength{\fboxsep}{2.5pt}
        \fbox{
          \begin{minipage}{0.975\textwidth}
            Similarly to~\cref{fig:01:clssf}, and for the instance
            $((1,2,2),1)$, for any point in feature space, the
            prediction is class 1 if and only if feature 1 is assigned
            value 1.
          \end{minipage}
        }
        \caption{Analysis of feature influence} \label{fig:02:disc}
    \end{subfigure}
  \end{minipage}
  \caption{Example classifier $\kappa_2$. The DT path
    $\langle1,3\rangle$, which is consistent with the instance
    $((1,2,2),1)$, is highlighted.} 
  \label{fig:02:clssf}
\end{figure}

\paragraph{Feature influence on predicted class.}
%~\\
Similar to the example in~\cref{ssec:ex:01}, for the concrete instance
$((1,2,2),1)$, we conclude that the value feature 1 determines the
predicted class when the prediction is 1 (see~\cref{fig:02:disc}).

\paragraph{Formal explanations \& feature relevancy.}
%~\\
The computation of formal explanations mimics the one
for~\cref{fig:01:clssf}, as shown in~\cref{tab:01:xps}. As a result,
we once again conclude that feature 1 is relevant (and necessary), and
that features 2 and 3 are irrelevant.

\paragraph{Features in adversarial examples.}
%~\\
By using the TR/DT in~\cref{fig:02:clssf}, an analysis similar to that
on~\cref{tab:01:aes} allows concluding that the only minimal $l_0$ AE
will only include feature 1, as the feature that must change value for
the prediction to change.

\begin{table}[t]
  \begin{subtable}[b]{0.4125\textwidth}
    \begin{center}
      \begin{tabular}{ccc} \toprule
        $\fml{S}$ & rows picked by $\fml{S}$ & $\phi(\fml{S})$
        \\ \toprule
        $\emptyset$ & 1..18 & $\nfrac{11}{18}$ \\
        $\{1\}$     & 10..18 & $1$ \\
        $\{2\}$     & 7..9,16..18 & $\nfrac{5}{6}$ \\
        $\{3\}$     & 3,6,9,12,15,18 & $\nfrac{5}{6}$ \\
        $\{1,2\}$   & 16..18 & $1$ \\
        $\{1,3\}$   & 12,15,18 & $1$ \\
        $\{2,3\}$   & 9,18 & $\nfrac{3}{2}$ \\
        $\{1,2,3\}$ & 18 & $1$ \\
        \bottomrule
      \end{tabular}
    \end{center}

    \begin{tabular}{c} ~\\[10pt] ~\\[10pt] ~\\[10pt] \end{tabular}
    \caption{Computing average values for the possible sets $\fml{S}$}
    \label{tab:02:svs:sets}
  \end{subtable}
  \hfill
  \begin{subtable}[b]{0.55\textwidth}
    \begin{center}
      \renewcommand{\tabcolsep}{0.45em}
      \begin{tabular}{cccccc} \toprule
        \multicolumn{6}{c}{$i=1$} \\
        \midrule
        $\fml{S}_i$ & $\phi(\fml{S}_i)$ & $\phi(\fml{S}_i\cup\{i\})$ &
        $\Delta(i,\fml{S}_i)$ & $\varsigma(\fml{S}_i)$ & $\sv(i)$ \\
        \toprule
        $\emptyset$ & $\nfrac{11}{18}$ & $1$ & $\nfrac{7}{18}$ &
        $\nfrac{1}{3}$ & -- \\
        $\{2\}$ & $\nfrac{5}{6}$ & $1$ & $\nfrac{1}{6}$ & $\nfrac{1}{6}$
        & -- \\
        $\{3\}$ & $\nfrac{5}{6}$ & $1$ & $\nfrac{1}{6}$ & $\nfrac{1}{6}$ & -- \\
        $\{2,3\}$ & $\nfrac{3}{2}$ & $1$ & -$\nfrac{1}{2}$ &
        $\nfrac{1}{3}$ & -- \\\midrule[0.25pt]
        & & & & & $\nfrac{2}{108}=0.019$ \\
        \midrule[0.7pt]
        \multicolumn{6}{c}{$i=2$} \\
        \midrule
        $\emptyset$ & $\nfrac{11}{18}$ & $\nfrac{5}{6}$ & $\nfrac{2}{9}$ &
        $\nfrac{1}{3}$ & -- \\
        $\{1\}$ & $1$ & $1$ & $0$ & $\nfrac{1}{6}$ & -- \\
        $\{3\}$ & $\nfrac{5}{6}$ & $\nfrac{3}{2}$ & $\nfrac{2}{3}$ & $\nfrac{1}{6}$ & -- \\
        $\{1,3\}$ & $1$ & $1$ & $0$ & $\nfrac{1}{3}$ & -- \\\midrule[0.25pt]
        & & & & & $\nfrac{10}{54}=0.185$ \\
        \midrule[0.7pt]
        \multicolumn{6}{c}{$i=3$} \\
        \midrule
      $\emptyset$ & $\nfrac{11}{18}$ & $\nfrac{5}{6}$ & $\nfrac{2}{9}$ &
        $\nfrac{1}{3}$ & -- \\
        $\{1\}$ & $1$ & $1$ & $0$ & $\nfrac{1}{6}$ & -- \\
        $\{2\}$ & $\nfrac{5}{6}$ & $\nfrac{3}{2}$ & $\nfrac{2}{3}$ & $\nfrac{1}{6}$ & -- \\
        $\{1,2\}$ & $1$ & $1$ & $0$ & $\nfrac{1}{3}$ & -- \\\midrule[0.25pt]
        & & & & & $-\nfrac{10}{54}=0.185$ \\
        %\midrule
        \bottomrule
      \end{tabular}
    \end{center}
    \caption{Shapley values computed using the intermediate values
      from~\cref{tab:02:svs:sets}}
    \label{tab:02:svs:tab}
  \end{subtable}
  \caption{Computation of Shapley values for the classifier
    $\kappa_2$ of~\cref{fig:02:clssf}.} \label{tab:02:svs}
\end{table}

\paragraph{Shapley values \& feature importance.}
%~\\
The computation of Shapley values for the classifier
of~\cref{fig:02:clssf}, and for the instance $((1,2,2),1)$ is shown
in~\cref{tab:02:svs}.
As can be observed, the relative order of feature importance obtained
is: 2, 3, 1 (or 3, 2, 1). The interpretation that can be made is:
feature 2 (or 3) is more important for the prediction than is feature
3 (or 2), and feature 3 (or 2) is  more important for the prediction
than is feature 1. This interpretation is in completely disagreement
with the analysis of feature influence, with the analysis of feature
relevancy, and with the analysis of adversarial examples.
The bottom line is that the features that bear no influence in
predicting class 1, are deemed the most important according to the
computed Shapley values.

\paragraph{Assessment.}
%~\\
As before in~\cref{ssec:ex:01}, we devised a classifier and an
instance for which the only relevant feature, and the feature that
bears some influence on the predicted class, is assigned an absolute
Shapley value that is smaller than the absolute Shapley values of two
other features, which are irrelevant for the prediction, and which are
clear not to influence the prediction.

\subsection{A Simple Discrete Parameterized Classifier}
\label{ssec:ex:03}

%\begin{example} \label{ex:03a}

\paragraph{Classifier.}
%~\\
Aiming to extend the conclusions of prevision sections, we now
consider a discrete classifier, defined on boolean features, with
$\mbb{D}_i=\mbb{B}=\{0,1\}$, with $1\le{i}\le{m}$.
For simplicity, we set $m=2$, and just represent the classifier with
a tabular representation as shown in~\cref{fig:03:clssf}.

\begin{figure}[ht]
  \begin{center}
    \begin{tabular}{cccc} \toprule
      $x_1$ & $x_2$ & $\kappa_3(\mbf{x})$ \\ \toprule
      0 & 0 & $\gamma$ \\
      0 & 1 & $\beta$ \\
      1 & 0 & $\delta$ \\
      1 & 1 & $\alpha$ \\
      \bottomrule
    \end{tabular}
  \end{center}
  \caption{Tabular representation for $\kappa_a$} \label{fig:03:clssf}
\end{figure}

For~\cref{fig:03:clssf}, $\mbb{F}=\mbb{B}^{2}$, and we let
$\alpha,\beta,\gamma\in\mbb{Z}$.
Moreover, we consider the instance $((1,1),\alpha)$.
It is easy to conclude that, as long as
$\alpha\not=\gamma\land\alpha\not=\beta$ and
$\delta=\alpha$, then feature 1 is relevant, and feature 2 is
irrelevant.
Given the table above, we set $\fml{K}=\{\gamma,\beta,\alpha\}$,
since we opt to pick $\delta=\alpha$. We also impose
$\gamma\not=\alpha\land\beta\not=\alpha$, as pointed out above.
As a result, throughout the remainder of this section, it will be
the case that $\delta=\alpha$, and that
$\gamma\not=\alpha\land\beta\not=\alpha$.
%
%\end{example}

\paragraph{Feature influence on predicted class.}
%~\\
For the instance $((1,1),\alpha)$, with
$\delta=\alpha,\gamma\not=\alpha,\beta\not=\alpha$, it is clear
that the predicted class is 1 if and only if feature 1 is assigned
value 1, and that the predicted class is other that 1 if and only if
feature 1 is assigned value 1.

\paragraph{Formal explanations \& feature relevancy.}
%~\\
%%\cref{fig:03:clssf} %(see~\cpageref{tab:03a})
Building on the examples in earlier sections, it is plain to conclude
that feature 1 is relevant and feature 2 is irrelevant.
%
%Alternatively, we can say that this classifiers illustrates the
%the significance of feature 1 being relevant and feature 2 being
%irrelevant. The point is
Observe that feature 2 is never necessary, neither as one of the
features required for keeping the prediction (i.e.\ included in some
AXp), nor as one of the features required for changing the prediction
(i.e.\ included in some CXp).

\paragraph{Features in adversarial examples.}
%~\\
As noted above, to change the predicted class changes if and only if
the value of feature 1 changes. No constraint is imposed on feature
2. Hence, minimal adversarial examples only require setting the value
of feature 1.

%\begin{remark}
%  Duality between AXp's and CXp's ensures that a feature is included
%  in some AXp iff it is included in some CXp. Hence, we can decide
%  relevancy by considering the set of AXp's or the set of CXp's.
%\end{remark}

\begin{table*}[t]%h
  \begin{mdframed}[linewidth=1.5pt,linecolor=darkblue,roundcorner=5pt]
  \begin{center}
    \begin{tabular}{ccccc} \toprule[1.0pt]
      Xp set $\fml{S}$ &
      $\fml{S}$ sufficient? &
      $\fml{S}$ irreducible? &
      \makecell{Pick $\fml{X}\subseteq\fml{S}$,\\ $\fml{X}$ sufficient
        \& irreducible} &
      %\lor\fml{X}=\fml{S}
      %\makecell{
      Meaning of $\fml{X}$ %\\ ... }
      relative to $\fml{S}$ \\ \toprule[1.0pt]
      $\emptyset$ & \nomark & -- & -- & -- \\ \midrule[0.25pt]
      $\{1\}$ & \yesmark & \yesmark & $\{1\}$ & $\fml{X}=\fml{S}$ is
      an AXp \\ \midrule[0.25pt]
      $\{2\}$ & \nomark & -- & -- & -- \\ \midrule[0.25pt]
      $\{1,2\}$ & \yesmark & \nomark & $\{1\}$ &
      $\begin{aligned}[c]\forall&(u_2\in\mbb{B}).\forall(\mbf{x}\in\mbb{F}).\\&[(x_1=1)\land(x_2=u_2)]\limply(\kappa_a(\mbf{x})=\alpha)\end{aligned}$\\
      \bottomrule[0.8pt]
    \end{tabular}
  \end{center}
  \caption{Explanations for example classifier in~\cref{fig:03:clssf}} \label{tab:03:xps}
  \end{mdframed}
\end{table*}

%\begin{example}

\paragraph{Shapley values \& feature importance.}
%~\\
As will be argued below, and under the stated assumptions, the goal is
for feature 1 to have a Shapley value of 0 and feature 2 to have a
non-zero Shapley value. This way, the information provided by Shapley
values is evidently misleading.
For the classifier of~\cref{fig:03:clssf}, and instance
$((1,1),\alpha)$, we can now compute the Shapley values as shown
in~\cref{tab:03:svs:tab}
(see \cite{msh-corr23} for the definitions).

\begin{table}[t] %[ht]
  \begin{center}
    \begin{tabular}{cccc} \toprule
      \multicolumn{4}{c}{$i=1$} \\ \midrule
      $\fml{S}_i$ & $\Delta(i,\fml{S}_i)$ & $\varsigma(\fml{S}_i)$ &
      $\sv(i)$ \\ \toprule
      $\emptyset$ & $\nfrac{(2\alpha-\beta-\gamma)}{4}$ & $\nfrac{1}{2}$ & -- \\[0.45pt]
      $\{2\}$ & $\nfrac{(\alpha-\beta)}{2}$ & $\nfrac{1}{2}$ & -- \\[0.45pt]
      -- & -- & -- & $\nfrac{\alpha}{2}-\nfrac{3\beta}{8}-\nfrac{\gamma}{8}$ \\[0.575pt]
      \toprule
      %\bottomrule
      %%
      \multicolumn{4}{c}{$i=2$} \\ \midrule
      $\fml{S}_i$ & $\Delta(i,\fml{S}_i)$ & $\varsigma(\fml{S}_i)$ &
      $\sv(i)$ \\ \toprule
      $\emptyset$ & $\nfrac{(\beta-\gamma)}{4}$ & $\nfrac{1}{2}$ & -- \\[0.45pt]
      $\{1\}$ & $0$ & $\nfrac{1}{2}$ & -- \\[0.45pt]
      -- & -- & -- & $\nfrac{\beta}{8}-\nfrac{\gamma}{8}$ \\[0.125pt]
      \bottomrule
    \end{tabular}
  \end{center}
  \caption{Shapley values for $\kappa_a$ and instance $((1,1),\alpha)$}
  \label{tab:03:svs:tab}
\end{table}

\paragraph{Instantiation.}
%~\\
Now, to achieve the goal of having $\sv(1)=0$ with feature 1
relevant, and $\sv(2)\not=0$ with feature 2 irrelevant, we must have
$\nfrac{\alpha}{2}-\nfrac{3\beta}{8}-\nfrac{\gamma}{8}=0$ and
$\nfrac{\beta}{8}-\nfrac{\gamma}{8}\not=0$, and the initial
constraint that $\gamma\not=\alpha$.
As an example, it is plain to conclude that
$\alpha=3,\beta=4,\gamma=0$ satisfies the constraints. Hence, we
manage to have a classifier with two features, and an example
instance such that feature 1 is relevant with a Shapley value of 0,
and feature 2 is irrelevant with a non-zero Shapley value.
Perhaps more importantly, if we pick the value of
$\alpha\in\mbb{Z}$, then it suffices to set
%$\beta=2\alpha$ and $\gamma=-2\alpha$.
$\alpha=\nfrac{3\beta}{4}+\nfrac{\gamma}{4}$ such that
$\beta\not=\gamma$.
Hence, we have arbitrarily many discrete classifiers, for each of
which a relevant feature has a Shapley value of 0, and an irrelevant
feature has a Shapley value with a non-zero value. (Therefore, this
example represents in fact a family of discrete classifiers.)
%
%%\end{example}

\paragraph{Assessment.}
%~\\
Whereas for the previous examples, the absolute Shapley value of the
relevant feature was small but non-zero, this example shows how one
can create classifiers where information provided by Shapley values
is completely misleading, i.e.\ with respect to feature 2, but also
with respect to feature 1.

%\begin{remark}
%  As in the case of~\cref{fig:01:clssf}, an adversarial example for the
%  instance $((1,1),\alpha)$ is $(0,1)$ requiring a single feature to
%  change value, and under the stated assumptions about the values of
%  $\alpha,\beta,\gamma$.
%\end{remark}

\subsection{Another Parameterized Discrete Classifier} \label{sec:ex:04}

\paragraph{Classifier.}
%~\\
%\begin{example} \label{ex:04a}
Motivated by the conclusions of the prevision section, we consider in
this section a somewhat more complex discrete classifier, defined on
boolean features, with $\mbb{D}_i=\mbb{B}=\{0,1\}$, with
$1\le{i}\le{m}$.
For simplicity, we set $m=3$, and just represent the classifier by a
tabular representation, as shown in~\cref{fig:04:clssf}.

\begin{figure}[t]%h
  \begin{center}
    \begin{tabular}{cccc} \toprule
      $x_1$ & $x_2$ & $x_3$ & $\kappa_b(\mbf{x})$ \\ \toprule
      0 & 0 & 0 & $\sigma_1$ \\
      0 & 0 & 1 & $\sigma_2$ \\
      0 & 1 & 0 & $\sigma_3$ \\
      0 & 1 & 1 & $\sigma_4$ \\
      1 & 0 & 0 & $\alpha$ \\
      1 & 0 & 1 & $\alpha$ \\
      1 & 1 & 0 & $\alpha$ \\
      1 & 1 & 1 & $\alpha$ \\
      \bottomrule
    \end{tabular}
  \end{center}
  \caption{Tabular representation for $\kappa_b$} \label{fig:04:clssf}
\end{figure}

For~\cref{fig:04:clssf}, $\mbb{F}=\mbb{B}^{3}$, and we let
$\alpha,\sigma_j\in\mbb{Z}$, with $j\in\{1,2,3,4\}$.
Moreover, we consider the instance $((1,1,1),\alpha)$.
%
%It is simple to conclude that, as long as $\sigma_i\not=\alpha$,
%$i=1,\ldots,4$, then feature 1 is relevant, and features 2 and 3 are
%irrelevant.
%
Given~\cref{fig:04:clssf}, we set
$\fml{K}=\{\alpha,\sigma_1,\ldots,\sigma_4\}$.
Thus, depending on the actual values assigned to $\alpha,\sigma_i$,
$1\le{i}\le4$, it holds that $|\fml{K}|\le5$.
We also impose $\sigma_i\not=\alpha, i=1,\ldots,4$, to ensure feature
relevancy as intended and as discussed in earlier examples.

%\end{example}

\paragraph{Feature influence on predicted class.}
%~\\
From~\cref{fig:04:clssf}, and based on the analysis of earlier
examples, it is plain to conclude that, for the instance
$((1,1,1),1)$, the predicted class is 1 if and only if feature 1 is
assigned value 1, and the predicted class is other than 1 if and only
if feature 1 is assigned value 0.
As before, features 2 and 3 bear no relevance in the predicting class
1, or in changing the predicted class 1 to something esel

\paragraph{Formal explanations \& feature relevancy.}
%~\\
In a similar way, it is immediate to conclude that, with
$\alpha\not=\alpha_j,j=1,\ldots,6$, there exists a single AXp $\{1\}$
and a single CXp $\{1\}$, which agrees with the analysis of the
influence of each feature on the predicted class 1.

\paragraph{Features in adversarial examples.}
%~\\
From~\cref{fig:04:clssf}, it is also plain that, for the parameterized
classifier of~\cref{fig:04:clssf}, any minimal $l_0$ adversarial
example must include feature 1, whereas features 2 and 3 serve no
purpose in changing the predicted class.

\begin{table}[t] %[ht]
  \begin{center}
    \renewcommand{\tabcolsep}{0.125em}
    \begin{tabular}{cccc} \toprule
      \multicolumn{4}{c}{$i=1$} \\ \midrule
      $\fml{S}_i$ & $\Delta(i,\fml{S}_i)$ & $\varsigma(\fml{S}_i)$ &
      $\sv(i)$ \\
      \toprule
      $\emptyset$ & $\nfrac{\alpha}{2}-\nfrac{(\sum\sigma_j)}{8}$ & $\nfrac{1}{3}$ & -- \\[0.45pt]
      $\{2\}$ & $\nfrac{\alpha}{2}-\nfrac{(\sigma_3+\sigma_4)}{4}$ & $\nfrac{1}{6}$ & -- \\[0.45pt]
      $\{3\}$ & $\nfrac{\alpha}{2}-\nfrac{(\sigma_2+\sigma_4)}{4}$ & $\nfrac{1}{6}$ & -- \\[0.45pt]
      $\{2,3\}$ & $\nfrac{\alpha}{2}-\nfrac{\sigma_4}{2}$ & $\nfrac{1}{3}$ & -- \\[0.45pt]
      -- & -- & -- & $\nfrac{\alpha}{2}-\nfrac{\sigma_1}{24}-\nfrac{\sigma_2}{12}-\nfrac{\sigma_3}{12}-\nfrac{7\sigma_4}{24}$ \\[0.575pt]
      \toprule
      \multicolumn{4}{c}{$i=2$} \\
      \midrule
      $\fml{S}_i$ & $\Delta(i,\fml{S}_i)$ & $\varsigma(\fml{S}_i)$ &
      $\sv(i)$ \\
      \toprule
      $\emptyset$ & $-\nfrac{(\sigma_1+\sigma_2)}{8}+\nfrac{(\sigma_3+\sigma_4)}{8}$ & $\nfrac{1}{3}$ & -- \\[0.45pt]
      $\{1\}$ & $0$ & $\nfrac{1}{6}$ & -- \\[0.45pt]
      $\{3\}$ & $-\nfrac{\sigma_2}{4}+\nfrac{\sigma_4}{4}$ & $\nfrac{1}{6}$ & -- \\[0.45pt]
      $\{1,3\}$ & $0$ & $\nfrac{1}{3}$ & -- \\[0.45pt]
      -- & -- & -- & $-\nfrac{\sigma_1}{24}-\nfrac{\sigma_2}{12}+\nfrac{\sigma_3}{24}+\nfrac{\sigma_4}{12}$ \\[0.575pt]
      \toprule
      \multicolumn{4}{c}{$i=3$} \\ \midrule
      $\fml{S}_i$ & $\Delta(i,\fml{S}_i)$ & $\varsigma(\fml{S}_i)$ &
      $\sv(i)$ \\
      \toprule
      $\emptyset$ & $-\nfrac{(\sigma_1+\sigma_3)}{8}+\nfrac{(\sigma_2+\sigma_4)}{8}$ & $\nfrac{1}{3}$ & -- \\[0.45pt]
      $\{1\}$ & $0$ & $\nfrac{1}{6}$ & -- \\[0.45pt]
      $\{2\}$ & $-\nfrac{\sigma_3}{4}+\nfrac{\sigma_4}{4}$ & $\nfrac{1}{6}$ & -- \\[0.45pt]
      $\{1,2\}$ & $0$ & $\nfrac{1}{3}$ & -- \\[0.45pt]
      -- & -- & -- & $-\nfrac{\sigma_1}{24}+\nfrac{\sigma_2}{24}-\nfrac{\sigma_3}{12}+\nfrac{\sigma_4}{12}$ \\[0.575pt]
      \bottomrule
    \end{tabular}
  \end{center}
  \caption{Computation of Shapley values for $\kappa_b$
    in~\cref{fig:04:clssf}} \label{tab:04:svs:tab}
\end{table}

\paragraph{Shapley values \& feature importance.}
%~\\
Building on the approach adopted in earlier sections, we can compute
the Shapley values for the parameterized classifier, and summarized
in~\cref{tab:04:svs:tab}.

\paragraph{Instantiations.}
%~\\
Given~\cref{tab:04:svs:tab}, and as before, our goal is to obtain
$\sv(1)=0$, with feature 1 relevant, and
$\sv(2)\not=0\land\sv(3)\not=0$, with features 2 and 3 irrelevant.
As a result, from the~\cref{tab:04:svs:tab} we get,
\[
\begin{array}{l}
  \nfrac{\alpha}{2}-\nfrac{\sigma_1}{24}-\nfrac{\sigma_2}{12}-\nfrac{\sigma_3}{12}-\nfrac{7\sigma_4}{24}
  = 0 \\[1pt]
  -\nfrac{\sigma_1}{24}-\nfrac{\sigma_2}{12}+\nfrac{\sigma_3}{24}+\nfrac{\sigma_4}{12}
  \not=0 \\[1pt]
  -\nfrac{\sigma_1}{24}+\nfrac{\sigma_2}{24}-\nfrac{\sigma_3}{12}+\nfrac{\sigma_4}{12}
  \not=0 \\
\end{array}
\]
As an example, these conditions can be satisfied by setting
$\sigma_1=\sigma_4=0$, $\sigma_2=\sigma_3=3$ and $\alpha=1$.
By plugging in these values in the expressions for the different
Shapley values, we then get $\sv(1)=0,\sv(2)=\sv(3)=-\nfrac{1}{8}$.
It is simple to make the difference in Shapley values more
significant by setting for example $\sigma_1=\sigma_4=0$,
$\sigma_2=\sigma_3=12$ and $\alpha=4$. In this case, we get
$\sv(1)=0,\sv(2)=\sv(3)=-\nfrac{1}{2}$.

\paragraph{Assessment.}
%~\\
Clearly, by suitably selecting the values of
$\sigma_1,\sigma_2,\sigma_3,\sigma_4$, we are able to find arbitrary
many examples of multi-valued classifiers defined on $\mbb{B}^{3}$,
such that $\sv(1)=0$ and $\sv(2)\not=0\land\sv(3)\not=0$, and where
feature 1 is relevant and features 2 and 3 are irrelevant.
(Therefore, and similarly to~\cref{fig:03:clssf}, this example
represents in fact a family of multi-valued classifiers.)
%
%Also important is that all classes have a non-negative value, in
%contrast with~\cref{fig:03:clssf}.
%
Finally, we should note that, although the selected instance was
$((1,1,1),\alpha)$, we could have considered other instances and/or
function definitions, as long as the computed values/classes were
changed accordingly.

\subsection{A More Complex Parameterized Discrete Example}
\label{ssec:ex:08}

\paragraph{Classifier.}
%~\\
%
This section studies a parameterized discrete classifier that
encompasses the classifier of~\cref{fig:02:clssf}. This parameterized
classifier is shown in~\cref{fig:08:clssf}.

From the table, we can conclude that $\fml{F}=\{1,2,3\}$,
$\mbb{D}_1=\{0,1\}$, $\mbb{D}_2=\mbb{D}_3=\{0,1,2\}$, and so
$\mbb{F}=\{0,1\}\times\{0,1,2\}^2$. Moreover, we also have
$\fml{K}=\{\alpha\}\cup\{\sigma_j\,|\,j=1,\ldots,9\}$.
We will require that $\alpha\not=\alpha_j,j=1,\ldots,9$;  this
constraint will be clarified below.
Finally, the target instance is $((1,2,2),\alpha)$.

\begin{figure}
  \begin{subtable}[b]{0.35\textwidth}%[t]
    \begin{center}
      \begin{tabular}{ccccc} \toprule
        row \# & $x_1$ & $x_2$ & $x_3$ & $\kappa_5(\mbf{x})$
        \\ \toprule 
        1      & 0     & 0     & 0     & $\sigma_1$ \\
        2      & 0     & 0     & 1     & $\sigma_2$ \\
        3      & 0     & 0     & 2     & $\sigma_3$ \\
        4      & 0     & 1     & 0     & $\sigma_4$ \\
        5      & 0     & 1     & 1     & $\sigma_5$ \\
        6      & 0     & 1     & 2     & $\sigma_6$ \\
        7      & 0     & 2     & 0     & $\sigma_7$ \\
        8      & 0     & 2     & 1     & $\sigma_8$ \\
        9      & 0     & 2     & 2     & $\sigma_9$ \\
        10     & 1     & 0     & 0     & $\alpha$ \\
        11     & 1     & 0     & 1     & $\alpha$ \\
        12     & 1     & 0     & 2     & $\alpha$ \\
        13     & 1     & 1     & 0     & $\alpha$ \\
        14     & 1     & 1     & 1     & $\alpha$ \\
        15     & 1     & 1     & 2     & $\alpha$ \\
        16     & 1     & 2     & 0     & $\alpha$ \\
        17     & 1     & 2     & 1     & $\alpha$ \\
        18     & 1     & 2     & 2     & $\alpha$ \\
        \bottomrule
      \end{tabular}
      \end{center}
    \caption{Tabular representation for $\kappa_{c}$}
    \label{tab:08:tt}
  \end{subtable}
  \begin{subtable}[b]{0.55\textwidth}
    \begin{center}
      \begin{tabular}{ccc} \toprule
        $\fml{S}$ & rows picked by $\fml{S}$ & $\phi(\fml{S})$  \\ \toprule
        $\emptyset$ & 1..18 & $\nfrac{(\sum_{j=1}^{9}\sigma_j)}{18}+\nfrac{\alpha}{2}$ \\
        $\{1\}$ & 10..18 & $\alpha$ \\
        $\{2\}$ & 7..9,16..18 & $\nfrac{(\sigma_7+\sigma_8+\sigma_9)}{6}+\nfrac{\alpha}{2}$ \\
        $\{3\}$ & 3,6,9,12,15,18 & $\nfrac{(\sigma_3+\sigma_6+\sigma_9)}{6}+\nfrac{\alpha}{2}$ \\
        $\{1,2\}$ & 16..18 & $\alpha$ \\
        $\{1,3\}$ & 12,15,18 & $\alpha$ \\
        $\{2,3\}$ & 9,18 & $\nfrac{\sigma_9}{2}+\nfrac{\alpha}{2}$ \\
        $\{1,2,3\}$ & 18 & $\alpha$ \\
        \bottomrule
      \end{tabular}
    \end{center}

    \begin{tabular}{c} \\[5pt] \\[5pt]\end{tabular}
    \caption{Computing $\phi(\fml{S})$, by inspecting the tabular
      representation}
    \label{tab:08:phi}
  \end{subtable}
  \caption{Example parameterized classifier $\kappa_{c}$}
  \label{fig:08:clssf}
\end{figure}

%%Instance: $((1,2,2),\alpha)$.
%%\clearpage

\paragraph{Feature influence on predicted class.}
%~\\
It is simple to conclude that the analysis applied to the previous
examples also holds in this case. Hence, for any point in feature
space, the predicted class is 1 if and only if feature 1 is assigned
value 1. The remaining features have no influence in predicting class
1 or in changing the predicted class to some other class different
from 1.

\paragraph{Formal explanations \& feature relevancy.}
%~\\
\cref{fig:08:xps} summarizes the computation of AXps and CXps for the
parameterized classifier in~\cref{tab:08:tt}.
(In this case, we opt to also highlight the computation of
contrastive explanations.)
As can be concluded, feature 1 is relevant (and necessary), whereas
features 2 and 3 are irrelevant. 

\begin{figure*}[t]
  \begin{mdframed}[linewidth=1.5pt,linecolor=darkblue,roundcorner=5pt]
    %skipabove=10pt
    %

    \medskip\smallskip
    \begin{center}
      \renewcommand{\tabcolsep}{0.5em}
      \begin{tabular}{cccc|cccc}
        \toprule[1pt]
        $\fml{S}$ &
        $\rows(\fml{S})$ &
        \makecell{$\waxp(\fml{S})$?\\$\fml{S}$ sufficient?} &
        \makecell{$\axp(\fml{S})$?\\$\fml{S}$ also minimal?} &
        $\fml{F}\setminus\fml{S}$ &
        $\rows(\fml{F}\setminus\fml{S})$ &
        \makecell{$\wcxp(\fml{S})$?\\$\fml{S}$ changes $\kappa$?} &
        \makecell{$\cxp(\fml{S})$?\\$\fml{S}$ also minimal?}
        \\
        \midrule[0.875pt]
        $\emptyset$ &
        1..12 & \nomark & &
        $\{1,2,3\}$ & 12 & \nomark & 
        \\
        $\{1\}$ &
        7,8,9,10,11,12 & \yesmark & \yesmark &
        $\{2,3\}$ & 6,12 & \yesmark & \yesmark
        \\
        $\{2\}$ &
        4,5,6,10,11,12 & \nomark & &
        $\{1,3\}$ & 9,12 & \nomark &
        \\
        $\{3\}$ &
        3,6,9,12 & \nomark & &
        $\{1,2\}$ & 10,11,12 & \nomark &
        \\
        $\{1,2\}$ & 10,11,12 & \yesmark & \nomark & 
        $\{3\}$ & 3,6,9,12 & \yesmark & \nomark
        \\
        $\{1,3\}$ & 9,12 & \yesmark & \nomark & 
        $\{2\}$ & 4,5,6,10,11,12 & \yesmark & \nomark
        \\
        $\{2,3\}$ & 6,12 & \nomark & & 
        $\{1\}$ & 7,8,9,10,11,12 & \nomark & 
        \\
        $\{1,2,3\}$ & 12 & \yesmark & \nomark & 
        $\emptyset$ & 1..12 & \yesmark & \nomark
        \\
        \bottomrule[1pt]
      \end{tabular}
    \end{center}
    \caption{Computing AXp's/CXp's for the example parameterized
      classifier shown in~\cref{fig:08:clssf} and instance
      $(\mbf{v},c)=((1,1,2),\alpha)$. All subsets of features are
      considered.
      For computing AXp's, and for some set $\fml{S}$, the features in
      $\fml{S}$ are fixed to their values as determined by $\mbf{v}$.
      The picked rows, i.e.\ $\rows(\fml{S})$, are the rows consistent
      with those fixed values.
      For example, if $\fml{S}=\{1,2\}$, then only rows 10, 11 and 12
      are consistent with having features 1 and 2 assigned value 1.
      Similarly, for computing CXp's, and for some set $\fml{S}$, the
      features in $\fml{F}\setminus\fml{S}$ are fixed to their values
      as determined by $\mbf{v}$. The picked rows are again the rows
      consistent with those fixed values. For example, if
      $\fml{S}=\{2\}$, then $\fml{F}\setminus\fml{S}=\{1,3\}$, and
      so only rows 9 and 12 are consistent with having feature 1
      assigned value 1 and feature 3 assigned value 2.
      An AXp is an irreducible set of features that is sufficient for
      the prediction. In this example, only $\{1\}$ respects the
      criteria.
      Moreover, a CXp is an irreducible set of features which, if
      allowed to take any value from their domain, the prediction
      changes value. For this example, $\{1\}$ respect the criteria,
      i.e.\ by only changing feature $\{1\}$, we are able to change
      the prediction.
    }
    \label{fig:08:xps}
  \end{mdframed}
\end{figure*}

\paragraph{Features in adversarial examples.}
~\\

\paragraph{Shapley values \& feature importance.}
%~\\
Given the average values for each possible set $\fml{S}$ shown
in~\cref{tab:08:phi}, the computation of Shapley values (for XAI) is
summarized in~\cref{fig:08:svs}.

%%\clearpage

\begin{figure*}[t]
  \begin{mdframed}[linewidth=1.5pt,linecolor=darkblue,roundcorner=5pt]
    %skipabove=10pt
    %

    \medskip\smallskip
    %
    %% Feature 1:
    \begin{subfigure}{1.0\textwidth}
      \begin{center}
        \scalebox{0.925}{
          \renewcommand{\tabcolsep}{0.45em}
          \begin{tabular}{cccccc}
            \toprule[1pt]
            $\fml{S}$ &
            $\phi(\fml{S})$ &
            $\phi(\fml{S}\cup\{1\})$ &
            $\Delta(\fml{S})$ &
            $\varsigma(\fml{S})$ &
            $\varsigma(\fml{S})\times\Delta(\fml{S})$
            \\
            \midrule[0.875pt]
            $\emptyset$ &
            $\nfrac{(\sum_{j=1}^{9}\sigma_j)}{18}+\nfrac{\alpha}{2}$ &
            $\alpha$ &
            $\nfrac{\alpha}{2}-\nfrac{(\sum_{j=1}^{9}\sigma_j)}{18}$ &
            $\sfrac{0!(3-0-1)!}{3!}=\sfrac{1}{3}$ &
            $\nfrac{\alpha}{6}-\nfrac{(\sum_{j=1}^{9}\sigma_j)}{54}$
            \\
            $\{2\}$ &
            $\nfrac{(\sigma_7+\sigma_8+\sigma_9)}{6}+\nfrac{\alpha}{2}$ &
            $\alpha$ &
            $\nfrac{\alpha}{2}-\nfrac{(\sigma_7+\sigma_8+\sigma_9)}{6}$ &
            $\sfrac{1!(3-1-1)!}{3!}=\sfrac{1}{6}$ &
            $\nfrac{\alpha}{12}-\nfrac{(\sigma_7+\sigma_8+\sigma_9)}{36}$
            \\
            $\{3\}$ &
            $\nfrac{(\sigma_3+\sigma_6+\sigma_9)}{6}+\nfrac{\alpha}{2}$ &
            $\alpha$ &
            $\nfrac{\alpha}{2}-\nfrac{(\sigma_3+\sigma_6+\sigma_9)}{6}$ &
            $\sfrac{1!(3-1-1)!}{3!}=\sfrac{1}{6}$ &
            $\nfrac{\alpha}{12}-\nfrac{(\sigma_3+\sigma_6+\sigma_9)}{36}$
            \\
            $\{2,3\}$ &
            $\nfrac{\sigma_9}{2}+\nfrac{\alpha}{2}$ &
            $\alpha$ &
            $\nfrac{\alpha}{2}-\nfrac{\sigma_9}{2}$ &
            $\sfrac{2!(3-2-1)!}{3!}=\sfrac{1}{3}$ &
            $\nfrac{\alpha}{6}-\nfrac{\sigma_9}{6}$
            \\
            \midrule[0.75pt]
            \multicolumn{5}{r}{Shapley value for feature 1 \hfill
              $\sv(1)~~=$} &
            $\nfrac{\alpha}{2}-\nfrac{(2\sigma_1+2\sigma_2+5\sigma_3+2\sigma_4+2\sigma_5+5\sigma_6+5\sigma_7+5\sigma_8+26\sigma_9)}{108}$ \\
            \bottomrule[1pt]
          \end{tabular}
        }
      \end{center}
    \end{subfigure}

    \medskip\medskip\medskip
    %
    %% Feature 2:
    \begin{subfigure}{1.0\textwidth}
      \begin{center}
        \scalebox{0.925}{
          \renewcommand{\tabcolsep}{0.275em}
          \begin{tabular}{cccccc}
            \toprule[1pt]
            $\fml{S}$ &
            $\phi(\fml{S})$ &
            $\phi(\fml{S}\cup\{2\})$ &
            $\Delta(\fml{S})$ &
            $\varsigma(\fml{S})$ &
            $\varsigma(\fml{S})\times\Delta(\fml{S})$
            \\
            \midrule[0.875pt]
            $\emptyset$ &
            $\nfrac{(\sum_{j=1}^{9}\sigma_j)}{18}+\nfrac{\alpha}{2}$ &
            $\nfrac{(\sigma_7+\sigma_8+\sigma_9)}{6}+\nfrac{\alpha}{2}$ &
            $-\nfrac{(\sum_{j=1}^{6}\sigma_j)}{18}+\nfrac{(\sigma_7+\sigma_8+\sigma_9)}{9}$ &
            $\sfrac{0!(3-0-1)!}{3!}=\sfrac{1}{3}$ &
            $-\nfrac{(\sum_{j=1}^{6}\sigma_j)}{54}+\nfrac{(\sigma_7+\sigma_8+\sigma_9)}{27}$
            \\
            $\{1\}$ &
            $\alpha$ &
            $\alpha$ &
            $0$ &
            $\sfrac{1!(3-1-1)!}{3!}=\sfrac{1}{6}$ &
            $0$
            \\
            $\{3\}$ &
            $\nfrac{(\sigma_3+\sigma_6+\sigma_9)}{6}+\nfrac{\alpha}{2}$ &
            $\nfrac{\sigma_9}{2}+\nfrac{\alpha}{2}$ &
            $-\nfrac{(\sigma_3+\sigma_6)}{6}+\nfrac{\sigma_9}{3}$ &
            $\sfrac{1!(3-1-1)!}{3!}=\sfrac{1}{6}$ &
            $-\nfrac{(\sigma_3+\sigma_6)}{36}+\nfrac{\sigma_9}{18}$
            \\
            $\{1,3\}$ &
            $\alpha$ &
            $\alpha$ &
            $0$ &
            $\sfrac{2!(3-2-1)!}{3!}=\sfrac{1}{3}$ &
            $0$
            \\
            \midrule[0.75pt]
            \multicolumn{5}{r}{Shapley value for feature 2 \hfill
              $\sv(2)~~=$} &
            $\nfrac{(-2\sum_{j=1,2,4,5}\sigma_j-5\sigma_3-5\sigma_6+4\sigma_7+4\sigma_8+10\sigma_9)}{108}$ \\
            \bottomrule[1pt]
          \end{tabular}
        }
      \end{center}
    \end{subfigure}

    \medskip\medskip\medskip
    %
    %% Feature 3:
    \begin{subfigure}{1.0\textwidth}
      \begin{center}
        \scalebox{0.925}{
          \renewcommand{\tabcolsep}{0.275em}
          \begin{tabular}{cccccc}
            \toprule[1pt]
            $\fml{S}$ &
            $\phi(\fml{S})$ &
            $\phi(\fml{S}\cup\{3\})$ &
            $\Delta(\fml{S})$ &
            $\varsigma(\fml{S})$ &
            $\varsigma(\fml{S})\times\Delta(\fml{S})$
            \\
            \midrule[0.875pt]
            $\emptyset$ &
            $\nfrac{(\sum_{j=1}^{9}\sigma_j)}{18}+\nfrac{\alpha}{2}$ &
            $\nfrac{(\sigma_3+\sigma_6+\sigma_9)}{6}+\nfrac{\alpha}{2}$ &
            $-\nfrac{(\sum_{j=1,2,4,5,7,8}\sigma_j)}{18}+\nfrac{(\sigma_3+\sigma_6+\sigma_9)}{9}$ &
            $\sfrac{0!(3-0-1)!}{3!}=\sfrac{1}{3}$ &
            $-\nfrac{(\sum_{j=1,2,4,5,7,8}\sigma_j)}{54}+\nfrac{(\sigma_3+\sigma_6+\sigma_9)}{27}$
            \\
            $\{1\}$ &
            $\alpha$ &
            $\alpha$ &
            $0$ &
            $\sfrac{1!(3-1-1)!}{3!}=\sfrac{1}{6}$ &
            $0$
            \\
            $\{2\}$ &
            $\nfrac{(\sigma_7+\sigma_8+\sigma_9)}{6}+\nfrac{\alpha}{2}$ &
            $\nfrac{\sigma_9}{2}+\nfrac{\alpha}{2}$ &
            $-\nfrac{(\sigma_7+\sigma_8)}{6}+\nfrac{\sigma_9}{3}$ &
            $\sfrac{1!(3-1-1)!}{3!}=\sfrac{1}{6}$ &
            $-\nfrac{(\sigma_7+\sigma_8)}{36}+\nfrac{\sigma_9}{18}$
            \\
            $\{1,2\}$ &
            $\alpha$ &
            $\alpha$ &
            $0$ &
            $\sfrac{2!(3-2-1)!}{3!}=\sfrac{1}{3}$ &
            $0$
            \\
            \midrule[0.75pt]
            \multicolumn{5}{r}{Shapley value for feature 3 \hfill
              $\sv(3)~~=$} &
            $\nfrac{(-2\sum_{j=1,2,4,5}\sigma_j+4\sigma_3+4\sigma_6-5\sigma_7-5\sigma_8+10\sigma_9)}{108}$ \\
            \bottomrule[1pt]
          \end{tabular}
        }
      \end{center}
    \end{subfigure}
    %
    %%\medskip\smallskip

    %\captionof{figure}{Shapley values for the example DT and instance $((0,0,0,0),0)$}
    \caption{Computation of Shapley values for the example parameterized classifier shown in~\cref{fig:08:clssf} and
      instance $((1,1,2),\alpha)$. For each feature $i$, the sets to
      consider are all the sets that do not include the feature.
      The average values are obtained by summing up the values of the
      classifier in the rows consistent with $\fml{S}$ and dividing by
      the total number of rows.
      %For $\fml{S}=\{2,4\}$, only row 3 in the truth table takes value
      %1, and so the average becomes $\sfrac{1}{4}$.
    }
    \label{fig:08:svs}
  \end{mdframed}
\end{figure*}

Given the computation of the Shapley values in~\cref{fig:08:svs}, and
the goal of have $\sv(1)=0$, $\sv(2)\not=0$ and $\sv(3)\not=0$, we
obtain the following constraints:
\begin{align}
  &\alpha=\nfrac{(2\sigma_1+2\sigma_2+5\sigma_3+2\sigma_4+2\sigma_5+5\sigma_6+5\sigma_7+5\sigma_8+26\sigma_9)}{54}
  \label{eq:08:svs:01}\\
  &\nfrac{(-2\sum_{j=1,2,4,5}\sigma_j-5\sigma_3-5\sigma_6+4\sigma_7+4\sigma_8+10\sigma_9)}{108}\not=0
  \label{eq:08:svs:02} \\
  &\nfrac{(-2\sum_{j=1,2,4,5}\sigma_j+4\sigma_3+4\sigma_6-5\sigma_7-5\sigma_8+10\sigma_9)}{108}\not=0
  \label{eq:08:svs:03}
\end{align}

Any pick of values of $\alpha$, $\sigma_j, j=1,\ldots,9$ that
satisfies the constraints above will represent a classifier where the
relative order of feature importance obtained with Shapley values is
misleading.

%%\clearpage

\begin{figure}[t]
  \begin{subtable}[b]{0.325\textwidth}%[t]
    \begin{center}
      \begin{tabular}{cccccc} \toprule
        row \# & $x_1$ & $x_2$ & $x_3$ & $\kappa_{c,1}(\mbf{x})$ & $\kappa_{c,2}(\mbf{x})$
        \\ \toprule
        1      & 0     & 0     & 0     & $0$ & $3$ \\
        2      & 0     & 0     & 1     & $2$ & $4$ \\
        3      & 0     & 0     & 2     & $0$ & $8$ \\
        4      & 0     & 1     & 0     & $0$ & $0$ \\
        5      & 0     & 1     & 1     & $5$ & $0$ \\
        6      & 0     & 1     & 2     & $0$ & $0$ \\
        7      & 0     & 2     & 0     & $0$ & $0$ \\
        8      & 0     & 2     & 1     & $8$ & $0$ \\
        9      & 0     & 2     & 2     & $0$ & $0$ \\
        10     & 1     & 0     & 0     & $1$ & $1$ \\
        11     & 1     & 0     & 1     & $1$ & $1$ \\
        12     & 1     & 0     & 2     & $1$ & $1$ \\
        13     & 1     & 1     & 0     & $1$ & $1$ \\
        14     & 1     & 1     & 1     & $1$ & $1$ \\
        15     & 1     & 1     & 2     & $1$ & $1$ \\
        16     & 1     & 2     & 0     & $1$ & $1$ \\
        17     & 1     & 2     & 1     & $1$ & $1$ \\
        18     & 1     & 2     & 2     & $1$ & $1$ \\
        \bottomrule
      \end{tabular}
    \end{center}
    \caption{Tabular representations for instantiated classifiers $\kappa_{c,1}$ and $\kappa_{c,2}$}
    \label{tab:08:tt2}
  \end{subtable}
  \hfill
  \begin{subfigure}[b]{0.2875\textwidth}
    \begin{center}
      % Concocted example
%%
%\tikzset{every label/.style={xshift=-0.35ex,
%  yshift=-5.25ex,
%  text width=1ex,
%  align=right, inner sep=1pt, font=\tiny, text=midblue}}
%%
%\tikzset{tlabel/.style={xshift=0.25ex, yshift=1.75ex, text width=1ex,
%    align=right, inner sep=1pt, font=\tiny, text=midblue}}
%%%\tikzset{every node/.style={---rectangle---}}
%
\forestset{
  BDT/.style={
    for tree={
      l=1.5cm,s sep=1.15cm,
      if n children=0{}{circle}, %rectangle
      %if n children=0{}{draw},
      draw=midblue,%draw=black,%
      text=midblue,%text=black,%
      edge={
        my edge
      },
      %if n=1{
      %  edge+={0 my edge},
      %}{},
      edge=thick,
    }
  },
}
\begin{forest}
  BDT
  [{$x_1$}, label={[yshift=-6.875ex]{{\tiny1}}} 
    [{$x_3$}, label={[yshift=-6.875ex]{{\tiny2}}}, %edge={very thick}, %top-left=x
      edge label={node[midway,left,xshift=-0.5pt] {{\scriptsize$\in\{0\}$}}}
      [{$x_2$}, label={[xshift=-3.075ex,yshift=-3.5ex]{{\tiny4}}}, %yshift=-6.875ex
        edge label={node[midway,left,xshift=-1.5pt] {{\scriptsize$\in\{1\}$}}}
        [\dghlight{\textbf{2}}, label={[yshift=-5.25ex]{{\tiny6}}},
          edge label={node[midway,left,xshift=-0.5pt] {{\scriptsize$\in\{0\}$}}}, rectangle, fill={tblue2!25} ]
        [\dghlight{\textbf{5}}, label={[yshift=-5.25ex]{{\tiny7}}},
          edge label={node[near end,right,xshift=-0.5pt] {{\scriptsize$\in\{1\}$}}}, rectangle, fill={tblue2!25} ]
        [\dghlight{\textbf{8}}, label={[yshift=-5.25ex]{{\tiny8}}},
          edge label={node[midway,right,xshift=-0.575pt] {{\scriptsize$\in\{2\}$}}}, rectangle, fill={tblue2!25} ]
      ]
      [\dghlight{\textbf{0}}, label={[yshift=-5.25ex]{{\tiny5}}},
        edge label={node[midway,right,xshift=-0.5pt] {{\scriptsize$\in\{0,2\}$}}},
        rectangle, fill={tblue2!20} ]
    ]
    [\dghlight{\textbf{1}}, label={[yshift=-5.25ex]{{\tiny3}}},
      edge={very thick, draw=purple3}, edge label={node[midway,right,xshift=0.5pt] {{\scriptsize$\in\{1\}$}}},
      rectangle, fill={tblue2!25} ]
  ]
\end{forest}
    \end{center}
    \begin{tabular}{c} \\[2pt] \\[3pt]\end{tabular}
    \caption{DT for instantiated classifier $\kappa_{c,1}$ in~\cref{tab:08:tt2}.}
    \label{tab:08:dt2}
  \end{subfigure}
  \hfill
  \begin{subfigure}[b]{0.2875\textwidth}
    \begin{center}
      % Concocted example
%%
%\tikzset{every label/.style={xshift=-0.35ex,
%  yshift=-5.25ex,
%  text width=1ex,
%  align=right, inner sep=1pt, font=\tiny, text=midblue}}
%%
%\tikzset{tlabel/.style={xshift=0.25ex, yshift=1.75ex, text width=1ex,
%    align=right, inner sep=1pt, font=\tiny, text=midblue}}
%%%\tikzset{every node/.style={---rectangle---}}
%
\forestset{
  BDT/.style={
    for tree={
      l=1.5cm,s sep=1.15cm,
      if n children=0{}{circle}, %rectangle
      %if n children=0{}{draw},
      draw=midblue,%draw=black,%
      text=midblue,%text=black,%
      edge={
        my edge
      },
      %if n=1{
      %  edge+={0 my edge},
      %}{},
      edge=thick,
    }
  },
}
\begin{forest}
  BDT
  [{$x_1$}, label={[yshift=-6.875ex]{{\tiny1}}} 
    [{$x_2$}, label={[yshift=-6.875ex]{{\tiny2}}}, %edge={very thick}, %top-left=x
      edge label={node[midway,left,xshift=-0.5pt] {{\scriptsize$\in\{0\}$}}}
      [{$x_3$}, label={[xshift=-3.075ex,yshift=-3.5ex]{{\tiny4}}}, %yshift=-6.875ex
        edge label={node[midway,left,xshift=-1.5pt] {{\scriptsize$\in\{0\}$}}}
        [\dghlight{\textbf{3}}, label={[yshift=-5.25ex]{{\tiny6}}},
          edge label={node[midway,left,xshift=-0.5pt] {{\scriptsize$\in\{0\}$}}}, rectangle, fill={tblue2!25} ]
        [\dghlight{\textbf{4}}, label={[yshift=-5.25ex]{{\tiny7}}},
          edge label={node[near end,right,xshift=-0.5pt] {{\scriptsize$\in\{1\}$}}}, rectangle, fill={tblue2!25} ]
        [\dghlight{\textbf{8}}, label={[yshift=-5.25ex]{{\tiny8}}},
          edge label={node[midway,right,xshift=-0.575pt] {{\scriptsize$\in\{2\}$}}}, rectangle, fill={tblue2!25} ]
      ]
      [\dghlight{\textbf{0}}, label={[yshift=-5.25ex]{{\tiny5}}},
        edge label={node[midway,right,xshift=-0.5pt] {{\scriptsize$\in\{1,2\}$}}},
        rectangle, fill={tblue2!20} ]
    ]
    [\dghlight{\textbf{1}}, label={[yshift=-5.25ex]{{\tiny3}}},
      edge={very thick, draw=purple3}, edge label={node[midway,right,xshift=0.5pt] {{\scriptsize$\in\{1\}$}}},
      rectangle, fill={tblue2!25} ]
  ]
\end{forest}
    \end{center}
    \begin{tabular}{c} \\[2pt] \\[3pt]\end{tabular}
    \caption{DT for instantiated classifier $\kappa_{c,2}$ in~\cref{tab:08:tt2}}
    \label{tab:08:dt3}
  \end{subfigure}
  \caption{Example DTs for instantiated classifiers, given
    parameterized classifier $\kappa_{c}$. The paths
    $\langle1,3\rangle$ in both DTs, which are consistent with the
    instance $((1,2,2),1)$, are highlighted.}
  \label{tab:08:tts} 
\end{figure}
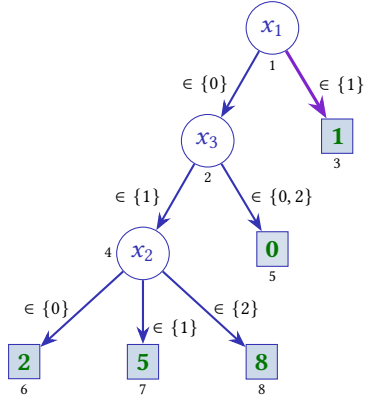
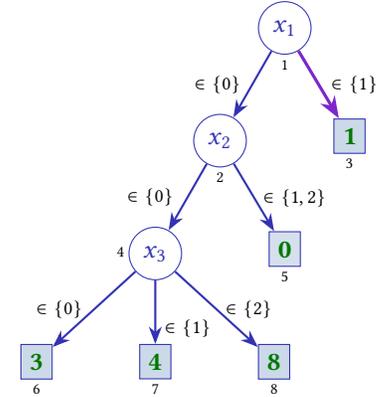

\paragraph{Instantiation.} Let us pick
$\sigma_1=\sigma_3=\sigma_4=\sigma_6=\sigma_7=\sigma_9=0$,
$\sigma_2=2$, $\sigma_5=5$ and $\sigma_8=8$, such that $\alpha=1$.
It is easy to conclude that these values
satisfy~\eqref{eq:08:svs:01},~\eqref{eq:08:svs:02},~\eqref{eq:08:svs:03}.
\cref{tab:08:tt2,tab:08:dt2} show the resulting tabular representation
and decision tree for the classifier $\kappa_{c,1}$.
In a similar way, we can pick
$\sigma_4=\sigma_5=\sigma_6=\sigma_7=\sigma_8=\sigma_9=0$,
$\sigma_1=3$, $\sigma_2=4$ and $\sigma_3=8$, such that $\alpha=1$.
It is again easy to conclude that these values
satisfy~\eqref{eq:08:svs:01},~\eqref{eq:08:svs:02},~\eqref{eq:08:svs:03}.
\cref{tab:08:tt2,tab:08:dt3} show the resulting tabular representation
and decision tree for the classifier $\kappa_{c,2}$.

\paragraph{Assessment.}
%~\\
As the instantiated examples of~\cref{tab:08:tts} illustrate, it is
simple to generate arbitrary many classifiers, given instance
$((1,2,2),\alpha$, for which only feature 1 bears some influence in
predicting class 1, only feature 1 is deemed relevant in terms of
explanations, only feature 1 occurs in adversarial examples, but such
that the computed Shapley value (for XAI) is 0, and such that the
remaining features, which bear no influence in predicting class 1,
that are irrelevant in terms of explanations, and that do not occur in
(minimal) adversarial examples, are assigned non-zero Shapley values.

\subsection{Discussion}

As the examples presented in this section reveal, it is
straightforward to devise very simple classifiers, and specific
instances, for which the computed Shapley values bear no relationship
whatsoever with the effective contribution of some features to the
predicted class.

In contrast with the contrived examples proposed in this section, the
next sections analyze published decision trees, but also OMDD
classifiers (which represent a special case of graph-based
classifiers)~\cite{hiims-kr21}.

%\section{Decision Tree Based Example Classifiers}
\section{Classifiers Defined by Decision Trees}
\label{sec:dt:exs}

This section studies two example DTs. However, in contrast with the
classifiers studied earlier in this document, the two DTs have been
studied in earlier
works~\cite{belmonte-ieee-access20,zhou2021machine}, and represent
concrete use cases.
The choice of DTs is motivated by their size, i.e.\ the DTs are not
small and so are not trivial to analyze, and by the fact that they
exhibit some of the issues with Shapley values that have been studied
in this and earlier
reports~\cite{hms-corr23a,msh-corr23,hms-corr23b}.

%(...)\\
%
%\paragraph{Binary DT.}
%~\\
%ToDo.
%
%\paragraph{Discrete DT.}
%~\\
%ToDo.

For both DTs, we investigate whether there are instances exhibiting
the following issue:
$\irrelevant(i) \land \relevant(j) \land (|\sv(i)| >|\sv(j)|)$.
For that, we use the polynomial-time algorithm for computing Shapley
values for d-DNFFs proposed in recent work~\cite{barcelo-jmlr23}.
Computation of explanations is based on earlier work as
well~\cite{hiims-kr21,iims-jair22}.
The experiments were performed on a MacBook Pro with a 6-Core Intel
Core i7 2.6 GHz processor with 16 GByte RAM, running macOS Ventura.

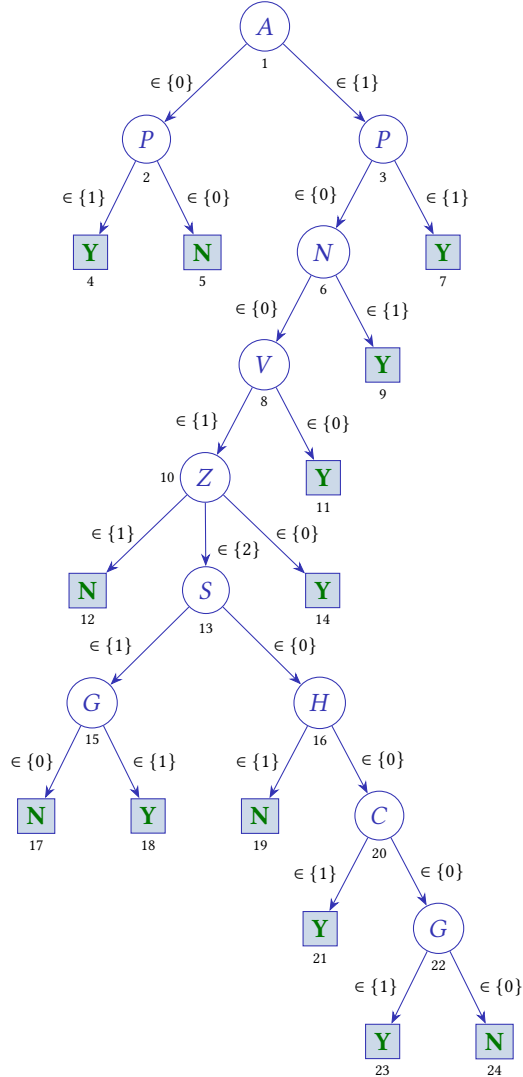
\begin{figure*}[t]
  % Example from Tanner et al. PLoS'08 paper
%%
%
\forestset{
  BDT/.style={
    for tree={
      %l=1.5cm,s sep=1.15cm,
      l=1.5cm,s sep=1.0cm,
      if n children=0{}{circle}, %rectangle
      %if n children=0{}{draw},
      draw=midblue,%draw=black,%
      text=midblue,%text=black,%
      %draw=black,%draw=midblue,
      %text=black,%text=midblue,
      edge={-{Stealth[]}},
      edge={
        my edge
      },
      %if n=1{}{
      %  edge+={0 my edge},
      %},
      %edge=thick,
      %font=\sffamily,
    }
  },
}
\begin{forest}
  BDT
  [{$A$}, label={[yshift=-6.75ex]{{\tiny1}}} %middle-middle=x
    [{$P$}, label={[yshift=-6.75ex]{{\tiny2}}}, %top-left=x
      %%edge=thick,
      edge label={node[midway,left,xshift=-0.5pt] {{\scriptsize${\in}\,\{0\}$}}}
          [\dghlight{\tbf{Y}}, label={[yshift=-5.375ex]{{\tiny4}}},
            %%edge=thick,
            edge label={node[midway,left,xshift=-0.5pt] {{\scriptsize${\in}\,\{1\}$}}},
            rectangle, fill={tblue2!25}
          ]
          [\dghlight{\tbf{N}}, label={[yshift=-5.375ex]{{\tiny5}}},
            %%edge=thick,
            edge label={node[midway,right,xshift=0.5pt] {{\scriptsize${\in}\,\{0\}$}}},
            rectangle, fill={tblue2!25}
          ]
    ]
    [{$P$}, label={[yshift=-6.75ex]{{\tiny3}}}, %top-left=x
      %%edge=thick,
      edge label={node[midway,right,xshift=0.5pt] {{\scriptsize${\in}\,\{1\}$}}}
      [{$N$}, label={[yshift=-6.885ex]{{\tiny6}}}, %top-left=x
        %%edge=thick,
        edge label={node[midway,left,xshift=-2.25pt] {{\scriptsize${\in}\,\{0\}$}}}
        [{$V$}, label={[yshift=-6.75ex]{{\tiny8}}}, %top-left=x
          %%edge=thick,
          edge label={node[midway,left,xshift=-1.75pt] {{\scriptsize${\in}\,\{0\}$}}}
          [{$Z$}, label={[xshift=-3.35ex,yshift=-3.5ex]{{\tiny10}}}, %top-left=x
            %%edge=thick,
            edge label={node[midway,left,xshift=-2.0pt] {{\scriptsize${\in}\,\{1\}$}}}
            [\dghlight{\tbf{N}}, label={[yshift=-5.375ex]{{\tiny12}}},
              %%edge=thick,
              edge label={node[midway,left,xshift=-0.5pt] {{\scriptsize${\in}\,\{1\}$}}},
              rectangle, fill={tblue2!25}
            ]
            [{$S$}, label={[yshift=-6.75ex]{{\tiny13}}}, %top-left=x
              %%edge=thick,
              edge label={node[near end,right,xshift=0.5pt] {{\scriptsize${\in}\,\{2\}$}}}
              [{$G$}, label={[yshift=-6.75ex]{{\tiny15}}}, %top-left=x
                %%edge=thick,
                edge label={node[midway,left,xshift=-2.5pt] {{\scriptsize${\in}\,\{1\}$}}}
                [\dghlight{\tbf{N}}, label={[yshift=-5.375ex]{{\tiny17}}},
                  %%edge=thick,
                  edge label={node[midway,left,xshift=0.5pt] {{\scriptsize${\in}\,\{0\}$}}},
                  rectangle, fill={tblue2!25}
                ]
                [\dghlight{\tbf{Y}}, label={[yshift=-5.375ex]{{\tiny18}}},
                  %%edge=thick,
                  edge label={node[midway,right,xshift=0.5pt] {{\scriptsize${\in}\,\{1\}$}}},
                  rectangle, fill={tblue2!25}
                ]
              ]
              [{$H$}, label={[yshift=-6.885ex]{{\tiny16}}}, %top-left=x
                %%edge=thick,
                edge label={node[midway,right,xshift=0.5pt] {{\scriptsize${\in}\,\{0\}$}}}
                [\dghlight{\tbf{N}}, label={[yshift=-5.375ex]{{\tiny19}}},
                  %%edge=thick,
                  edge label={node[midway,left,xshift=0.5pt] {{\scriptsize${\in}\,\{1\}$}}},
                  rectangle, fill={tblue2!25}
                ]
                [{$C$}, label={[yshift=-6.75ex]{{\tiny20}}}, %top-left=x
                  %%edge=thick,
                  edge label={node[midway,right,xshift=0.5pt] {{\scriptsize${\in}\,\{0\}$}}}
                  [\dghlight{\tbf{Y}}, label={[yshift=-5.375ex]{{\tiny21}}},
                    %%edge=thick,
                    edge label={node[midway,left,xshift=-0.5pt] {{\scriptsize${\in}\,\{1\}$}}},
                    rectangle, fill={tblue2!25}
                  ]
                  [{$G$}, label={[yshift=-6.75ex]{{\tiny22}}}, %top-left=x
                    %%edge=thick,
                    edge label={node[midway,right,xshift=0.5pt] {{\scriptsize${\in}\,\{0\}$}}}
                    [\dghlight{\tbf{Y}}, label={[yshift=-5.375ex]{{\tiny23}}},
                      %%edge=thick,
                      edge label={node[midway,left,xshift=-0.5pt] {{\scriptsize${\in}\,\{1\}$}}},
                      rectangle, fill={tblue2!25}
                    ]
                    [\dghlight{\tbf{N}}, label={[yshift=-5.375ex]{{\tiny24}}},
                      %%edge=thick,
                      edge label={node[midway,right,xshift=0.5pt] {{\scriptsize${\in}\,\{0\}$}}},
                      rectangle, fill={tblue2!25}
                    ]
                  ]
                ]
              ]
            ]
            [\dghlight{\tbf{Y}}, label={[yshift=-5.375ex]{{\tiny14}}},
              %%edge=thick,
              edge label={node[midway,right,xshift=0.5pt] {{\scriptsize${\in}\,\{0\}$}}},
              rectangle, fill={tblue2!25}
            ]
          ]
          [\dghlight{\tbf{Y}}, label={[yshift=-5.375ex]{{\tiny11}}},
            %%edge=thick,
            edge label={node[midway,right,xshift=0.5pt] {{\scriptsize${\in}\,\{0\}$}}},
            rectangle, fill={tblue2!25}
          ]
        ]
        [\dghlight{\tbf{Y}}, label={[yshift=-5.375ex]{{\tiny9}}},
          %%edge=thick,
          edge label={node[midway,right,xshift=0.5pt] {{\scriptsize${\in}\,\{1\}$}}},
          rectangle, fill={tblue2!25}
        ]
      ]
      [\dghlight{\tbf{Y}}, label={[yshift=-5.375ex]{{\tiny7}}},
        %%edge=thick,
        edge label={node[midway,right,xshift=0.5pt] {{\scriptsize${\in}\,\{1\}$}}},
        rectangle, fill={tblue2!25}
      ]
    ]
  ]
\end{forest}
  \caption{Example DT, adapted from~\cite{belmonte-ieee-access20}}
  \label{fig:dt03}
\end{figure*}

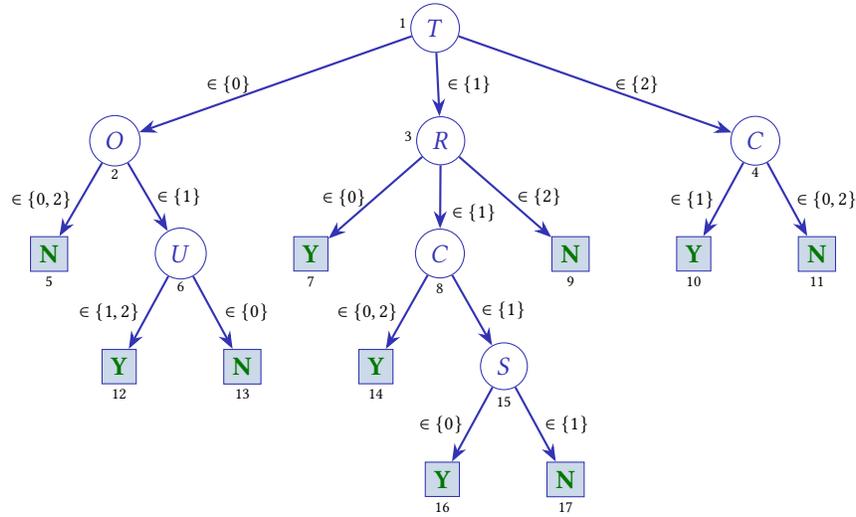
\begin{figure*}[t]
  % Example from Zhou's book, 2021.
%%
%\tikzset{every label/.style={xshift=-0.35ex,
%  yshift=-5.25ex,
%  text width=1ex,
%  align=right, inner sep=1pt, font=\tiny, text=midblue}}
%%
%\tikzset{tlabel/.style={xshift=0.25ex, yshift=1.75ex, text width=1ex,
%    align=right, inner sep=1pt, font=\tiny, text=midblue}}
%%%\tikzset{every node/.style={---rectangle---}}
%
\forestset{
  BDT/.style={
    for tree={
      l=1.5cm,s sep=1.15cm,
      if n children=0{}{circle}, %rectangle
      %if n children=0{}{draw},
      draw=midblue,%draw=black,%
      text=midblue,%text=black,%
      edge={
        my edge
      },
      %if n=1{
      %  edge+={0 my edge},
      %}{},
      edge=thick,
    }
  },
}
\begin{forest}
  BDT
  [{$T$}, label={[xshift=-2.85ex,yshift=-3.0ex]{{\tiny1}}} %,yshift=-6.875ex
    [{$O$}, label={[yshift=-6.45ex]{{\tiny2}}}, 
      edge label={node[midway,left,xshift=-5.5pt] {{\scriptsize${\in}\,\{0\}$}}}
      [\dghlight{\textbf{N}}, label={[yshift=-5.25ex]{{\tiny5}}},
        edge label={node[midway,left,xshift=0.5pt] {{\scriptsize${\in}\,\{0,2\}$}}},
        rectangle, fill={tblue2!25} ]
      [{$U$}, label={[yshift=-6.5ex]{{\tiny6}}},
        edge label={node[midway,right,xshift=-0.5pt] {{\scriptsize${\in}\,\{1\}$}}}
        [\dghlight{\textbf{Y}}, label={[yshift=-5.25ex]{{\tiny12}}},
          edge label={node[midway,left,xshift=0.5pt] {{\scriptsize${\in}\,\{1,2\}$}}}, rectangle, fill={tblue2!25} ]
        [\dghlight{\textbf{N}}, label={[yshift=-5.25ex]{{\tiny13}}},
          edge label={node[midway,right,xshift=0.5pt] {{\scriptsize${\in}\,\{0\}$}}}, rectangle, fill={tblue2!25} ]
      ]
    ]
    [{$R$}, label={[xshift=-2.85ex,yshift=-3.0ex]{{\tiny3}}},
      edge label={node[midway,right,xshift=-0.5pt] {{\scriptsize${\in}\,\{1\}$}}}
      [\dghlight{\textbf{Y}}, label={[yshift=-5.25ex]{{\tiny7}}},
        edge label={node[midway,left,xshift=0.5pt] {{\scriptsize${\in}\,\{0\}$}}},
        rectangle, fill={tblue2!25} ]
      [{$C$}, label={[yshift=-6.5ex]{{\tiny8}}},
        edge label={node[near end,right,xshift=0.5pt] {{\scriptsize${\in}\,\{1\}$}}}
        [\dghlight{\textbf{Y}}, label={[yshift=-5.25ex]{{\tiny14}}},
          edge label={node[midway,left,xshift=0.5pt] {{\scriptsize${\in}\,\{0,2\}$}}},
          rectangle, fill={tblue2!25} ]
        [{$S$}, label={[yshift=-6.5ex]{{\tiny15}}},
          edge label={node[midway,right,xshift=-0.5pt] {{\scriptsize${\in}\,\{1\}$}}}
          [\dghlight{\textbf{Y}}, label={[yshift=-5.25ex]{{\tiny16}}},
            edge label={node[midway,left,xshift=0.5pt] {{\scriptsize${\in}\,\{0\}$}}}, rectangle, fill={tblue2!25} ]
          [\dghlight{\textbf{N}}, label={[yshift=-5.25ex]{{\tiny17}}},
            edge label={node[midway,right,xshift=-0.5pt] {{\scriptsize${\in}\,\{1\}$}}}, rectangle, fill={tblue2!25} ]
        ]
      ]
      [\dghlight{\textbf{N}}, label={[yshift=-5.25ex]{{\tiny9}}},
        edge label={node[midway,right,xshift=0.5pt] {{\scriptsize${\in}\,\{2\}$}}},
        rectangle, fill={tblue2!25} ]
    ]
    [{$C$}, label={[yshift=-6.25ex]{{\tiny4}}},
      edge label={node[midway,right,xshift=3.5pt] {{\scriptsize${\in}\,\{2\}$}}}
      [\dghlight{\textbf{Y}}, label={[yshift=-5.25ex]{{\tiny10}}},
        edge label={node[midway,left,xshift=0.5pt] {{\scriptsize${\in}\,\{1\}$}}},
        rectangle, fill={tblue2!25} ]
      [\dghlight{\textbf{N}}, label={[yshift=-5.25ex]{{\tiny11}}},
        edge label={node[midway,right,xshift=-0.5pt] {{\scriptsize${\in}\,\{0,2\}$}}},
        rectangle, fill={tblue2!25} ]
    ]
  ]
\end{forest}
  \caption{Example DT, adapted from~\cite{zhou2021machine}}
  \label{fig:dt04}
\end{figure*}

\paragraph{Example Decision Trees.}
We consider two publicly available decision trees with discrete features
and classes, one adapted from~\cite[Figure~9]{belmonte-ieee-access20}
and the other from~\cite[Figure~4.8]{zhou2021machine}. The DTs are
shown in~\cref{fig:dt03,fig:dt04}. For simplicity, the DTs use set
notation for the literals, as proposed in recent
work~\cite{iims-jair22}.
\cref{tab:domlgb} shows the feature domains of the DT
in~\cref{fig:dt03}, while \cref{tab:domz} shows the feature domains of
the DT in~\cref{fig:dt04}.

\begin{table*}
\caption{Mapping of original features for the DT
from~\cite{belmonte-ieee-access20}. The original classes
$\{\tn{MD},\tn{Non-MD}\}$ are mapped to $\{\tbf{Y},\tbf{N}\}$.} \label{tab:domlgb}
    \begin{center}
      \begin{tabular}{ccccc}
        \toprule
        Feature Name & Short Name & Original Domain & Feature Number $i$ & Mapped Domain \\
        \toprule
        Age & $A$ &
        $\{ A\le5, A>5 \}$ &
        1 & $\{0, 1\}$
        \\
        Petechiae & $P$ &
        $\{\tn{no}, \tn{yes} \}$ &
        2 & $\{0,1\}$ 
        \\
        Neck Stiffness & $N$ &
        $\{\tn{no}, \tn{yes}\}$ &
        3 & $\{0,1\}$
        \\
        Vomiting & $V$ &
        $\{\tn{no}, \tn{yes}\}$ &
        4 & $\{0,1\}$
        \\
        Zone & $Z$ &
        $\{\tn{rural}, \tn{peri-urban}, \tn{urban}\}$ &
        5 & $\{0,1,2\}$
        \\
        Seizures & $S$ &
        $\{\tn{no}, \tn{yes}\}$ &
        6 & $\{0,1\}$
        \\
        Headche & $H$ &
        $\{\tn{no}, \tn{yes}\}$ &
        7 & $\{0,1\}$
        \\
        Comma & $C$ &
        $\{\tn{no}, \tn{yes}\}$ &
        8 & $\{0,1\}$
        \\
        Gender & $G$ &
        $\{\tn{female}, \tn{male}\}$ &
        9 & $\{0,1\}$
        \\
        \bottomrule
      \end{tabular}
    \end{center}
\end{table*}

%%\clearpage

\begin{table*}
\caption{Mapping of original features for the DT
from~\cite{zhou2021machine}. The original classes
$\{\tn{ripe},\tn{unripe}\}$ are mapped to
$\{\tn{Y}, \tn{N}\}$.} \label{tab:domz}
    \begin{center}
      \begin{tabular}{ccccc}
        \toprule
        Feature Name & Short Name & Original Domain & Feature Number $i$ & Mapped Domain \\
        \toprule
        Texture & $T$ &
        $\{\tn{slightly blurry}, \tn{clear}, \tn{blurry} \}$ &
        1 & $\{0, 1, 2\}$
        \\
        Root & $R$ &
        $\{\tn{curly}, \tn{slightly curly}, \tn{straight} \}$ &
        2 & $\{0,1,2\}$
        \\
        Color & $C$ &
        $\{\tn{green}, \tn{dark}, \tn{light} \}$ &
        3 & $\{0,1,2\}$
        \\
        Surface & $S$ &
        $\{\tn{hard}, \tn{soft}\}$ &
        4 & $\{0,1\}$
        \\
        Sound & $O$ &
        $\{\tn{dull}, \tn{muffled}, \tn{crisp} \}$ &
        5 & $\{0,1,2\}$ 
        \\
        Umbilicus & $U$ &
        $\{\tn{hollow}, \tn{slightly hollow}, \tn{flat} \}$ &
        6 & $\{0,1,2\}$
        \\
        \bottomrule
      \end{tabular}
    \end{center}
\end{table*}

\paragraph{Summary of results.}
\begin{figure*}
    \begin{center}
    \includegraphics[width=.675\textwidth]{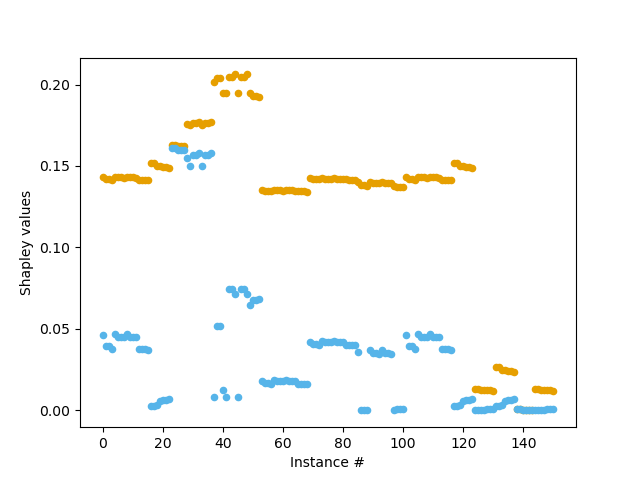}
    \caption{
    Whether there exist irrelevant features (dots in yellow) with higher scores than relevant features (dots in blue) in absolute value,
    for the DT in~\cref{fig:dt03}.}
    \label{fig:svs:res1}
    \end{center}
\end{figure*}

\begin{figure*}
    \begin{center}
    \includegraphics[width=.675\textwidth]{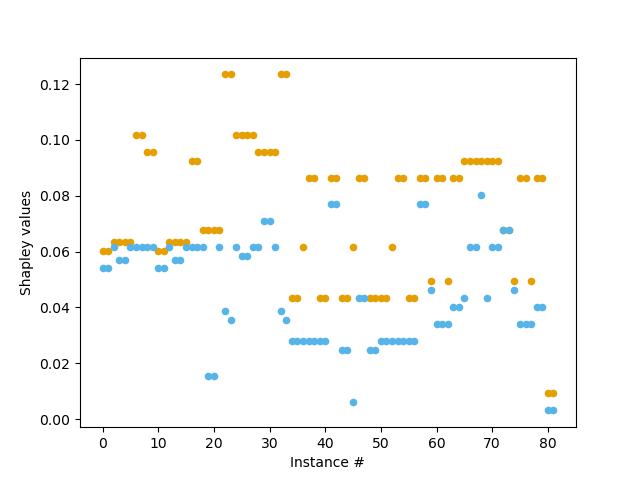}
    \caption{
    Whether there exist irrelevant features (dots in yellow) with higher scores than relevant features (dots in blue) in absolute value,
    for the DT in~\cref{fig:dt04}.}
    \label{fig:svs:res2}
    \end{center}
\end{figure*}

For each instance, all AXps are enumerated. This serves to decide
which features are relevant and which are irrelevant.
Then we compute the Shapley values for each feature and analyze
whether the issue
$\irrelevant(i) \land \relevant(j) \land (|\sv(i)| > |\sv(j)|)$
occurs. 
If an instance exhibits such an issue, we plot a pair of values
$(v_i,v_j)$. 
More specifically, $v_i = \max\{|\sv(k)|\,|\,k\not\in\fml{F}_{\mbb{A}(\fml{E})}\}$
and $v_j = \min\{|\sv(k)|\,|\,k\in\fml{F}_{\mbb{A}(\fml{E})}\}$.
(Observe that this means that the relative order of feature importance
will be misleading.)
We then plot $v_i$ in yellow and $v_j$ in blue, these pairs of values
are depicted in \cref{fig:svs:res1,fig:svs:res2}.
Another observation is the occurrence of issues with Shapley values is
non-negligible. For the DT in~\cref{fig:dt03}, 151 out of 768
instances exhibit the aforementioned issue, i.e.\ 19.7\% of the total.
Moreover, for the DT in~\cref{fig:dt04}, 82 out of 486 instances
exhibit the same issue, i.e.\ 16.8\% of the total.

Moreover, for the DT in~\cref{fig:dt03}, we found that
for the instance $((1, 0, 0, 0, 0, 0, 1, 1, 1), 1)$,
there exist two AXps: $\{1, 5\}$ and $\{1, 4\}$
and the Shapley values are: $\sv(1)=0.3572$, $\sv(2)=-0.1428$,
$\sv(3)=-0.0178$, $\sv(4)=0.0449$, $\sv(5)=0.0449$, $\sv(6)=-0.0029$,
$\sv(7)=-0.002$, $\sv(8)=0.0005$, $\sv(9)=0.0005$.
As can be concluded, for this instance, feature 2 is irrelevant and
feature 3 and 4 are relevant. However, we have $|\sv(2)| > |\sv(3)|$
and $|\sv(2)| > |\sv(4)|$.
Additionally, for the same DT, we found two instances such that
relevant features assigned with a Shapley value of 0.
Specifically, for the instance $((1, 1, 1, 0, 2, 1, 1, 0, 1), 1)$,
we can compute four AXps: $\{2\}$, $\{1, 5, 6, 7\}$, $\{1, 4\}$, and $\{1, 3\}$.
The Shapley values are: $\sv(1)=0.1172$, $\sv(2)=0.1373$,
$\sv(3)=0.0123$, $\sv(4)=0.0123$, $\sv(5)=0$, $\sv(6)=0.0016$,
$\sv(7)=0.0016$, $\sv(8)=-0.0003$, $\sv(9)=0.0004$.
Clearly, feature 5 is relevant but its Shapley value is 0.
For another instance $((1, 1, 1, 0, 2, 1, 1, 1, 0), 1)$, we can
compute four AXps: $\{2\}$, $\{1, 5, 6, 7\}$, $\{1, 4\}$, and
$\{1,3\}$. The Shapley values are: $\sv(1)=0.1172$, $\sv(2)=0.1373$,
$\sv(3)=0.0123$, $\sv(4)=0.0123$, $\sv(5)=0$, $\sv(6)=0.0016$,
$\sv(7)=0.0016$, $\sv(8)=0.0004$, $\sv(9)=-0.0003$. Clearly, for the
relevant feature 5, it has a Shapley value of 0.

\section{Classifiers Defined by OMDDs} \label{sec:mdd:exs}
In this section, we consider five publicly available datasets
and analyze whether there are instances exhibit the issue $\irrelevant(i) \land \relevant(j) \land (|\sv(i)| > |\sv(j)|)$.
These five datasets are from the Penn Machine Learning Benchmarks~\cite{Olson2017PMLB}, with discrete features and classes.
For each dataset, we picked a consistent subset of samples (i.e. no two instances are contradictory)
for building Ordered Multi-Valued Decision Diagrams (OMDDs)~\cite{brayton-tr90}.
For example, for the dataset \emph{post\-oper\-ati\-ve\_pa\-tient\_data}, there are only 88 instances, 
and a consistent subset of samples include 66 instances.
OMDD’s were built heuristically using a publicly available package 
MEDDLY~\footnote{\url{https://asminer.github.io/meddly/}}, which is implemented in C/C++.
For computing Shapley values, we assumed uniform data distribution for each dataset.
Beside, for each dataset we test randomly picked 200 instances or all instances if there are less than 200 rows in the dataset.

For all the five OMDDs, we investigate whether there are instances exhibiting the following issue:
$\irrelevant(i) \land \relevant(j) \land (|\sv(i)| >|\sv(j)|)$.
The method computing Shapley values is based on~\cref{eq:sv}.
However, it is known that OMDDs~\cite{niveau2011representing} are \emph{deterministic} and \emph{decomposable}.
Moreover, they also supports the query \emph{polytime model counting},
and the transformation \emph{polytime conditioning}~\cite{brayton-tr90,niveau2011representing}.
This means the algorithm proposed in~\cite{barcelo-jmlr23} for computing Shapley values of d-DNNFs
can be extended to the case of OMDDs.
Computation of explanations is based on earlier work as well~\cite{hiims-kr21,iims-jair22}.
The experiments were performed on a MacBook Pro with a 6-Core Intel
Core i7 2.6 GHz processor with 16 GByte RAM, running macOS Ventura.

\paragraph{Description of the datasets.}

\begin{table*}
\caption{Description of the OMDDs.} \label{tab:mdd_description}
  \begin{center}
    \begin{tabular}{ccccc}
      \toprule
      Dataset & Number of Features & Feature Domains & Number of
      Classes & Number of OMDD Nodes\\
      \toprule
      car & 6 & $4 \times 4 \times 4 \times 3 \times 3 \times 3$ & 4 & 248 \\
      monk1 & 6 & $3 \times 3 \times 2 \times 3 \times 4 \times 2$ & 2 & 68 \\
      monk2 & 6 & $3 \times 3 \times 2 \times 3 \times 4 \times 2$ & 2 & 70 \\
      monk3 & 6 & $3 \times 3 \times 2 \times 3 \times 4 \times 2$ & 2 & 74 \\
      postoperative\_patient & 8 & $3 \times 3 \times 2 \times 3 \times 2 \times 3 \times 3 \times 5$ & 2 & 109 \\
      \bottomrule
    \end{tabular}
  \end{center}
\end{table*}

\cref{tab:mdd_description} shows the description of the five OMDDs used in the experiment.

\paragraph{Summary of results.}
\begin{figure*}
    \begin{center}
    \subcaptionbox{car~\label{subfig:svs_mdds01}}{
        \includegraphics[width=.48\textwidth]{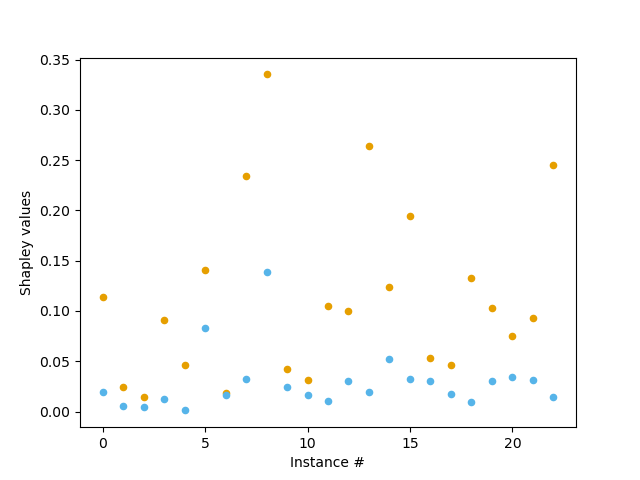}
    }
    \subcaptionbox{monk1~\label{subfig:svs_mdds02}}{
        \includegraphics[width=.48\textwidth]{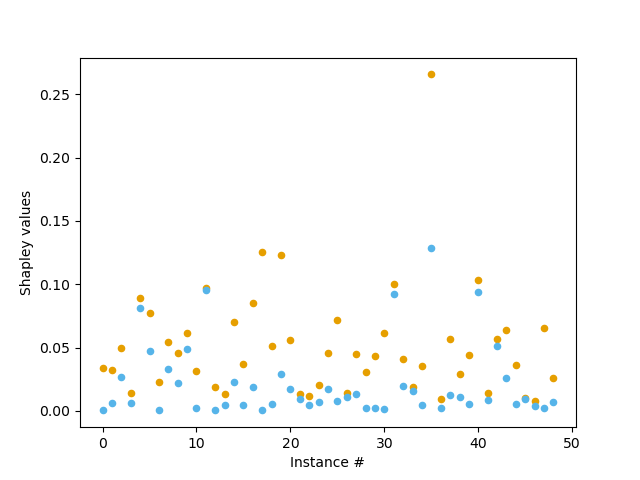}
    }
    \subcaptionbox{monk2~\label{subfig:svs_mdds03}}{
        \includegraphics[width=.48\textwidth]{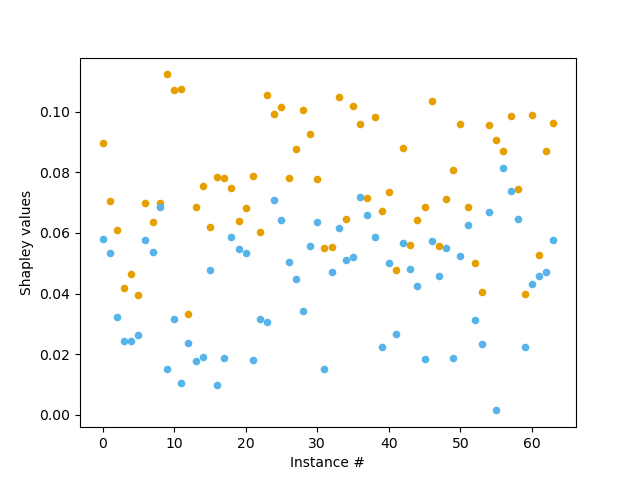}
    }
    \subcaptionbox{monk3~\label{subfig:svs_mdds04}}{
        \includegraphics[width=.48\textwidth]{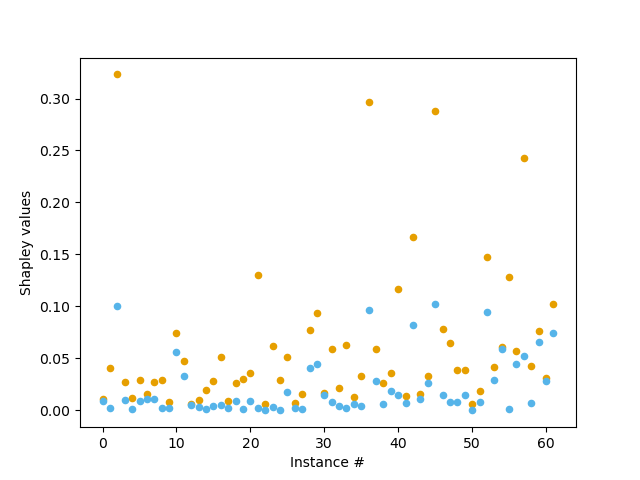}
    }
    \subcaptionbox{postoperative\_patient\_data~\label{subfig:svs_mdds05}}{
        \includegraphics[width=.48\textwidth]{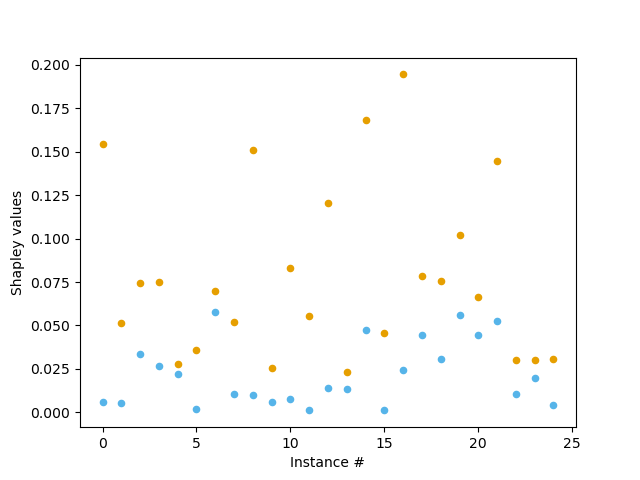}
    }
    \caption{
    Whether there exist irrelevant features (dots in yellow) with higher scores than relevant features (dots in blue) in absolute value.}
    \label{fig:svs_mdds}
    \end{center}
\end{figure*}

For the case of OMDDs, we repeat the experiment conducted in~\cref{sec:dt:exs} and plot their results.
These results are depicted in \cref{fig:svs_mdds}.

%For each instance, all AXps are enumerated. This serves to decide
%which features are relevant and which are irrelevant.
%%
%Then we compute the Shapley values for each feature and analyze
%whether the issue
%$\irrelevant(i) \land \relevant(j) \land (|\sv(i)| > |\sv(j)|)$
%occurs. 
%%
%If an instance exhibits such an issue, we plot a pair of values
%$(v_i,v_j)$. 
%More specifically, $v_i = \max\{|\sv(k)|\,|\,k\not\in\fml{F}_{\mbb{A}(\fml{E})}\}$
%and $v_j = \min\{|\sv(k)|\,|\,k\in\fml{F}_{\mbb{A}(\fml{E})}\}$.
%(Observe that this means that the relative order of feature importance
%will be misleading.)
%We then plot $v_i$ in yellow and $v_j$ in blue, these pairs of values
%are depicted in \cref{fig:svs_mdds}.

An observation is the occurrence of issues with Shapley values is
non-negligible. For the OMDD in~\ref{subfig:svs_mdds01}, 23 out of 200
instances (i.e.\ 11.5\%) exhibit the aforementioned issue.
For the OMDD in~\cref{subfig:svs_mdds02}, 49 out of 200
instances (i.e.\ 24.5\%) exhibit the same issue.
For the OMDDs in~\cref{subfig:svs_mdds03,subfig:svs_mdds04}, 64 out of 200
instances (i.e.\ 32\%) exhibit the same issue.
And for the OMDD in~\cref{subfig:svs_mdds05}, 22 out of 66
instances (i.e.\ 33.3\%) exhibit the same issue.

\section{Apparent Threats to Validity of Results \& Their Rebuttal}
\label{sec:threat}

This section addresses and rebuts a number of possible criticisms to
the results presented in this and earlier
reports~\cite{hms-corr23a,msh-corr23,hms-corr23b}\footnote{%
In fact, some of the apparent threats to validity discussed in this
section represent comments that were made with respect to earlier
reports~\cite{hms-corr23a,msh-corr23,hms-corr23b}.}.

\paragraph{Definition of (ir)relevant features.}
Our definition of (ir)relevant features mirrors the one proposed and
studied in logic-based abduction~\cite{gottlob-jacm95} since the early
and mid 90s.
(Logic-based abduction formalizes the concept of abduction, studied in
logic and philosophy for more than a century~\cite{peirce-works31}.)
Nevertheless, we explicitly consider subset-minimality for the
definition of (abductive) explanation, whereas logic-based abduction
contemplates other possible definitions~\cite{gottlob-jacm95}. For
example, there are other definitions of (minimal) explanation which
involve a user indicating some sort of preference among hypotheses (or 
features), that can involve some sort of prioritization or
penalization~\cite{gottlob-jacm95}. Since Shapley values are not
defined in terms of user-specified preferences, this sort of
preference-minimal explanations are inapplicable in our setting.
In addition, another definition of explanations involves those that
are cardinality-minimal~\cite{gottlob-jacm95}. The following is a
straightforward observation.

\begin{proposition}
  Any feature that is deemed irrelevant under a subset-minimal
  definition of explanation must also be an irrelevant feature under a
  cardinality-minimal definition of explanation.
\end{proposition}

Most of the examples in this and earlier
reports~\cite{hms-corr23a,msh-corr23,hms-corr23b} already consider a
single explanation which is necessarily cardinality-min\-imal. Hence,
replacing a subset-minimal definition of explanation by a
cardinality-minimal definition would not impact the implications of
the results presented in this and earlier
reports~\cite{hms-corr23a,msh-corr23,hms-corr23b} in terms of the
inadequacy of Shapley values for XAI.

Furthermore, the results presented in this and earlier
reports~\cite{hms-corr23a,msh-corr23,hms-corr23b} demonstrate that
Shapley values for XAI do not correlate with the information obtained
from adversarial examples. Moreover, some of results in this report
demonstrate the inadequacy of Shapley values for XAI simply by
analysis of the classifier's function.

\paragraph{Definition of Shapley values for XAI.}
Although this and earlier
reports~\cite{hms-corr23a,msh-corr23,hms-corr23b} consider a
well-established definition of Shapley values for XAI, specifically
the one proposed in a number of well-known
references~\cite{lundberg-nips17,barcelo-aaai21,vandenbroeck-aaai21,vandenbroeck-jair22,barcelo-jmlr23},
one possible criticism to the results in this and earlier
reports~\cite{hms-corr23a,msh-corr23,hms-corr23b} is that there are
other definitions of Shapley values besides the one being used.
One example is the use of
baselines~\cite{najmi-icml20,blobaum-aistats20}. Our initial
experiments suggest that the use of baselines is even more problematic
than the original definitions of Shapley for XAI. Concretely, the
percentages of detected issues for Boolean classifiers far exceed
those reported in earlier work~\cite{hms-corr23a}.
Future work will build on these initial experiments, and will document
the issues that are also observed when using Shapley values for XAI
based on baselines.

\paragraph{Evidence from practical examples.}
Since this and earlier reports
\cite{hms-corr23a,msh-corr23,hms-corr23b} study a restricted
set of example classifiers, one possible criticism is that
counterexamples to the theory of Shapley values for XAI should be
drawn from practical examples, including from those representing
complex classifiers.
The previous two sections (see~\cref{sec:dt:exs,sec:mdd:exs}) shows
results on practical DTs and OMDDs, thus confirming the existing of
issues with Shapley values in practical classifiers.
Moreover, given the complexity of computing Shapley values,
in general and for XAI in particular, it is in practice completely
unrealistic to obtain exact Shapley values in the case of the complex
classifiers used in many %(if not all)
practical applications.
Nevertheless, such evidence would be beyond the point that is being
made, in that \emph{no} sound theory can withstand a \emph{single}
counterexample. The vast number of counterexamples that this
and earlier reports~\cite{hms-corr23a,msh-corr23,hms-corr23b} have
identified already serve as comprehensive evidence to the fact that
Shapley values will necessarily provide human decision-makers with
misleading information regarding relative feature importance, for
arbitrarily many classifiers. If that were not to be the case, then
future work should identify the families of classifiers for which
Shapley values are provably guaranteed not to provide misleading
information to human decision-makers. At present, that is an open
research topic.

Furthermore, this report also includes initial experimental results, 
obtained on publicly available classifiers, that confirm that Shapley
values for XAI can produce misleading information regarding relative
feature importance.

\paragraph{Shapley values for XAI unrelated with formal explanations.}
One additional criticism to the results in this and earlier
reports \cite{hms-corr23a,msh-corr23,hms-corr23b} is that the fact
that Shapley values for XAI do not capture feature relevancy is not
problematic per se, and it might be the case that we could be talking
about different and unrelated measures of feature importance, one
provided by feature attribution and the other provided by feature
selection.
As shown in this report, we can construct classifiers with features
that are of paramount importance for a prediction, but that are
assigned a Shapley value of 0 (i.e.\ denoting no importance whatsoever
for the prediction). Similarly, we can construct classifiers (actually
the same classifier can be used!) with features that serve no purpose
in terms of explanations, and that also serve no purpose in terms of 
creating adversarial examples, but which are assigned the largest
absolute Shapley value.
In such situations, it would mystify the authors of this report if
there could exist some ascribed meaning to computed Shapley values
such that the information they convey would not be misleading for
human decision-makers.
Furthermore, existing interpretations of Shapley
values~\cite{kononenko-jmlr10} are disproved by the results presented
in this and earlier reports \cite{hms-corr23a,msh-corr23,hms-corr23b}.
Concretely, the uses of Shapley values in explainability have been
justified by very significant and very claims. For example,
from~\cite{kononenko-jmlr10}:
\begin{itemize}
\item \nohlight{\emph{``According to the 2nd axiom, if two features
  values have an identical influence on the prediction they are
  assigned contributions of equal size. The 3rd axiom says that if a
  \rhlight{\textbf{feature has no influence}} on the prediction
  \rhlight{\textbf{it is assigned a contribution of 0}}.''}}\\ 
  (Note: the axioms above refer to the axiomatic characterization of 
  Shapley values in~\cite{kononenko-jmlr10}.)
\item \nohlight{\emph{``When viewed together, these properties ensure
  that \rhlight{\textbf{any effect the features might have on the
      classifiers output will be reflected in the generated  
      contributions}}, which effectively deals with the issues of
  previous general explanation methods.''}}
\end{itemize}
Although it is the case that terms such as ``influence'' or ``effect''
are used in earlier work~\cite{kononenko-jmlr10} without a formal
definition, it is also the case that, by assuming commonly ascribed
meanings to these terms, our results prove that Shapley values for XAI
do not respect those meanings.
Thus, assuming those commonly ascribed meanings, our results disprove
the above claims. 

%%\input{res}

%\section{Discussion}
\section{Conclusions}
\label{sec:conc}

This paper significantly extends earlier evidence on the inadequacy of
Shapley values for XAI. Besides the boolean classifiers analyzed in
earlier work \cite{hms-corr23a,msh-corr23,hms-corr23b}, this paper
considers both multi-valued and discrete classifiers, exhibiting
additional examples of the issues raised by the use of Shapley values
for XAI.
Perhaps more importantly, the inadequacy of Shapley values is
also demonstrated for DTs published in recent
years~\cite{belmonte-ieee-access20,zhou2021machine}, as well as
OMDD classifiers~\cite{brayton-tr90,hiims-kr21}.

Furthermore, the paper shows that the relative order of feature
importance obtained with Shapley values for XAI does not correlate
with the features that can serve for producing $l_0$-minimal
adversarial examples, i.e.\ those that are sufficiently close to the
original instance. Thus, besides Shapley values for XAI not being
correlated with feature relevancy, it is also the case that Shapley
values for XAI do not relate with adversarial examples.

%Besides further underlining the lack of correlation between Shapley
%values and feature relevancy, the paper also analyzes the lack of
%correlation between adversarial examples and the relative order of
%feature importance obtained with Shapley values for XAI.

%%(...)\\

%\clearpage
\paragraph{Acknowledgments.}
  This work was supported by the AI Interdisciplinary Institute ANITI,
  funded by the French program ``Investing for the Future -- PIA3''
  under Grant agreement no.\ ANR-19-PI3A-0004,
  and by the H2020-ICT38 project COALA ``Cognitive Assisted agile 
  manufacturing for a Labor force supported by trustworthy Artificial
  intelligence''.
  This work was motivated in part by discussions with several colleagues
  including L.~Bertossi, A.~Ignatiev, N.~Narodytska, M.~Cooper, Y.~Izza,
  R.\ Passos, J.\ Planes and N.~Asher.
  %
  %Finally, JMS acknowledges
JMS also acknowledges
the incentive provided by the ERC who, by not funding this research
nor a handful of other grant applications between 2012 and 2022, has
had a lasting impact in framing the research presented in this paper.

\let\oldaddcontentsline\addcontentsline% Store \addcontentsline
\renewcommand{\addcontentsline}[3]{}% Make \addcontentsline a no-op

% RequiredL: \usepackage{etoolbox}
%\providetoggle{mkbbl}
\newtoggle{mkbbl}
% Contents if using bibtex: "\settoggle{mkbbl}{true}"
% Contents if inputing pre-generated file: "\settoggle{mkbbl}{false}"

\settoggle{mkbbl}{false}
 % file is automatically generated

%%\cleardoublepage %% TENTATIVE, and required if bibliography starts page...
%%\addcontentsline{toc}{section}{References}
%%\vskip 0.2in
% For arxix paper production, and since arXiv does not allow for
% bibtex, we need to create a .bbl file to include upon submission
% to arXiv.
\iftoggle{mkbbl}{
  % Run bibtex, i.e. generate .bbl gile
  %\bibliographystyle{splncs04}
  \bibliographystyle{ACM-Reference-Format}
  \bibliography{refs}
}{
  % Import bibl (original .bbl) file

}
%\bibliographystyle{abbrv}
%\bibliography{refs,xtra}
%\input{wip}
%
%\renewcommand{\addcontentsline}[3]{\oldaddcontentsline} % Restore \addcontentsline
\let\addcontentsline\oldaddcontentsline% Restore \addcontentsline

\condaddapp{
  %%\appendix
  %%\clearpage
  \begin{appendices}
    
%\section{Additional Examples}

\clearpage

\begin{figure*}[t]
  \begin{mdframed}[linewidth=1.5pt,linecolor=darkblue,roundcorner=5pt] %skipabove=10pt
    \begin{subfigure}[b]{0.29\textwidth}
      \begin{center}
        % Concocted example
%%
%\tikzset{every label/.style={xshift=-0.35ex,
%  yshift=-5.25ex,
%  text width=1ex,
%  align=right, inner sep=1pt, font=\tiny, text=midblue}}
%%
%\tikzset{tlabel/.style={xshift=0.25ex, yshift=1.75ex, text width=1ex,
%    align=right, inner sep=1pt, font=\tiny, text=midblue}}
%%%\tikzset{every node/.style={---rectangle---}}
%
\forestset{
  BDT/.style={
    for tree={
      l=1.5cm,s sep=1.15cm,
      if n children=0{}{circle}, %rectangle
      %if n children=0{}{draw},
      draw=midblue,%draw=black,%
      text=midblue,%text=black,%
      edge={
        my edge
      },
      %if n=1{
      %  edge+={0 my edge},
      %}{},
      edge=thick,
    }
  },
}
\begin{forest}
  BDT
  [{$x_1$}, label={[yshift=-6.875ex]{{\tiny1}}} 
    [{$x_3$}, label={[yshift=-6.875ex]{{\tiny2}}}, %edge={very thick}, %top-left=x
      edge label={node[midway,left,xshift=-0.5pt] {{\scriptsize$\in\{0\}$}}}
      [{$x_2$}, label={[yshift=-6.875ex]{{\tiny4}}}, %xshift=-3.075ex,yshift=-3.5ex
        edge label={node[midway,left,xshift=-1.5pt] {{\scriptsize$\in\{1\}$}}}
        [\dghlight{\textbf{4}}, label={[yshift=-5.25ex]{{\tiny6}}},
          edge label={node[midway,left,xshift=-0.5pt] {{\scriptsize$\in\{0\}$}}}, rectangle, fill={tblue2!25} ]
        [\dghlight{\textbf{7}}, label={[yshift=-5.25ex]{{\tiny7}}},
          edge label={node[midway,right,xshift=-0.575pt] {{\scriptsize$\in\{1\}$}}}, rectangle, fill={tblue2!25} ]
      ]
      [\dghlight{\textbf{0}}, label={[yshift=-5.25ex]{{\tiny5}}},
        edge label={node[midway,right,xshift=-0.5pt] {{\scriptsize$\in\{0,2\}$}}},
        rectangle, fill={tblue2!20} ]
    ]
    [\dghlight{\textbf{1}}, label={[yshift=-5.25ex]{{\tiny3}}},
      edge={very thick}, edge label={node[midway,right,xshift=0.5pt] {{\scriptsize$\in\{1\}$}}},
      rectangle, fill={tblue2!25} ]
  ]
\end{forest}
        %\begin{tabular}{c} \\[5pt] \\[5pt] \\[5pt] \end{tabular}
        %%%$\kappa(\cdot) = \ldots$
      \end{center}
      \caption{Decision tree (DT) for $\kappa_1$} \label{fig:00:dt}
    \end{subfigure}
    \begin{subfigure}[b]{0.39\textwidth}
      \begin{center}
        \begin{tabular}{ccccccc} \toprule
          row \# & $x_1$ & $x_2$ & $x_3$ & $\kappa_1(\mbf{x})$ & $\kappa_2(\mbf{x})$ &
          \\ \toprule
          1 & 0 & 0 & 0 & 0 & 0 \\
          2 & 0 & 0 & 1 & 4 & 3 \\
          3 & 0 & 0 & 2 & 0 & 0 \\
          4 & 0 & 1 & 0 & 0 & 0 \\
          5 & 0 & 1 & 1 & 7 & 2 \\
          6 & 0 & 1 & 2 & 0 & 0 \\
          7 & 1 & 0 & 0 & 1 & 1 \\
          8 & 1 & 0 & 1 & 1 & 1 \\
          9 & 1 & 0 & 2 & 1 & 1 \\
          10 & 1 & 1 & 0 & 1 & 1 \\
          11 & 1 & 1 & 1 & 1 & 1 \\
          12 & 1 & 1 & 2 & 1 & 1 \\
          \bottomrule
        \end{tabular}
      \end{center}
      \medskip %%\begin{tabular}{c} \\ \end{tabular}
      \caption{Tabular representations (TRs) for $\kappa_1$ and $\kappa_2$} \label{fig:00:tt}
    \end{subfigure}
    \begin{subfigure}[b]{0.29\textwidth}
      \begin{center}
        % Concocted example
%%
%\tikzset{every label/.style={xshift=-0.35ex,
%  yshift=-5.25ex,
%  text width=1ex,
%  align=right, inner sep=1pt, font=\tiny, text=midblue}}
%%
%\tikzset{tlabel/.style={xshift=0.25ex, yshift=1.75ex, text width=1ex,
%    align=right, inner sep=1pt, font=\tiny, text=midblue}}
%%%\tikzset{every node/.style={---rectangle---}}
%
\forestset{
  BDT/.style={
    for tree={
      l=1.5cm,s sep=1.15cm,
      if n children=0{}{circle}, %rectangle
      %if n children=0{}{draw},
      draw=midblue,%draw=black,%
      text=midblue,%text=black,%
      edge={
        my edge
      },
      %if n=1{
      %  edge+={0 my edge},
      %}{},
      edge=thick,
    }
  },
}
\begin{forest}
  BDT
  [{$x_1$}, label={[yshift=-6.875ex]{{\tiny1}}} 
    [{$x_3$}, label={[yshift=-6.875ex]{{\tiny2}}}, %edge={very thick}, %top-left=x
      edge label={node[midway,left,xshift=-0.5pt] {{\scriptsize$\in\{0\}$}}}
      [{$x_2$}, label={[yshift=-6.875ex]{{\tiny4}}}, %xshift=-3.075ex,yshift=-3.5ex
        edge label={node[midway,left,xshift=-1.5pt] {{\scriptsize$\in\{1\}$}}}
        [\dghlight{\textbf{3}}, label={[yshift=-5.25ex]{{\tiny6}}},
          edge label={node[midway,left,xshift=-0.5pt] {{\scriptsize$\in\{0\}$}}}, rectangle, fill={tblue2!25} ]
        [\dghlight{\textbf{2}}, label={[yshift=-5.25ex]{{\tiny7}}},
          edge label={node[midway,right,xshift=-0.575pt] {{\scriptsize$\in\{1\}$}}}, rectangle, fill={tblue2!25} ]
      ]
      [\dghlight{\textbf{0}}, label={[yshift=-5.25ex]{{\tiny5}}},
        edge label={node[midway,right,xshift=-0.5pt] {{\scriptsize$\in\{0,2\}$}}},
        rectangle, fill={tblue2!20} ]
    ]
    [\dghlight{\textbf{1}}, label={[yshift=-5.25ex]{{\tiny3}}},
      edge={very thick}, edge label={node[midway,right,xshift=0.5pt] {{\scriptsize$\in\{1\}$}}},
      rectangle, fill={tblue2!25} ]
  ]
\end{forest}
        %\begin{tabular}{c} \\[5pt] \\[5pt] \\[5pt] \end{tabular}
        %%%$\kappa(\cdot) = \ldots$
      \end{center}
      \caption{Decision tree (DT) for $\kappa_2$} \label{fig:00:dt2}
    \end{subfigure}

    \smallskip
    \begin{subfigure}[b]{0.975\textwidth}
      \setlength{\fboxrule}{0.875pt}
      \setlength{\fboxsep}{2.5pt}
      \fbox{
        \begin{minipage}{\textwidth}
          Analysis of the two classifiers: \\
          %DT \& truth table (TT): \\ %Instance $((1,1,2),1)$: \\
          By inspection of the DTs/TRs for both $\kappa_1$ and
          $\kappa_2$, it is immediate that: (i) if $x_1=1$, then the
          prediction must be 1; (ii) if $x_1=0$, then the prediction
          must not be 1. Thus, the classifiers predict class 1 (or
          predict a class other than 1) \emph{independently} of the
          values assigned to $x_2$ and $x_3$.
          %clear that the prediction is 1 whenever $x_1=1$,
          %\emph{independently} of the values of $x_2$ and $x_3$. %\\
          Hence, for point $\mbf{v}=(1,1,2)$, the prediction will be 1
          as long as $x_1=1$, \emph{independently} of $x_2$ and
          $x_3$.
          %These conclusions apply to both $\kappa_1$ and  $\kappa_2$.
          To change the prediction from class, the value of $x_1$ must
          be changed. In this case, the prediction will change to a
          class other than 1 \emph{independently} of the values
          assigned to $x_2$ and $x_3$. %\\
          %Thus, given the instance $((1,1,2),1)$, the value of the
          %prediction is determined \emph{uniquely} by the value of
          %$x_1$, and it is independent of the values of $x_2$ and
          %$x_3$. A change of prediction is also \emph{uniquely}
          %determined by the value of $x_1$, and it is again
          %independent of $x_2$ and $x_3$. Once again,
          These observations apply to both $\kappa_1$ and $\kappa_2$.
        \end{minipage}
      }
      
      \caption{Influence of features on $\kappa_1$ and $\kappa_2$ for
        $((1,1,2),1)$}
      \label{fig:00:disc}
    \end{subfigure}

    \caption{Example classifiers -- decision trees and their tabular
      representations.
     For these two classifiers, we have $\fml{F}=\{1,2,3\}$,
     $\mbb{D}_1=\mbb{D}_2=\{0,1\}$, and $\mbb{D}_3=\{0,1,2\}$,
     $\mbb{F}=\{0,1\}^2\times\{0,1,2\}$, and
     $\fml{K}=\{0,1,2,3,4,5,6,7\}$, albeit the DTs and TRs only use a
     subset of the classes.
     Literals in the DTs are represented with set notation, as used
     in earlier work~\cite{iims-jair22}; this solution yields more
     compact DTs.
     The classification functions are given by the decision trees, or
     alternatively by the tabular representations.
     The instance considered is $((1,1,2),1)$, which is consistent
     with path $\langle1,3\rangle$ in both DTs; this path is
     highlighted in the two DTs. The prediction is 1, as indicated in
     terminal node 3.%\\
    }
    \label{fig:00}
  \end{mdframed}
\end{figure*}

\section{Example for Journal Submission -- Deprecated}

Besides~\cref{fig:00}, the rest of the analysis mimics that
of~\cref{fig:02}.

Given the discussion in~\cref{sec:ex07}, we have that,

%\begin{align}
%  \sv(1)
%  =~&\nfrac{\alpha}{2}-\nfrac{(2\sigma_1+2\sigma_2+5\sigma_3+4\sigma_4+4\sigma_5+19\sigma_6)}{72}
%  = 0 \label{eq:00:cond01} \\
%  \sv(2) =~&\nfrac{(-2\sigma_1-2\sigma_2-5\sigma_3+2\sigma_4+2\sigma_5+5\sigma_6)}{72}
%  \not= 0 \label{eq:00:cond02} \\
%  \sv(3) =~&\nfrac{(-\sigma_1-\sigma_2+2\sigma_3-2\sigma_4-2\sigma_5+4\sigma_6)}{36}
%  \not=0 \label{eq:00:cond03} %%\\
%\end{align}
%
%Indeed, with $\alpha=1$, $\sigma_1=\sigma_2=\sigma_3=4$ and
%$\sigma_4=\sigma_5=\sigma_6=0$, we have that,
%\[
%\begin{array}{l}
%  \sv(1) = 0 \\
%  \sv(2) = -\nfrac{1}{2} \not= 0 \\
%  \sv(3) = 0 \quad \tn{(!!)}
%\end{array}
%\]

%%\begin{figure}[t]
%%  \input{./texfigs/dt00b}
%%  \caption{Similar DT, but less problematic} \label{fig:00b}
%%\end{figure}

\paragraph{Remarks.}
%~\\
Compare the DT in~\cref{fig:00:dt2} with the one in~\cref{fig:00:dt}.

\begin{table}[t]
  \begin{tabular}{cccc} \toprule
    DT & $\sv(1)$ & $\sv(2)$ & $\sv(3)$
    \\ \toprule
    \cref{fig:00:dt}   & 0 & 0.167 & -0.5 \\
    \cref{fig:00:dt2}  & 0.306 & 0.028 & -0.194 \\
    \bottomrule
  \end{tabular}
  \caption{Comparison of Shapley values}
\end{table}

%%Example 2: 
%%s1=s4=s5=s6=0 ; s2=2 ; s5=3 ; alpha=1
%%
\begin{align}
  \sv(1)
  =~&\nfrac{\alpha}{2}-\nfrac{(2\sigma_1+2\sigma_2+5\sigma_3+4\sigma_4+4\sigma_5+19\sigma_6)}{72}
  % = 1/2 - (0+4+0+0+15+0)/72
  \\
  \sv(2)
  =~&\nfrac{(-2\sigma_1-2\sigma_2-5\sigma_3+2\sigma_4+2\sigma_5+5\sigma_6)}{72} 
  % = (-0-4-0+0+6+0)/72 
  \\
  \sv(3)
  =~&\nfrac{(-\sigma_1-\sigma_2+2\sigma_3-2\sigma_4-2\sigma_5+4\sigma_6)}{36}
  % = (-0 -2 + 0 + 0 -6 + 0) / 36
\end{align}

\begin{remark}
  If only $\sigma_2$ and $\sigma_5$ are required to be assigned a
  non-zero value, then it is impossible to have $\sv(2)=\sv(3)=0$,
  even when it is clear that the prediction for $(1,1,2)$ is
  independent of the values assigned to features 2 and 3.
\end{remark}

\begin{remark}
  If we pick $\sigma_1=\sigma_3=\sigma_4=\sigma_6=\tau$,
  then if we impose either $\sigma_2\not=\tau$ or
  $\sigma_5\not=\tau$, then either $\sv(2)\not=$ or $\sv(3)\not=0$.
\end{remark}

\clearpage

\begin{figure*}[t]
  \begin{mdframed}[linewidth=1.5pt,linecolor=darkblue,roundcorner=5pt] %skipabove=10pt
    \begin{subfigure}[b]{0.325\textwidth}
      % Concocted example
%%
%\tikzset{every label/.style={xshift=-0.35ex,
%  yshift=-5.25ex,
%  text width=1ex,
%  align=right, inner sep=1pt, font=\tiny, text=midblue}}
%%
%\tikzset{tlabel/.style={xshift=0.25ex, yshift=1.75ex, text width=1ex,
%    align=right, inner sep=1pt, font=\tiny, text=midblue}}
%%%\tikzset{every node/.style={---rectangle---}}
%
\forestset{
  BDT/.style={
    for tree={
      l=1.5cm,s sep=1.15cm,
      if n children=0{}{circle}, %rectangle
      %if n children=0{}{draw},
      draw=midblue,%draw=black,%
      text=midblue,%text=black,%
      edge={
        my edge
      },
      %if n=1{
      %  edge+={0 my edge},
      %}{},
      edge=thick,
    }
  },
}
\begin{forest}
  BDT
  [{$x_1$}, label={[yshift=-6.875ex]{{\tiny1}}} 
    [{$x_2$}, label={[yshift=-6.875ex]{{\tiny2}}}, %edge={very thick}, %top-left=x
      edge label={node[midway,left,xshift=-0.5pt] {{\scriptsize$\in\{0\}$}}}
      [{$x_3$}, label={[xshift=-3.075ex,yshift=-3.5ex]{{\tiny4}}}, %yshift=-6.875ex
        edge label={node[midway,left,xshift=-1.5pt] {{\scriptsize$\in\{0\}$}}}
        [\dghlight{\textbf{0}}, label={[yshift=-5.25ex]{{\tiny6}}},
          edge label={node[midway,left,xshift=-0.5pt] {{\scriptsize$\in\{0\}$}}}, rectangle, fill={tblue2!25} ]
        [\dghlight{\textbf{3}}, label={[yshift=-5.25ex]{{\tiny7}}},
          edge label={node[near end,right,xshift=-0.575pt] {{\scriptsize$\in\{1\}$}}}, rectangle, fill={tblue2!25} ]
        [\dghlight{\textbf{6}}, label={[yshift=-5.25ex]{{\tiny8}}},
          edge label={node[midway,right,xshift=0.5pt] {{\scriptsize$\in\{2\}$}}}, rectangle, fill={tblue2!25} ]
      ]
      [\dghlight{\textbf{0}}, label={[yshift=-5.25ex]{{\tiny5}}},
        edge label={node[midway,right,xshift=-0.5pt] {{\scriptsize$\in\{1\}$}}},
        rectangle, fill={tblue2!20} ]
    ]
    [\dghlight{\textbf{1}}, label={[yshift=-5.25ex]{{\tiny3}}},
      edge={very thick}, edge label={node[midway,right,xshift=0.5pt] {{\scriptsize$\in\{1\}$}}},
      rectangle, fill={tblue2!25} ]
  ]
\end{forest}
      %%%$\kappa(\cdot) = \ldots$
      \caption{Decision tree (DT) for $\kappa$} \label{fig:01:dt}
    \end{subfigure}
    \begin{subfigure}[b]{0.295\textwidth}
      \begin{center}
        \begin{tabular}{cccccc} \toprule
          row \# & $x_1$ & $x_2$ & $x_3$ & $\kappa(\mbf{x})$
          \\ \toprule
          1 & 0 & 0 & 0 & 0 \\
          2 & 0 & 0 & 1 & 3 \\
          3 & 0 & 0 & 2 & 6 \\
          4 & 0 & 1 & 0 & 0 \\
          5 & 0 & 1 & 1 & 0 \\
          6 & 0 & 1 & 2 & 0 \\
          7 & 1 & 0 & 0 & 1 \\
          8 & 1 & 0 & 1 & 1 \\
          9 & 1 & 0 & 2 & 1 \\
          10 & 1 & 1 & 0 & 1 \\
          11 & 1 & 1 & 1 & 1 \\
          12 & 1 & 1 & 2 & 1 \\
          \bottomrule
        \end{tabular}
      \end{center}
      \medskip
      %%\begin{tabular}{c} \\ \end{tabular}
      \caption{Tabular representation (TR) for $\kappa$} \label{fig:01:tt}
    \end{subfigure}
    %\captionof{figure}{Example classifier -- decision tree} \label{fig:example01}
    %
    \begin{subfigure}[b]{0.3565\textwidth}
      \setlength{\fboxrule}{0.875pt}
      \setlength{\fboxsep}{2pt}
      \fbox{
        \begin{minipage}{\textwidth}
          Analysis of the classifier: \\
          %DT \& truth table (TT): \\ %Instance $((1,1,2),1)$: \\
          From the DT/TR, it is clear the prediction is 1 whenever
          $x_1=1$, \emph{independently} of the values of $x_2$ and
          $x_3$. %\\
          Hence, for point $\mbf{v}=(1,1,2)$, the prediction will be 1 as
          long as $x_1=1$, \emph{independently} of $x_2$ and $x_3$.
          To change the prediction, the value of $x_1$ must be
          changed. In this case, the prediction will change,
          \emph{independently} of the values of $x_2$ and $x_3$.\\
          Thus, given the instance $((1,1,2),1)$, the value of the
          prediction is determined \emph{uniquely} by the value of
          $x_1$, and it is independent of the values of $x_2$ and
          $x_3$. A change of prediction is also \emph{uniquely}
          determined by the value of $x_1$, and it is again
          independent of $x_2$ and $x_3$.
        \end{minipage}
      }
      \caption{Influence of features on $\kappa$ for $((1,1,2),1)$}
      \label{fig:01:disc}
    \end{subfigure}

    \caption{Example classifier -- decision tree and its tabular
      representation.
     For this classifier, we have $\fml{F}=\{1,2,3\}$,
     $\mbb{D}_1=\mbb{D}_2=\{0,1\}$, and $\mbb{D}_3=\{0,1,2\}$,
     $\mbb{F}=\{0,1\}^2\times\{0,1,2\}$, and
     $\fml{K}=\{0,1,2,3,4,5,6\}$, albeit the DT and TR only use 0, 1,
     3 and 6.
     Literals in the DT are represented with set notation, as proposed
     in earlier work~\cite{iims-jair22}, which yields more compact DTs.
     The classification function is given by the decision tree, or
     alternatively by the tabular representation.
     The instance considered is $((1,1,2),1)$, which is consistent
     with path $\langle1,3\rangle$; this path is highlighted in the
     DT. The prediction is 1, as indicated in terminal node 3.%\\
    }
    \label{fig:01}
  \end{mdframed}
\end{figure*}

\section{Example for Journal Submission -- Deprecated}

Besides~\cref{fig:01}, the rest of the analysis mimics that
of~\cref{fig:02}.

\clearpage

\begin{figure*}[t]
  \begin{mdframed}[linewidth=1.5pt,linecolor=darkblue,roundcorner=5pt] %skipabove=10pt
    \begin{subfigure}[b]{0.325\textwidth}
      % Example from Rudin et al. NeurIPS'19 paper
%%
%\tikzset{every label/.style={xshift=-0.35ex,
%  yshift=-5.25ex,
%  text width=1ex,
%  align=right, inner sep=1pt, font=\tiny, text=midblue}}
%%
%\tikzset{tlabel/.style={xshift=0.25ex, yshift=1.75ex, text width=1ex,
%    align=right, inner sep=1pt, font=\tiny, text=midblue}}
%%%\tikzset{every node/.style={---rectangle---}}
%
\forestset{
  BDT/.style={
    for tree={
      l=1.5cm,s sep=1.15cm,
      if n children=0{}{circle}, %rectangle
      %if n children=0{}{draw},
      draw=midblue,%draw=black,%
      text=midblue,%text=black,%
      edge={
        my edge
      },
      %if n=1{
      %  edge+={0 my edge},
      %}{},
      edge=thick,
    }
  },
}
%
% middle-middle=x : x_1
% top-left=x : x_2
% bottom-right=x : x_3
% bottom-left=x : x_4
% top-right=x : x_5
%
\begin{forest}
  BDT
  [{$x_1$}, label={[yshift=-6.875ex]{{\tiny1}}} %middle-middle=x
    [{$x_2$}, label={[yshift=-6.875ex]{{\tiny2}}}, %edge={very thick}, %top-left=x
      edge label={node[midway,left,xshift=-0.5pt] {{\scriptsize$\in\{0\}$}}}
      [{$x_3$}, label={[yshift=-6.875ex]{{\tiny4}}}, %edge={very thick}, %bottom-right=x
        edge label={node[midway,left,xshift=-2.5pt] {{\scriptsize$\in\{0\}$}}}
        [\dghlight{\textbf{2}}, label={[yshift=-5.25ex]{{\tiny6}}}, %edge={very thick}, 
          edge label={node[midway,left,xshift=-1.5pt] {{\scriptsize$\in\{0\}$}}},
          rectangle, fill={tblue2!25} ]
        [\dghlight{\textbf{0}}, label={[yshift=-5.25ex]{{\tiny7}}},
          edge label={node[midway,right,xshift=0.5pt] {{\scriptsize$\in\{1,2\}$}}},
          rectangle, fill={tblue2!25} ]
      ]
      [{$x_3$}, label={[yshift=-6.875ex]{{\tiny5}}}, %bottom-left=x
        edge label={node[midway,right,xshift=1.5pt] {{\scriptsize$\in\{1\}$}}}
        [\dghlight{\textbf{4}}, label={[yshift=-5.25ex]{{\tiny8}}},
          edge label={node[midway,left,xshift=-0.5pt] {{\scriptsize$\in\{0,1\}$}}},
          rectangle, fill={tblue2!20} ]
        [\dghlight{\textbf{0}}, label={[yshift=-5.25ex]{{\tiny9}}},
          edge label={node[midway,right,xshift=0.5pt] {{\scriptsize$\in\{2\}$}}},
          rectangle, fill={tblue2!25} ]
      ]
    ]
    [\dghlight{\textbf{1}}, label={[yshift=-5.25ex]{{\tiny3}}},
      edge={very thick}, edge label={node[midway,right,xshift=0.5pt] {{\scriptsize$\in\{1\}$}}},
      rectangle, fill={tblue2!25} ]
  ]
\end{forest}
      %%%$\kappa(\cdot) = \ldots$
      \caption{Decision tree (DT) for $\kappa$} \label{fig:02:dt}
    \end{subfigure}
    \begin{subfigure}[b]{0.295\textwidth}
      \begin{center}
        \begin{tabular}{cccccc} \toprule
          row \# & $x_1$ & $x_2$ & $x_3$ & $\kappa(\mbf{x})$
          \\ \toprule
          1 & 0 & 0 & 0 & 2 \\
          2 & 0 & 0 & 1 & 0 \\
          3 & 0 & 0 & 2 & 0 \\
          4 & 0 & 1 & 0 & 4 \\
          5 & 0 & 1 & 1 & 4 \\
          6 & 0 & 1 & 2 & 0 \\
          7 & 1 & 0 & 0 & 1 \\
          8 & 1 & 0 & 1 & 1 \\
          9 & 1 & 0 & 2 & 1 \\
          10 & 1 & 1 & 0 & 1 \\
          11 & 1 & 1 & 1 & 1 \\
          12 & 1 & 1 & 2 & 1 \\
          \bottomrule
        \end{tabular}
      \end{center}
      \medskip %\begin{tabular}{c} \\ \end{tabular}
      \caption{Tabular representation (TR) for $\kappa$} \label{fig:02:tt}
    \end{subfigure}
    %\captionof{figure}{Example classifier -- decision tree} \label{fig:example02}
    %
    \begin{subfigure}[b]{0.3565\textwidth}
      \setlength{\fboxrule}{0.875pt}
      \setlength{\fboxsep}{2pt}
      \fbox{
        \begin{minipage}{\textwidth}
          Analysis of the classifier: \\
          %DT \& truth table (TT): \\ %Instance $((1,1,2),1)$: \\
          From the DT/TR, it is clear the prediction is 1 whenever
          $x_1=1$, \emph{independently} of the values of $x_2$ and
          $x_3$. %\\
          Hence, for point $\mbf{v}=(1,1,2)$, the prediction will be 1 as
          long as $x_1=1$, \emph{independently} of $x_2$ and $x_3$.
          To change the prediction, the value of $x_1$ must be
          changed. In this case, the prediction will change,
          \emph{independently} of the values of $x_2$ and $x_3$.\\
          Thus, given the instance $((1,1,2),1)$, the value of the
          prediction is determined \emph{uniquely} by the value of
          $x_1$, and it is independent of the values of $x_2$ and
          $x_3$. A change of prediction is also \emph{uniquely}
          determined by the value of $x_1$, and it is again
          independent of $x_2$ and $x_3$.
        \end{minipage}
      }
      \caption{Influence of features on $\kappa$ for $((1,1,2),1)$}
      \label{fig:02:disc}
    \end{subfigure}

    \caption{Example classifier -- decision tree and its tabular
      representation.
     For this classifier, we have $\fml{F}=\{1,2,3\}$,
     $\mbb{D}_1=\mbb{D}_2=\{0,1\}$, and $\mbb{D}_3=\{0,1,2\}$,
     $\mbb{F}=\{0,1\}^2\times\{0,1,2\}$, and
     $\fml{K}=\{0,1,2,3,4\}$.
     Literals in the DT are represented with set notation, as proposed
     in earlier work~\cite{iims-jair22}, which yields more compact DTs.
     The classification function is given by the decision tree, or
     alternatively by the tabular representation.
     The instance considered is $((1,1,2),1)$, which is consistent
     with path $\langle1,3\rangle$; this path is highlighted in the
     DT. The prediction is 1, as indicated in terminal node 3.%\\
    }
    \label{fig:02}
  \end{mdframed}
\end{figure*}

%\begin{figure*}
%  \input{./texfigs/dt02}
%\end{figure*}

\section{Example for CACM Journal Submission -- Deprecated} \label{sec:ex07}

This section studies the classifier shown in~\cref{fig:02}.
The decision tree (DT) for the classifier is depicted
in~\cref{fig:02:dt}, and the corresponding tabular representation (TR)
of the classifier is presented in~\cref{fig:02:tt}.
The instance considered is $(\mbf{v},c)=((1,1,2),1)$.
A brief analysis of the influence of features on the classifier's
behavior is presented in~\cref{fig:02:disc}.
This analysis is based solely on inspecting the function computed by
the classifier, as shown either in the DT or the TR.

\paragraph{An hypothetical scenario.}
%~\\
To motivate the analysis of the classifier in~\cref{fig:02}, we
consider the following hypothetical scenario\footnote{%
It would be fairly straightforward to create a dataset, e.g.\ the DT
itself, from which the DT shown in~\cref{fig:02:dt} would be induced
using existing tree learning tools.}.
A small college aims to predict the number of extra-curricular
activities of each of its students, where this number can be between 0
and 4. Let feature 1 represent whether the student is a honors student
(0 for no, and 1 for yes). Let feature 2 represent where the student
originates from a urban/non-urban household (0 for non-urban, and 1
for urban). Finally, let feature 3 represent whether the student's
field of study is humanities, arts or sciences (0 for humanities, 1
for arts and 2 for sciences). Thus, the target instance $((1,1,2),1)$
denotes a honors student from an urban household, studying sciences,
for whom the predicted number of extra-curricular activities is 1.

\begin{table}[t]
  \begin{tabular}{ccccc} \toprule
    row \# & $x_1$ & $x_2$ & $x_3$ & $\kappa_5(\mbf{x})$ \\ \toprule
    1      & 0     & 0     & 0     & $\sigma_1$ \\
    2      & 0     & 0     & 1     & $\sigma_2$ \\
    3      & 0     & 0     & 2     & $\sigma_3$ \\
    4      & 0     & 1     & 0     & $\sigma_4$ \\
    5      & 0     & 1     & 1     & $\sigma_5$ \\
    6      & 0     & 1     & 2     & $\sigma_6$ \\
    7      & 1     & 0     & 0     & $\alpha$ \\
    8      & 1     & 0     & 1     & $\alpha$ \\
    9      & 1     & 0     & 2     & $\alpha$ \\
    10     & 1     & 1     & 0     & $\alpha$ \\
    11     & 1     & 1     & 1     & $\alpha$ \\
    12     & 1     & 1     & 2     & $\alpha$ \\
    \bottomrule
  \end{tabular}
  \caption{Tabular representation for $\kappa_5$} \label{tab:05:tt}
\end{table}

\paragraph{Parameterized example.}
%~\\
Throughout the remainder of this section, we will consider a more
general classifier, shown in~\cref{tab:05:tt}. As clarified below, we
impose that,
\begin{equation} \label{eq:05:cond00}
  \alpha\not=\sigma_i, i=1\ldots,6
\end{equation}
For simplicity, we also require that
$\alpha,\sigma_i\in\mbb{Z}^{+}_0, i=1,\ldots,6$.
It is plain that the DT of~\cref{fig:02} is a concrete
instantiation of the parameterized classifier shown
in~\cref{tab:05:tt}, by setting $\sigma_1=2$,
$\sigma_2=\sigma_3=\sigma_6=0$, $\sigma_4=\sigma_5=4$ and $\alpha=1$.
For the parameterized example, the instance that will be considered is
$((1,1,2),\alpha)$.

\paragraph{Formal explanations.}
%~\\
For both the classifier of~\cref{fig:02} or the parameterized
classifier of~\cref{tab:05:tt} (under the stated assumptions), it is
simple to show that $\mbb{A} = \{ \{1\} \}$ and that
$\mbb{C} = \{ \{1\} \}$. The computation of AXp's/CXp's is shown
in~\cref{fig:xps}. (Observe that computation of AXp's/CXp's shown
holds as long as $\alpha$ \emph{differs} from each of the $\sigma_i$,
$i=1,\ldots,6$, i.e.~\eqref{eq:05:cond00} holds.)
Thus, for any of these classifiers, feature 1 is relevant (in fact it
is necessary), and features 2 and 3 are irrelevant.
%These results agree with the analysis of the
%classifier (see~\cref{fig:02}) in terms of feature influence, in that
%feature 1 occurs in explanations, and features 2 and 3 do not.

\begin{figure*}[t]
  \begin{mdframed}[linewidth=1.5pt,linecolor=darkblue,roundcorner=5pt]
    %skipabove=10pt
    %

    \medskip\smallskip
    \begin{center}
      \renewcommand{\tabcolsep}{0.5em}
      \begin{tabular}{cccc|cccc}
        \toprule[1pt]
        $\fml{S}$ &
        $\rows(\fml{S})$ &
        \makecell{$\waxp(\fml{S})$?\\$\fml{S}$ sufficient?} &
        \makecell{$\axp(\fml{S})$?\\$\fml{S}$ also minimal?} &
        $\fml{F}\setminus\fml{S}$ &
        $\rows(\fml{F}\setminus\fml{S})$ &
        \makecell{$\wcxp(\fml{S})$?\\$\fml{S}$ changes $\kappa$?} &
        \makecell{$\cxp(\fml{S})$?\\$\fml{S}$ also minimal?}
        \\
        \midrule[0.875pt]
        $\emptyset$ &
        1..12 & \nomark & &
        $\{1,2,3\}$ & 12 & \nomark & 
        \\
        $\{1\}$ &
        7,8,9,10,11,12 & \yesmark & \yesmark &
        $\{2,3\}$ & 6,12 & \yesmark & \yesmark
        \\
        $\{2\}$ &
        4,5,6,10,11,12 & \nomark & &
        $\{1,3\}$ & 9,12 & \nomark &
        \\
        $\{3\}$ &
        3,6,9,12 & \nomark & &
        $\{1,2\}$ & 10,11,12 & \nomark &
        \\
        $\{1,2\}$ & 10,11,12 & \yesmark & \nomark & 
        $\{3\}$ & 3,6,9,12 & \yesmark & \nomark
        \\
        $\{1,3\}$ & 9,12 & \yesmark & \nomark & 
        $\{2\}$ & 4,5,6,10,11,12 & \yesmark & \nomark
        \\
        $\{2,3\}$ & 6,12 & \nomark & & 
        $\{1\}$ & 7,8,9,10,11,12 & \nomark & 
        \\
        $\{1,2,3\}$ & 12 & \yesmark & \nomark & 
        $\emptyset$ & 1..12 & \yesmark & \nomark
        \\
        \bottomrule[1pt]
      \end{tabular}
    \end{center}
    \caption{Computing AXp's/CXp's for the example parameterized
      classifier shown in~\cref{tab:05:tt} and instance
      $(\mbf{v},c)=((1,1,2),\alpha)$. All subsets of features are
      considered.
      For computing AXp's, and for some set $\fml{S}$, the features in
      $\fml{S}$ are fixed to their values as determined by $\mbf{v}$.
      The picked rows, i.e.\ $\rows(\fml{S})$, are the rows consistent
      with those fixed values.
      For example, if $\fml{S}=\{1,2\}$, then only rows 10, 11 and 12
      are consistent with having features 1 and 2 assigned value 1.
      Similarly, for computing CXp's, and for some set $\fml{S}$, the
      features in $\fml{F}\setminus\fml{S}$ are fixed to their values
      as determined by $\mbf{v}$. The picked rows are again the rows
      consistent with those fixed values. For example, if
      $\fml{S}=\{2\}$, then $\fml{F}\setminus\fml{S}=\{1,3\}$, and
      so only rows 9 and 12 are consistent with having feature 1
      assigned value 1 and feature 3 assigned value 2.
      An AXp is an irreducible set of features that is sufficient for
      the prediction. In this example, only $\{1\}$ respects the
      criteria.
      Moreover, a CXp is an irreducible set of features which, if
      allowed to take any value from their domain, the prediction
      changes value. For this example, $\{1\}$ respect the criteria,
      i.e.\ by only changing feature $\{1\}$, we are able to change
      the prediction.
    }
    \label{fig:xps}
  \end{mdframed}
\end{figure*}

As can be concluded, the computed abductive and contrastive
explanations agree with the analysis shown in~\cref{fig:02:disc} in
terms of feature influence. Indeed, features 2 and 3, which have no
influence in determining nor in changing the prediction, are not
included in the computed explanations. In contrast, feature 1, which
is solely responsible for the prediction, is included in the computed
explanations.

\paragraph{Feature attribution with Shapley values.}
%~\\
We will consider one of the most popular XAI approaches, which
consists in attributing relative feature importance by exploiting
Shapley
values~\cite{conklin-asmbi01,kononenko-jmlr10,kononenko-kis14,lundberg-nips17,barcelo-aaai21,vandenbroeck-aaai21,vandenbroeck-jair22,barcelo-jmlr23}.
To compute the Shapley values, it will be convenient to evaluate the
value of $\phi$ (see~\eqref{eq:phi}), i.e.\ the average value of the
classifier when the features of a given set $\fml{S}$ are fixed, for
each of the possible sets $\fml{S}$ we will need.  This is shown
in~\cref{tab:05:phi}.

\begin{table}[t]
  \begin{tabular}{ccc} \toprule
    $\fml{S}$ & rows picked by $\fml{S}$ & $\phi(\fml{S})$  \\ \toprule
    $\emptyset$ & 1..12 & $\nfrac{(\sum_{j=1}^{6}\sigma_j)}{12}+\nfrac{\alpha}{2}$ \\
    $\{1\}$ & 7..12 & $\alpha$ \\
    $\{2\}$ & 4..6,10..12 & $\nfrac{(\sigma_4+\sigma_5+\sigma_6)}{6}+\nfrac{\alpha}{2}$ \\
    $\{3\}$ & 3,6,9,12 & $\nfrac{(\sigma_3+\sigma_6)}{4}+\nfrac{\alpha}{2}$ \\
    $\{1,2\}$ & 10..12 & $\alpha$ \\
    $\{1,3\}$ & 9,12 & $\alpha$ \\
    $\{2,3\}$ & 6,12 & $\nfrac{\sigma_6}{2}+\nfrac{\alpha}{2}$ \\
    $\{1,2,3\}$ & 12 & $\alpha$ \\
    \bottomrule
  \end{tabular}
  \caption{Computing $\phi(\fml{S})$, by inspecting the tabular
    representation}
  \label{tab:05:phi}
\end{table}

\begin{figure*}[t]
  \begin{mdframed}[linewidth=1.5pt,linecolor=darkblue,roundcorner=5pt]
    %skipabove=10pt
    %

    \medskip\smallskip
    %
    %% Feature 1:
    \begin{subfigure}{1.0\textwidth}
      \begin{center}
        \renewcommand{\tabcolsep}{0.45em}
        \begin{tabular}{cccccc}
          \toprule[1pt]
          $\fml{S}$ &
          $\phi(\fml{S})$ &
          $\phi(\fml{S}\cup\{1\})$ &
          $\Delta(\fml{S})$ &
          $\varsigma(\fml{S})$ &
          $\varsigma(\fml{S})\times\Delta(\fml{S})$
          \\
          \midrule[0.875pt]
          $\emptyset$ &
          $\nfrac{(\sum_{j=1}^{6}\sigma_j)}{12}+\nfrac{\alpha}{2}$ &
          $\alpha$ &
          $\nfrac{\alpha}{2}-\nfrac{(\sum_{j=1}^{6}\sigma_j)}{12}$ &
          $\sfrac{0!(3-0-1)!}{3!}=\sfrac{1}{3}$ &
          $\nfrac{\alpha}{6}-\nfrac{(\sum_{j=1}^{6}\sigma_j)}{36}$
          \\
          $\{2\}$ &
          $\nfrac{(\sigma_4+\sigma_5+\sigma_6)}{6}+\nfrac{\alpha}{2}$ &
          $\alpha$ &
          $\nfrac{\alpha}{2}-\nfrac{(\sigma_4+\sigma_5+\sigma_6)}{6}$ &
          $\sfrac{1!(3-1-1)!}{3!}=\sfrac{1}{6}$ &
          $\nfrac{\alpha}{12}-\nfrac{(\sigma_4+\sigma_5+\sigma_6)}{36}$
          \\
          $\{3\}$ &
          $\nfrac{(\sigma_3+\sigma_6)}{4}+\nfrac{\alpha}{2}$ &
          $\alpha$ &
          $\nfrac{\alpha}{2}-\nfrac{(\sigma_3+\sigma_6)}{4}$ &
          $\sfrac{1!(3-1-1)!}{3!}=\sfrac{1}{6}$ &
          $\nfrac{\alpha}{12}-\nfrac{(\sigma_3+\sigma_6)}{24}$
          \\
          $\{2,3\}$ &
          $\nfrac{\sigma_6}{2}+\nfrac{\alpha}{2}$ &
          $\alpha$ &
          $\nfrac{\alpha}{2}-\nfrac{\sigma_6}{2}$ &
          $\sfrac{2!(3-2-1)!}{3!}=\sfrac{1}{3}$ &
          $\nfrac{\alpha}{6}-\nfrac{\sigma_6}{6}$
          \\
          \midrule[0.75pt]
          \multicolumn{5}{r}{Shapley value for feature 1 \hfill
            $\sv(1)~~=$} &
          $\nfrac{\alpha}{2}-\nfrac{(2\sigma_1+2\sigma_2+5\sigma_3+4\sigma_4+4\sigma_5+19\sigma_6)}{72}$ \\
          \bottomrule[1pt]
        \end{tabular}
      \end{center}
    \end{subfigure}
    %%\captionof{table}{Shapley value for feature 1} \label{tab:sv1}

    \medskip\medskip\medskip
    %
    %% Feature 2:
    \begin{subfigure}{1.0\textwidth}
      \begin{center}
        \renewcommand{\tabcolsep}{0.5em}
        \begin{tabular}{cccccc}
          \toprule[1pt]
          $\fml{S}$ &
          $\phi(\fml{S})$ &
          $\phi(\fml{S}\cup\{2\})$ &
          $\Delta(\fml{S})$ &
          $\varsigma(\fml{S})$ &
          $\varsigma(\fml{S})\times\Delta(\fml{S})$
          \\
          \midrule[0.875pt]
          $\emptyset$ &
          $\nfrac{(\sum_{j=1}^{6}\sigma_j)}{12}+\nfrac{\alpha}{2}$ &
          $\nfrac{(\sigma_4+\sigma_5+\sigma_6)}{6}+\nfrac{\alpha}{2}$ &
          $-\nfrac{(\sigma_1+\sigma_2+\sigma_3)}{12}+\nfrac{(\sigma_4+\sigma_5+\sigma_6)}{12}$ &
          $\sfrac{0!(3-0-1)!}{3!}=\sfrac{1}{3}$ &
          $-\nfrac{(\sigma_1+\sigma_2+\sigma_3)}{36}+\nfrac{(\sigma_4+\sigma_5+\sigma_6)}{36}$
          \\
          $\{1\}$ &
          $\alpha$ &
          $\alpha$ &
          $0$ &
          $\sfrac{1!(3-1-1)!}{3!}=\sfrac{1}{6}$ &
          $0$
          \\
          $\{3\}$ &
          $\nfrac{(\sigma_3+\sigma_6)}{4}+\nfrac{\alpha}{2}$ &
          $\nfrac{\sigma_6}{2}+\nfrac{\alpha}{2}$ &
          $-\nfrac{\sigma_3}{4}+\nfrac{\sigma_6}{4}$ &
          $\sfrac{1!(3-1-1)!}{3!}=\sfrac{1}{6}$ &
          $-\nfrac{\sigma_3}{24}+\nfrac{\sigma_6}{24}$
          \\
          $\{1,3\}$ &
          $\alpha$ &
          $\alpha$ &
          $0$ &
          $\sfrac{2!(3-2-1)!}{3!}=\sfrac{1}{3}$ &
          $0$
          \\
          \midrule[0.75pt]
          \multicolumn{5}{r}{Shapley value for feature 2 \hfill
            $\sv(2)~~=$} &
          $\nfrac{(-2\sigma_1-2\sigma_2-5\sigma_3+2\sigma_4+2\sigma_5+5\sigma_6)}{72}$ \\
          \bottomrule[1pt]
        \end{tabular}
      \end{center}
    \end{subfigure}
    %%\captionof{table}{Shapley value for feature 2} \label{tab:sv2}

    \medskip\medskip\medskip
    %
    %% Feature 3:
    \begin{subfigure}{1.0\textwidth}
      \begin{center}
        \renewcommand{\tabcolsep}{0.5em}
        \begin{tabular}{cccccc}
          \toprule[1pt]
          $\fml{S}$ &
          $\phi(\fml{S})$ &
          $\phi(\fml{S}\cup\{3\})$ &
          $\Delta(\fml{S})$ &
          $\varsigma(\fml{S})$ &
          $\varsigma(\fml{S})\times\Delta(\fml{S})$
          \\
          \midrule[0.875pt]
          $\emptyset$ &
          $\nfrac{(\sum_{j=1}^{6}\sigma_j)}{12}+\nfrac{\alpha}{2}$ &
          $\nfrac{(\sigma_3+\sigma_6)}{4}+\nfrac{\alpha}{2}$ &
          $-\nfrac{(\sigma_1+\sigma_2+\sigma_4+\sigma_5)}{12}+\nfrac{(\sigma_3+\sigma_6)}{6}$ &
          $\sfrac{0!(3-0-1)!}{3!}=\sfrac{1}{3}$ &
          $-\nfrac{(\sigma_1+\sigma_2+\sigma_4+\sigma_5)}{36}+\nfrac{(\sigma_3+\sigma_6)}{18}$
          \\
          $\{1\}$ &
          $\alpha$ &
          $\alpha$ &
          $0$ &
          $\sfrac{1!(3-1-1)!}{3!}=\sfrac{1}{6}$ &
          $0$
          \\
          $\{2\}$ &
          $\nfrac{(\sigma_4+\sigma_5+\sigma_6)}{6}+\nfrac{\alpha}{2}$ &
          $\nfrac{\sigma_6}{2}+\nfrac{\alpha}{2}$ &
          $-\nfrac{(\sigma_4+\sigma_5)}{6}+\nfrac{\sigma_6}{3}$ &
          $\sfrac{1!(3-1-1)!}{3!}=\sfrac{1}{6}$ &
          $-\nfrac{(\sigma_4+\sigma_5)}{36}+\nfrac{\sigma_6}{18}$
          \\
          $\{1,2\}$ &
          $\alpha$ &
          $\alpha$ &
          $0$ &
          $\sfrac{2!(3-2-1)!}{3!}=\sfrac{1}{3}$ &
          $0$
          \\
          \midrule[0.75pt]
          \multicolumn{5}{r}{Shapley value for feature 3 \hfill
            $\sv(3)~~=$} &
          $\nfrac{(-\sigma_1-\sigma_2+2\sigma_3-2\sigma_4-2\sigma_5+4\sigma_6)}{36}$ \\
          \bottomrule[1pt]
        \end{tabular}
      \end{center}
    \end{subfigure}
    
    %
    %%\medskip\smallskip

    %\captionof{figure}{Shapley values for the example DT and instance $((0,0,0,0),0)$}
    \caption{Computation of Shapley values for the example DT and
      instance $((1,1,2),\alpha)$. For each feature $i$, the sets to
      consider are all the sets that do not include the feature.
      The average values are obtained by summing up the values of the
      classifier in the rows consistent with $\fml{S}$ and dividing by
      the total number of rows.
      %For $\fml{S}=\{2,4\}$, only row 3 in the truth table takes value
      %1, and so the average becomes $\sfrac{1}{4}$.
    }
    \label{fig:05:svs}
  \end{mdframed}
\end{figure*}

Next, given the values obtained in~\cref{tab:05:phi}, we can compute
the Shapley value of each feature as shown in~\cref{fig:05:svs}.
The computed Shapley values are symbolic, and depend on the values of
the predicted classes.
We now demonstrate that there is surprising flexibility in choosing
the influence of each feature. Concretely, we want to show that
Shapley values yield misleading information.
%
%It is clear (as shown in Table[ToDo]), that feature 1 is relevant (and
%necessary), and that features 2 and 3 are irrelevant.
%
As a result, given what we already know about feature 1, 2 and 3 with
respect to the instance $((1,1,2),1)$, our goal is to devise
classifiers such that $\sv(1)=0$, $\sv(2)\not=0$ and $\sv(3)\not=0$.
This goal is justified in that these Shapley values disagree both with
the argued influence of features (see~\eqref{fig:02:disc}) and with
feature relevancy, as obtained from computed formal explanations.
Given the computation of Shapley values in~\cref{fig:05:svs}, we
obtain the following constraints.
\begin{align}
  \sv(1) =~&\nfrac{\alpha}{2}-\nfrac{(2\sigma_1+2\sigma_2+5\sigma_3+4\sigma_4+4\sigma_5+19\sigma_6)}{72}
  = 0 \label{eq:05:cond01} \\
  \sv(2) =~&\nfrac{(-2\sigma_1-2\sigma_2-5\sigma_3+2\sigma_4+2\sigma_5+5\sigma_6)}{72}
  \not= 0 \label{eq:05:cond02} \\
  \sv(3) =~&\nfrac{(-\sigma_1-\sigma_2+2\sigma_3-2\sigma_4-2\sigma_5+4\sigma_6)}{36}
  \not=0 \label{eq:05:cond03} %%\\
\end{align}

Thus, we can express $\alpha$ in terms of $\sigma_1\ldots,\sigma_6$,
i.e.
\begin{equation} \label{eq:def:alpha}
  \alpha=\nfrac{(2\sigma_1+2\sigma_2+5\sigma_3+4\sigma_4+4\sigma_5+19\sigma_6)}{36}
\end{equation}
In addition, we can identify values for $\sigma_i$, $i=1,\ldots,6$,
such that the two remaining conditions \eqref{eq:05:cond02}
and~\eqref{eq:05:cond03} are satisfied.
For example, we can assign the values of $\alpha$ and $\sigma_i$,
$i=1,\ldots,6$, as shown in~\cref{fig:02:tt} and~\cref{fig:02:dt}.
Concretely, we pick $\sigma_1=2,
\sigma_4=\sigma_5=4,\sigma_2=\sigma_3=\sigma_6=0$.
%
%As a result, from~\eqref{eq:05:cond01},~\eqref{eq:05:cond02}
%and~\eqref{eq:05:cond03} we get,
%%
%\[
%%\begin{array}{l}
%\alpha = 1 \land %\\
%%-4 +0 +0 +8 +8 +0 
%{+12} \not= 0 \land %\\
%%-2 +0 +0 -8 -8 +0
%{-18} \not= 0 %%\\
%%\end{array}
%\]
%%
%
From~\eqref{eq:def:alpha}, we get $\alpha=1$, and so $\sv(1)=0$.
Moreover, the selection of values for $\alpha_i,i=1,\ldots,6$ ensures
that $\sv(2)=\nfrac{12}{72}=\nfrac{1}{6}\not=0$ and 
$\sv(3)=-\nfrac{18}{36}=-\nfrac{1}{2}\not=0$.
%
%
%Observe that the choice of values for $\sigma_i$ and $\alpha$
%corresponds to the DT in~\cref{fig:02}.
%
Given the discussion in~\cref{fig:02:disc}, we conclude that feature
1, which determines the prediction, and which must be modified to
change the prediction, is assigned a Shapley value of 0, which is
supposed to signify~\cite{kononenko-jmlr10} that it has \emph{no}
influence on the prediction.
Similarly, features 2 and 3, which were shown in~\cref{fig:02:disc}
to bear \emph{no} influence in determining or changing the prediction,
are assigned non-zero Shapley values, which is supposed to
signify~\cite{kononenko-jmlr10} \emph{some} influence on the
prediction.
Thus, as the example demonstrates, the computed Shapley values
rank \emph{incorrectly} the features in terms of their influence in
determining or changing the prediction.

\condaddval{
  \paragraph{Validation.}
  %~\\
  To validate the results shown above, we get:
  \[\begin{array}{l}
  \sum_{j=1}^{6} \sigma_j = 2 + 0 + 0 + 4 + 4 + 0 = 10 \\
  \kappa(\mbf{v}) = \kappa(1,1,2) = \alpha = 1 \\
  \phi(\emptyset) = \nfrac{10}{12} + \nfrac{1}{2} = \nfrac{4}{3}\\
  \sum_i \sv(i) = 0 + \nfrac{1}{6} - \nfrac{1}{2} = -\nfrac{1}{3}\\
  \end{array}
  \]
  Thus, we should have,
  \[
  \begin{array}{rcc}
    -\nfrac{1}{3} + \nfrac{4}{3} = 1 & \Leftrightarrow & \\
    \nfrac{3}{3} = 1 & ~~ & \tn{\yesmark}\\
  \end{array}
  \]
}

\paragraph{Generalization.}
%~\\
Since the example is parameterized in $\alpha$, $\sigma_i$,
$i=1,\ldots,6$, we can define arbitrary many classifiers for which the
limitations of Shapley values for XAI are revealed.
For example, it is plain to conclude that by setting $\alpha=\psi$ and 
$\sigma_1=2\psi,\sigma_2=\sigma_3=\sigma_6=0,\sigma_4=\sigma_5=4\psi$, 
e.g.\ with $\psi\in\mbb{Z}^{+}$, we obtain a classifier where feature
1 is relevant (and necessary), features 2 and 3 are irrelevant, and
such that $\sv(1)=0$, $\sv(2)\not=0$ and $\sv(3)\not=0$.

Furthermore, other combinations of values can be selected as long as
conditions \eqref{eq:05:cond00}, \eqref{eq:05:cond01},
\eqref{eq:05:cond02} and \eqref{eq:05:cond03} hold true. Each such
combination of values captures an arbitrary number of classifiers, as
illustrated above.

\paragraph{Consequences for the hypothetical scenario.}
%~\\
Although from a formal perspective, the proposed example suffices to
demonstrate the inadequacy of Shapley values for XAI, it can be more
intuitive to discuss the hypothetical scenario outlined at the outset
of this section.
As argued earlier, the instance $((1,1,2),1)$ denotes a honors student
from an urban household that is studying sciences. Formal
explanations, either abductive or contrastive, consist of feature 1
(honors student).
In contrast, in terms of Shapley values for XAI, feature 1 would be
assigned no importance, albeit it determines the prediction, and
features 2  (type of household) and 3  (field of study) would be
assigned some importance, albeit none of these features has any
influence on the prediction.
Given the information provided by Shapley values, a decision maker
would have to deem features 2 and 3 as influencing the prediction, and
feature 1 and not influencing the prediction. However, as clearly
shown by the DT and TR in~\cref{fig:02:dt,fig:02:tt}, feature 1 is
actually the only feature bearing influence in the prediction.
Clearly, the inevitable conclusion is that the information provided by
Shapley values for XAI is misleading.

\paragraph{Discussion.}
%~\\
As observed above, for the family of DTs discussed in this section,
namely those respecting \eqref{eq:05:cond00}, \eqref{eq:05:cond01},
\eqref{eq:05:cond02} and \eqref{eq:05:cond03}, and in the concrete
case of point $(1,1,2)$, the computed Shapley values will provide
misleading information.
Concretely, feature 1, which must be used to fix or to change the
prediction, is assigned a Shapley value of 0, thus signifying no
influence on the prediction~\cite{kononenko-jmlr10}. Moreover,
features 2 and 3, which are never used to fix or change the
prediction, are assigned a non-zero Shapley value, thus signifying
some influence on the prediction.
Therefore, for this family of classifiers, the computed Shapley values
produce information that bears \emph{no} correlation with actual
feature influence in either fixing or changing the prediction.
Similar results have been reported in a number of recent
reports~\cite{hms-corr23a,msh-corr23,hms-corr23b}.

\paragraph{Pinpointing the problem.}
%~\\
How come Shapley values can provide such misleading information?
A more detailed analysis of the example reveals that several sets of
fixed features, which play no role in abductively explaining the
prediction, serve to bias the computed Shapley values in a way such
that these bear no correlation with the relative importance of
features in for the given prediction. As an example, this is the case
with the sets $\{1,2\}$, $\{1,3\}$ and $\{1,2,3\}$, i.e.\ any proper
superset of the computed AXp.
In contrast, if we were to look at the sets of AXp's or CXp's we would
only account for the features that play some role either in
determining or in changing the prediction.

The results reported in this and earlier reports have already
motivated the proposal of alternatives to the use of Shapley
values~\cite{hms-corr23a,ignatiev-corr23,izza-corr23}, which aim at
computing relative orders of feature importance that respect the
actual logic operation of the ML model.

%\clearpage
%
%\input{exampl08}

\clearpage

\section{Example for the IJAR Journal Submission}

We consider the parameterized classifier shown in~\cref{tab:09:tt}.
From the table, we can conclude that $\fml{F}=\{1,2,3,4\}$,
$\mbb{D}_1=\mbb{D}_2=\mbb{D}_3=\{0,1\}$, $\mbb{D}_4=\{0,1,2\}$, and so
$\mbb{F}=\{0,1\}^3\times\{0,1,2\}$. Moreover, we also have
$\fml{K}=\{\alpha\}\cup\{\sigma_j\,|\,j=1,\ldots,9\}$.
We will require that $\alpha\not=\alpha_j,j=1,\ldots,9$;  this
constraint will be clarified below.
Finally, the target instance is $((1,2,2),\alpha)$.

\begin{table}[t]
  \begin{tabular}{cccccc} \toprule
    row \# & $x_1$ & $x_2$ & $x_3$ & $x_4$ & $\kappa_5(\mbf{x})$ \\ \toprule
    1      & 0     & 0     & 0     & 0 & 0 \\
    2      & 0     & 0     & 0     & 1 & $\sigma_1$ \\
    3      & 0     & 0     & 0     & 2 & 0 \\
    4      & 0     & 0     & 1     & 0 & 0 \\
    5      & 0     & 0     & 1     & 1 & $\sigma_2$ \\
    6      & 0     & 0     & 1     & 2 & 0 \\
    7      & 0     & 1     & 0     & 0 & 0 \\
    8      & 0     & 1     & 0     & 1 & $\sigma_3$ \\
    9      & 0     & 1     & 0     & 2 & 0 \\
    10     & 0     & 1     & 1     & 0 & 0 \\
    11     & 0     & 1     & 1     & 1 & $\sigma_4$ \\
    12     & 0     & 1     & 1     & 2 & 0 \\
    13     & 1     & 0     & 0     & 0 & $\alpha$ \\
    14     & 1     & 0     & 0     & 1 & $\alpha$ \\
    15     & 1     & 0     & 0     & 2 & $\alpha$ \\
    16     & 1     & 0     & 1     & 0 & $\alpha$ \\
    17     & 1     & 0     & 1     & 1 & $\alpha$ \\
    18     & 1     & 0     & 1     & 2 & $\alpha$ \\
    19     & 1     & 1     & 0     & 0 & $\alpha$ \\
    20     & 1     & 1     & 0     & 1 & $\alpha$ \\
    21     & 1     & 1     & 0     & 2 & $\alpha$ \\
    22     & 1     & 1     & 1     & 0 & $\alpha$ \\
    23     & 1     & 1     & 1     & 1 & $\alpha$ \\
    24     & 1     & 1     & 1     & 2 & $\alpha$ \\
    \bottomrule
  \end{tabular}
  \caption{Tabular representation for $\kappa$} \label{tab:09:tt}
\end{table}

%%Instance: $((1,2,2),\alpha)$.

%%\clearpage

\cref{fig:09:xps} summarizes the computation of AXps and CXps for the
parameterized classifier in~\cref{tab:09:tt}. As can be concluded,
feature 1 is relevant (and necessary), whereas features 2, 3 and 4 are
irrelevant.

\begin{figure*}[t]
  \begin{mdframed}[linewidth=1.5pt,linecolor=darkblue,roundcorner=5pt]
    %skipabove=10pt
    %

    \medskip\smallskip
    \begin{center}
      \renewcommand{\tabcolsep}{0.5em}
      \begin{tabular}{cccc}%{cccc|cccc}
        \toprule[1pt]
        $\fml{S}$ &
        $\rows(\fml{S})$ &
        \makecell{$\waxp(\fml{S})$?\\$\fml{S}$ sufficient?} &
        \makecell{$\axp(\fml{S})$?\\$\fml{S}$ also minimal?} %&
        %
        %$\fml{F}\setminus\fml{S}$ &
        %$\rows(\fml{F}\setminus\fml{S})$ &
        %\makecell{$\wcxp(\fml{S})$?\\$\fml{S}$ changes $\kappa$?} &
        %\makecell{$\cxp(\fml{S})$?\\$\fml{S}$ also minimal?}
        \\
        \midrule[0.875pt]
        $\emptyset$ &
        1..24 & \nomark & %&
        %$\{1,2,3,4\}$ & 12 & \nomark & 
        \\
        $\{1\}$ &
        13..24 & \yesmark & \yesmark %&
        %$\{2,3,4\}$ & 12,24 & \yesmark & \yesmark
        \\
        $\{2\}$ &
        7..12,19..24 & \nomark & %&
        %$\{1,3,4\}$ & 18,24 & \nomark &
        \\
        $\{3\}$ &
        4..6,10..12,16..18,22..24 & \nomark & %&
        %$\{1,2,4\}$ & 10,11,12 & \nomark &
        \\
        $\{4\}$ &
        3,6,9,12,15,18,21,24 &
        \\
        $\{1,2\}$ & 10,11,12 & \yesmark & \nomark %& 
        %$\{3\}$ & 3,6,9,12 & \yesmark & \nomark
        \\
        $\{1,3\}$ & 9,12 & \yesmark & \nomark %& 
        %$\{2\}$ & 4,5,6,10,11,12 & \yesmark & \nomark
        \\
        $\{1,4\}$ &
        \\
        $\{2,3\}$ & 6,12 & \nomark & %& 
        %$\{1\}$ & 7,8,9,10,11,12 & \nomark & 
        \\
        $\{2,4\}$ & 
        \\
        $\{3,4\}$ & 
        \\
        $\{1,2,3\}$ & 12 & \yesmark & \nomark %& 
        %$\emptyset$ & 1..12 & \yesmark & \nomark
        \\
        $\{1,2,4\}$ &
        \\
        $\{1,3,4\}$ &
        \\
        $\{2,3,4\}$ &
        \\
        $\{1,2,3,4\}$ &
        \\
        \bottomrule[1pt]
      \end{tabular}
    \end{center}
    \caption{Computing AXp's/CXp's for the example parameterized
      classifier shown in~\cref{tab:05:tt} and instance
      $(\mbf{v},c)=((1,1,2),\alpha)$. All subsets of features are
      considered.
      For computing AXp's, and for some set $\fml{S}$, the features in
      $\fml{S}$ are fixed to their values as determined by $\mbf{v}$.
      The picked rows, i.e.\ $\rows(\fml{S})$, are the rows consistent
      with those fixed values.
      For example, if $\fml{S}=\{1,2\}$, then only rows 10, 11 and 12
      are consistent with having features 1 and 2 assigned value 1.
      Similarly, for computing CXp's, and for some set $\fml{S}$, the
      features in $\fml{F}\setminus\fml{S}$ are fixed to their values
      as determined by $\mbf{v}$. The picked rows are again the rows
      consistent with those fixed values. For example, if
      $\fml{S}=\{2\}$, then $\fml{F}\setminus\fml{S}=\{1,3\}$, and
      so only rows 9 and 12 are consistent with having feature 1
      assigned value 1 and feature 3 assigned value 2.
      An AXp is an irreducible set of features that is sufficient for
      the prediction. In this example, only $\{1\}$ respects the
      criteria.
      Moreover, a CXp is an irreducible set of features which, if
      allowed to take any value from their domain, the prediction
      changes value. For this example, $\{1\}$ respect the criteria,
      i.e.\ by only changing feature $\{1\}$, we are able to change
      the prediction.
    }
    \label{fig:09:xps}
  \end{mdframed}
\end{figure*}

%%
%%\clearpage

\begin{table}[t]
  \begin{tabular}{ccc} \toprule
    $\fml{S}$ & rows picked by $\fml{S}$ & $\phi(\fml{S})$  \\ \toprule
    $\emptyset$ & 1..18 & $\nfrac{(\sum_{j=1}^{4}\sigma_j)}{24}+\nfrac{\alpha}{2}$ \\
    $\{1\}$ & 13..24 & $\alpha$ \\
    $\{2\}$ & 7..12,19..24 & $\nfrac{(\sigma_3+\sigma_4)}{12}+\nfrac{\alpha}{2}$ \\
    $\{3\}$ & 4..6,10..12,16..18,22..24 & $\nfrac{(\sigma_2+\sigma_4)}{12}+\nfrac{\alpha}{2}$ \\
    $\{4\}$ & 3,6,9,12,15,18,21,24 & $\nfrac{\alpha}{2}$ \\
    $\{1,2\}$ & 19..24 & $\alpha$ \\
    $\{1,3\}$ & 16..18,22..24 & $\alpha$ \\
    $\{1,4\}$ & 15,18,21,24 & $\alpha$ \\
    $\{2,3\}$ & 10..12,22..24 & $\nfrac{\sigma_4}{6}+\nfrac{\alpha}{2}$ \\
    $\{2,4\}$ & 9,12,21,24 & $\nfrac{\alpha}{2}$ \\
    $\{3,4\}$ & 6,12,18,24 & $\nfrac{\alpha}{2}$ \\
    $\{1,2,3\}$ & 22..24 & $\alpha$ \\
    $\{1,2,4\}$ & 21,24 & $\alpha$ \\
    $\{2,3,4\}$ & 12,24 & $\nfrac{\alpha}{2}$ \\
    $\{1,2,3,4\}$ & 24 & $\alpha$ \\
    \bottomrule
  \end{tabular}
  \caption{Computing $\phi(\fml{S})$, by inspecting the tabular
    representation}
  \label{tab:09:phi}
\end{table}

\begin{figure*}[t]
  \begin{mdframed}[linewidth=1.5pt,linecolor=darkblue,roundcorner=5pt]
    %skipabove=10pt
    %

    \medskip\smallskip
    %
    %% Feature 1:
    \begin{subfigure}{1.0\textwidth}
      \begin{center}
        \scalebox{0.995}{
          \renewcommand{\tabcolsep}{0.75em}
          \begin{tabular}{cccccc}
            \toprule[1pt]
            $\fml{S}$ &
            $\phi(\fml{S})$ &
            $\phi(\fml{S}\cup\{1\})$ &
            $\Delta(\fml{S})$ &
            $\varsigma(\fml{S})$ &
            $\varsigma(\fml{S})\times\Delta(\fml{S})$
            \\
            \midrule[0.875pt]
            $\emptyset$ &
            $\nfrac{(\sum_{j=1}^{4}\sigma_j)}{24}+\nfrac{\alpha}{2}$ &
            $\alpha$ &
            $\nfrac{\alpha}{2}-\nfrac{(\sum_{j=1}^{4}\sigma_j)}{24}$ &
            $\sfrac{0!(4-0-1)!}{4!}=\sfrac{1}{4}$ &
            $\nfrac{\alpha}{8}-\nfrac{(\sum_{j=1}^{4}\sigma_j)}{96}$
            \\
            $\{2\}$ &
            $\nfrac{(\sigma_3+\sigma_4)}{12}+\nfrac{\alpha}{2}$ &
            $\alpha$ &
            $\nfrac{\alpha}{2}-\nfrac{(\sigma_3+\sigma_4)}{12}$ &
            $\sfrac{1!(4-1-1)!}{4!}=\sfrac{1}{12}$ &
            $\nfrac{\alpha}{24}-\nfrac{(\sigma_3+\sigma_4)}{144}$
            \\
            $\{3\}$ &
            $\nfrac{(\sigma_2+\sigma_4)}{12}+\nfrac{\alpha}{2}$ &
            $\alpha$ &
            $\nfrac{\alpha}{2}-\nfrac{(\sigma_2+\sigma_4)}{12}$ &
            $\sfrac{1!(4-1-1)!}{4!}=\sfrac{1}{12}$ &
            $\nfrac{\alpha}{24}-\nfrac{(\sigma_2+\sigma_4)}{144}$
            \\
            $\{4\}$ &
            $\nfrac{\alpha}{2}$ &
            $\alpha$ &
            $\nfrac{\alpha}{2}$ &
            $\sfrac{1!(4-1-1)!}{4!}=\sfrac{1}{12}$ &
            $\nfrac{\alpha}{24}$
            \\
            $\{2,3\}$ &
            $\nfrac{\sigma_4}{6}+\nfrac{\alpha}{2}$ &
            $\alpha$ &
            $\nfrac{\alpha}{2}-\nfrac{\sigma_4}{6}$ &
            $\sfrac{2!(4-2-1)!}{4!}=\sfrac{1}{12}$ &
            $\nfrac{\alpha}{24}-\nfrac{\sigma_4}{72}$
            \\
            $\{2,4\}$ &
            $\nfrac{\alpha}{2}$ &
            $\alpha$ &
            $\nfrac{\alpha}{2}$ &
            $\sfrac{2!(4-2-1)!}{4!}=\sfrac{1}{12}$ &
            $\nfrac{\alpha}{24}$
            \\
            $\{3,4\}$ &
            $\nfrac{\alpha}{2}$ &
            $\alpha$ &
            $\nfrac{\alpha}{2}$ &
            $\sfrac{2!(4-2-1)!}{4!}=\sfrac{1}{12}$ &
            $\nfrac{\alpha}{24}$
            \\
            $\{2,3,4\}$ &
            $\nfrac{\alpha}{2}$ &
            $\alpha$ &
            $\nfrac{\alpha}{2}$ &
            $\sfrac{3!(4-3-1)!}{4!}=\sfrac{1}{4}$ &
            $\nfrac{\alpha}{8}$
            \\
            \midrule[0.75pt]
            \multicolumn{5}{r}{Shapley value for feature 1 \hfill
              $\sv(1)~~=$} &
            $\nfrac{\alpha}{2}-\nfrac{(3\sigma_1+5\sigma_2+5\sigma_3+11\sigma_4)}{288}$ \\
            \bottomrule[1pt]
          \end{tabular}
        }
      \end{center}
    \end{subfigure}
    %%% Obs: s1=6 ; s2=12 ; s3=12 ; s4=10    XXXXXXXX
    %%% Obs: Sum=144: s1=5 ; s2=2 ; s3=4 ; s4=9
    %%% 9x11 + 4*5 + 2*5 + 5*3 = 99 + 20 + 10 + 15 = 144
    
    \medskip\medskip\medskip
    %
    %% Feature 2:
    \begin{subfigure}{1.0\textwidth}
      \begin{center}
        \scalebox{0.995}{
          \renewcommand{\tabcolsep}{0.5em}
          \begin{tabular}{cccccc}
            \toprule[1pt]
            $\fml{S}$ &
            $\phi(\fml{S})$ &
            $\phi(\fml{S}\cup\{2\})$ &
            $\Delta(\fml{S})$ &
            $\varsigma(\fml{S})$ &
            $\varsigma(\fml{S})\times\Delta(\fml{S})$
            \\
            \midrule[0.875pt]
            $\emptyset$ &
            $\nfrac{(\sum_{j=1}^{4}\sigma_j)}{24}+\nfrac{\alpha}{2}$ &
            $\nfrac{(\sigma_3+\sigma_4)}{12}+\nfrac{\alpha}{2}$ &
            $-\nfrac{(\sigma_1+\sigma_2)}{24}+\nfrac{(\sigma_3+\sigma_4)}{24}$ &
            $\sfrac{0!(4-0-1)!}{4!}=\sfrac{1}{4}$ &
            $-\nfrac{(\sigma_1+\sigma_2)}{96}+\nfrac{(\sigma_3+\sigma_4)}{96}$
            \\
            $\{1\}$ &
            $\alpha$ &
            $\alpha$ &
            $0$ &
            $\sfrac{1!(4-1-1)!}{4!}=\sfrac{1}{12}$ &
            $0$
            \\
            $\{3\}$ &
            $\nfrac{(\sigma_2+\sigma_4)}{12}+\nfrac{\alpha}{2}$ &
            $\nfrac{\sigma_4}{6}+\nfrac{\alpha}{2}$ &
            $-\nfrac{(\sigma_2+\sigma_4)}{12}+\nfrac{\sigma_4}{6}$ &
            $\sfrac{1!(4-1-1)!}{4!}=\sfrac{1}{12}$ &
            $-\nfrac{(\sigma_2+\sigma_4)}{144}+\nfrac{\sigma_4}{72}$
            \\
            $\{4\}$ &
            $\nfrac{\alpha}{2}$ &
            $\nfrac{\alpha}{2}$ &
            $0$ &
            $\sfrac{1!(4-1-1)!}{4!}=\sfrac{1}{12}$ &
            $0$
            \\
            $\{1,3\}$ &
            $\alpha$ &
            $\alpha$ &
            $0$ &
            $\sfrac{2!(4-2-1)!}{4!}=\sfrac{1}{12}$ &
            $0$
            \\
            $\{1,4\}$ &
            $\alpha$ &
            $\alpha$ &
            $0$ &
            $\sfrac{2!(4-2-1)!}{4!}=\sfrac{1}{12}$ &
            $0$
            \\
            $\{3,4\}$ &
            $\nfrac{\alpha}{2}$ &
            $\nfrac{\alpha}{2}$ &
            $0$ &
            $\sfrac{2!(4-2-1)!}{4!}=\sfrac{1}{12}$ &
            $0$
            \\
            $\{1,3,4\}$ &
            $\alpha$ &
            $\alpha$ &
            $0$ &
            $\sfrac{3!(4-3-1)!}{4!}=\sfrac{1}{4}$ &
            $0$
            \\
            \midrule[0.75pt]
            \multicolumn{5}{r}{Shapley value for feature 2 \hfill
              $\sv(2)~~=$} &
            $\nfrac{(-3\sigma_1-5\sigma_2+3\sigma_3+5\sigma_4)}{288}$ \\
            \bottomrule[1pt]
          \end{tabular}
        }
      \end{center}
    \end{subfigure}
    
    \medskip\medskip\medskip
    %
    %% Feature 3:
    \begin{subfigure}{1.0\textwidth}
      \begin{center}
        \scalebox{0.995}{
          \renewcommand{\tabcolsep}{0.5em}
          \begin{tabular}{cccccc}
            \toprule[1pt]
            $\fml{S}$ &
            $\phi(\fml{S})$ &
            $\phi(\fml{S}\cup\{3\})$ &
            $\Delta(\fml{S})$ &
            $\varsigma(\fml{S})$ &
            $\varsigma(\fml{S})\times\Delta(\fml{S})$
            \\
            \midrule[0.875pt]
            $\emptyset$ &
            $\nfrac{(\sum_{j=1}^{4}\sigma_j)}{24}+\nfrac{\alpha}{2}$ &
            $\nfrac{(\sigma_2+\sigma_4)}{12}+\nfrac{\alpha}{2}$ &
            $-\nfrac{(\sigma_1+\sigma_3)}{24}+\nfrac{(\sigma_2+\sigma_4)}{24}$ &
            $\sfrac{0!(4-0-1)!}{4!}=\sfrac{1}{4}$ &
            $-\nfrac{(\sigma_1+\sigma_3)}{96}+\nfrac{(\sigma_2+\sigma_4)}{96}$
            \\
            $\{1\}$ &
            $\alpha$ &
            $\alpha$ &
            $0$ &
            $\sfrac{1!(4-1-1)!}{4!}=\sfrac{1}{12}$ &
            $0$
            \\
            $\{2\}$ &
            $\nfrac{(\sigma_3+\sigma_4)}{12}+\nfrac{\alpha}{2}$ &
            $\nfrac{\sigma_4}{6}+\nfrac{\alpha}{2}$ &
            $-\nfrac{(\sigma_3)}{12}+\nfrac{\sigma_4}{12}$ &
            $\sfrac{1!(4-1-1)!}{4!}=\sfrac{1}{12}$ &
            $-\nfrac{(\sigma_3)}{144}+\nfrac{\sigma_4}{144}$
            \\
            $\{4\}$ &
            $\nfrac{\alpha}{2}$ &
            $\nfrac{\alpha}{2}$ &
            $0$ &
            $\sfrac{1!(4-1-1)!}{4!}=\sfrac{1}{12}$ &
            $0$
            \\
            $\{1,2\}$ &
            $\alpha$ &
            $\alpha$ &
            $0$ &
            $\sfrac{2!(4-2-1)!}{4!}=\sfrac{1}{12}$ &
            $0$
            \\
            $\{1,4\}$ &
            $\alpha$ &
            $\alpha$ &
            $0$ &
            $\sfrac{2!(4-2-1)!}{4!}=\sfrac{1}{12}$ &
            $0$
            \\
            $\{2,4\}$ &
            $\nfrac{\alpha}{2}$ &
            $\nfrac{\alpha}{2}$ &
            $0$ &
            $\sfrac{2!(4-2-1)!}{4!}=\sfrac{1}{12}$ &
            $0$
            \\
            $\{1,2,4\}$ &
            $\alpha$ &
            $\alpha$ &
            $0$ &
            $\sfrac{3!(4-3-1)!}{4!}=\sfrac{1}{4}$ &
            $0$
            \\
            \midrule[0.75pt]
            \multicolumn{5}{r}{Shapley value for feature 3 \hfill
              $\sv(3)~~=$} &
            $\nfrac{(-3\sigma_1+3\sigma_2-5\sigma_3+5\sigma_4)}{288}$ \\
            \bottomrule[1pt]
          \end{tabular}
        }
      \end{center}
    \end{subfigure}
    
    \medskip\medskip\medskip
    %
    %% Feature 4:
    \begin{subfigure}{1.0\textwidth}
      \begin{center}
        \scalebox{0.995}{
          \renewcommand{\tabcolsep}{0.75em}
          \begin{tabular}{cccccc}
            \toprule[1pt]
            $\fml{S}$ &
            $\phi(\fml{S})$ &
            $\phi(\fml{S}\cup\{4\})$ &
            $\Delta(\fml{S})$ &
            $\varsigma(\fml{S})$ &
            $\varsigma(\fml{S})\times\Delta(\fml{S})$
            \\
            \midrule[0.875pt]
            $\emptyset$ &
            $\nfrac{(\sum_{j=1}^{4}\sigma_j)}{24}+\nfrac{\alpha}{2}$ &
            $\nfrac{\alpha}{2}$ &
            $\nfrac{(\sum_{j=1}^{4}\sigma_j)}{24}$ &
            $\sfrac{0!(4-0-1)!}{4!}=\sfrac{1}{4}$ &
            $\nfrac{(\sum_{j=1}^{4}\sigma_j)}{96}$
            \\
            $\{1\}$ &
            $\alpha$ &
            $\alpha$ &
            $0$ &
            $\sfrac{1!(4-1-1)!}{4!}=\sfrac{1}{12}$ &
            $0$
            \\
            $\{2\}$ &
            $\nfrac{(\sigma_3+\sigma_4)}{12}+\nfrac{\alpha}{2}$ &
            $\nfrac{\alpha}{2}$ &
            $-\nfrac{(\sigma_3+\sigma_4)}{12}$ &
            $\sfrac{1!(4-1-1)!}{4!}=\sfrac{1}{12}$ &
            $-\nfrac{(\sigma_3+\sigma_4)}{144}$
            \\
            $\{3\}$ &
            $\nfrac{(\sigma_2+\sigma_4)}{12}+\nfrac{\alpha}{2}$ &
            $\nfrac{\alpha}{2}$ &
            $-\nfrac{(\sigma_3+\sigma_4)}{12}$ &
            $\sfrac{1!(4-1-1)!}{4!}=\sfrac{1}{12}$ &
            $-\nfrac{(\sigma_2+\sigma_4)}{144}$
            \\
            $\{1,2\}$ &
            $\alpha$ &
            $\alpha$ &
            $0$ &
            $\sfrac{2!(4-2-1)!}{4!}=\sfrac{1}{12}$ &
            $0$
            \\
            $\{1,3\}$ &
            $\alpha$ &
            $\alpha$ &
            $0$ &
            $\sfrac{2!(4-2-1)!}{4!}=\sfrac{1}{12}$ &
            $0$
            \\
            $\{2,3\}$ &
            $\nfrac{(\sigma_4)}{6}+\nfrac{\alpha}{2}$ &
            $\nfrac{\alpha}{2}$ &
            $-\nfrac{\sigma_4}{6}$ &
            $\sfrac{2!(4-2-1)!}{4!}=\sfrac{1}{12}$ &
            $-\nfrac{\sigma_4}{72}$
            \\
            $\{1,2,3\}$ &
            $\alpha$ &
            $\alpha$ &
            $0$ &
            $\sfrac{3!(4-3-1)!}{4!}=\sfrac{1}{4}$ &
            $0$
            \\
            \midrule[0.75pt]
            \multicolumn{5}{r}{Shapley value for feature 4 \hfill
              $\sv(4)~~=$} &
            $\nfrac{(3\sigma_1+\sigma_2+\sigma_3-5\sigma_4)}{288}$ \\
            \bottomrule[1pt]
          \end{tabular}
        }
      \end{center}
    \end{subfigure}

    \caption{Computation of Shapley values for the example DT and
      instance $((1,1,1,2),\alpha)$. For each feature $i$, the sets to
      consider are all the sets that do not include the feature.
      The average values are obtained by summing up the values of the
      classifier in the rows consistent with $\fml{S}$ and dividing by
      the total number of rows.
      %For $\fml{S}=\{2,4\}$, only row 3 in the truth table takes value
      %1, and so the average becomes $\sfrac{1}{4}$.
    }
    \label{fig:09:svs}
  \end{mdframed}
\end{figure*}

Given the computation of the Shapley values in~\cref{fig:09:svs}, and
the goal of have $\sv(1)=0$, $\sv(2)\not=0$, $\sv(3)\not=0$, and
$\sv(4)\not=0$, we obtain the following constraints:
\begin{align}
  & \alpha=\nfrac{(3\sigma_1+5\sigma_2+5\sigma_3+11\sigma_4)}{144}
  \label{eq:09:svs:01} \\
  & \nfrac{(-3\sigma_1-5\sigma_2+3\sigma_3+5\sigma_4)}{288}
  \label{eq:09:svs:02}\not=0 \\
  & \nfrac{(-3\sigma_1+3\sigma_2-5\sigma_3+5\sigma_4)}{288}
  \label{eq:09:svs:03}\not=0 \\
  & \nfrac{(3\sigma_1+\sigma_2+\sigma_3-5\sigma_4)}{288}
  \label{eq:09:svs:04}\not=0 %%\\
\end{align}

Any pick of values of $\alpha$, $\sigma_j, j=1,\ldots,9$ that
satisfies the constraints above will represent a classifier where the
relative order of feature importance obtained with Shapley values is
misleading.

%%\clearpage

\paragraph{Instantiation.} Let us pick
    %%% Obs: Sum=144: s1=5 ; s2=2 ; s3=4 ; s4=9
$\sigma_1=5, \sigma_2=2, \sigma_3=4, \sigma_4=9$, such that $\alpha=1$.
It is easy to conclude that these values
satisfy~\eqref{eq:09:svs:01},~\eqref{eq:09:svs:02},~\eqref{eq:09:svs:03}.
\cref{tab:09:tt2,tab:09:dt2} show the resulting tabular representation
and decision tree.

\begin{equation}
  \begin{array}{l}
    \alpha=1 %\nfrac{(3\sigma_1+5\sigma_2+5\sigma_3+11\sigma_4)}{144}
    \\[3pt]
    \begin{aligned}
      -3\sigma_1-5\sigma_2+3\sigma_3+5\sigma_4\not=0 & \Leftrightarrow\\
      -15-10+12+45 \not=0 & \Leftrightarrow\\
      32\not=0 & \\[2.5pt]
      -3\sigma_1+3\sigma_2-5\sigma_3+5\sigma_4\not=0 & \Leftrightarrow\\
      -15+6-20+45\not=0 & \Leftrightarrow \\
      56\not=0 & \\[2.5pt]
      3\sigma_1+\sigma_2+\sigma_3-5\sigma_4 \not=0 & \Leftrightarrow\\
      15+2+4-45 \not=0 & \Leftrightarrow \\
      -24\not=0 &
    \end{aligned}
  \end{array}
\end{equation}

Furthermore, we get
$\sv(1)=0$,
$\sv(2)=\nfrac{32}{288}=\nfrac{1}{9}=0.111$,
$\sv(3)=\nfrac{56}{288}=\nfrac{7}{36}=0.194$,
$\sv(4)=-\nfrac{24}{288}=-\nfrac{1}{12}=-0.083$.

\begin{table}[t]
  \begin{tabular}{cccccc} \toprule
    row \# & $x_1$ & $x_2$ & $x_3$ & $x_4$ & $\kappa_5(\mbf{x})$ \\ \toprule
    1      & 0     & 0     & 0     & 0 & 0 \\
    2      & 0     & 0     & 0     & 1 & $5$ \\
    3      & 0     & 0     & 0     & 2 & 0 \\
    4      & 0     & 0     & 1     & 0 & 0 \\
    5      & 0     & 0     & 1     & 1 & $2$ \\
    6      & 0     & 0     & 1     & 2 & 0 \\
    7      & 0     & 1     & 0     & 0 & 0 \\
    8      & 0     & 1     & 0     & 1 & $4$ \\
    9      & 0     & 1     & 0     & 2 & 0 \\
    10     & 0     & 1     & 1     & 0 & 0 \\
    11     & 0     & 1     & 1     & 1 & $9$ \\
    12     & 0     & 1     & 1     & 2 & 0 \\
    13     & 1     & 0     & 0     & 0 & $1$ \\
    14     & 1     & 0     & 0     & 1 & $1$ \\
    15     & 1     & 0     & 0     & 2 & $1$ \\
    16     & 1     & 0     & 1     & 0 & $1$ \\
    17     & 1     & 0     & 1     & 1 & $1$ \\
    18     & 1     & 0     & 1     & 2 & $1$ \\
    19     & 1     & 1     & 0     & 0 & $1$ \\
    20     & 1     & 1     & 0     & 1 & $1$ \\
    21     & 1     & 1     & 0     & 2 & $1$ \\
    22     & 1     & 1     & 1     & 0 & $1$ \\
    23     & 1     & 1     & 1     & 1 & $1$ \\
    24     & 1     & 1     & 1     & 2 & $1$ \\
    \bottomrule
  \end{tabular}
  \caption{Tabular representation for $\kappa_5$} \label{tab:09:tt2}
\end{table}

\begin{figure}[t]
  % Concocted example
%%
%\tikzset{every label/.style={xshift=-0.35ex,
%  yshift=-5.25ex,
%  text width=1ex,
%  align=right, inner sep=1pt, font=\tiny, text=midblue}}
%%
%\tikzset{tlabel/.style={xshift=0.25ex, yshift=1.75ex, text width=1ex,
%    align=right, inner sep=1pt, font=\tiny, text=midblue}}
%%%\tikzset{every node/.style={---rectangle---}}
%
\forestset{
  BDT/.style={
    for tree={
      l=1.5cm,s sep=1.15cm,
      if n children=0{}{circle}, %rectangle
      %if n children=0{}{draw},
      draw=midblue,%draw=black,%
      text=midblue,%text=black,%
      edge={
        my edge
      },
      %if n=1{
      %  edge+={0 my edge},
      %}{},
      edge=thick,
    }
  },
}
\begin{forest}
  BDT
  [{$x_1$}, label={[yshift=-6.875ex]{{\tiny1}}} 
    [{$x_4$}, label={[yshift=-6.875ex]{{\tiny2}}}, %edge={very thick}, %top-left=x
      edge label={node[midway,left,xshift=-0.5pt] {{\scriptsize$\in\{0\}$}}}
      [{$x_2$}, label={[yshift=-6.875ex]{{\tiny4}}}, %xshift=-3.075ex,yshift=-3.575ex
        edge label={node[midway,left,xshift=-1.5pt] {{\scriptsize$\in\{0\}$}}}
        [{$x_3$}, label={[yshift=-6.875ex]{{\tiny6}}}, %xshift=-3.075ex,yshift=-3.575ex
          edge label={node[midway,left,xshift=-1.5pt] {{\scriptsize$\in\{1\}$}}}
          [\dghlight{\textbf{5}}, label={[yshift=-5.25ex]{{\tiny8}}},
            edge label={node[midway,left,xshift=-0.5pt] {{\scriptsize$\in\{0\}$}}}, rectangle, fill={tblue2!25} ]
          [\dghlight{\textbf{2}}, label={[yshift=-5.25ex]{{\tiny9}}},
            edge label={node[midway,right,xshift=-0.575pt] {{\scriptsize$\in\{1\}$}}}, rectangle, fill={tblue2!25} ]
        ]
        [{$x_3$}, label={[yshift=-6.875ex]{{\tiny7}}}, %xshift=-3.075ex,yshift=-3.575ex
          edge label={node[midway,right,xshift=0.25pt] {{\scriptsize$\in\{1\}$}}}
          [\dghlight{\textbf{4}}, label={[yshift=-5.25ex]{{\tiny10}}},
            edge label={node[midway,left,xshift=-0.5pt] {{\scriptsize$\in\{0\}$}}}, rectangle, fill={tblue2!25} ]
          [\dghlight{\textbf{9}}, label={[yshift=-5.25ex]{{\tiny11}}},
            edge label={node[midway,right,xshift=-0.575pt] {{\scriptsize$\in\{1\}$}}}, rectangle, fill={tblue2!25} ]
        ]
      ]
      [\dghlight{\textbf{0}}, label={[yshift=-5.25ex]{{\tiny5}}},
        edge label={node[midway,right,xshift=-0.5pt] {{\scriptsize$\in\{0,2\}$}}}, rectangle, fill={tblue2!20} ]
    ]
    [\dghlight{\textbf{1}}, label={[yshift=-5.25ex]{{\tiny3}}},
      edge={very thick}, edge label={node[midway,right,xshift=0.5pt] {{\scriptsize$\in\{1\}$}}}, rectangle, fill={tblue2!25} ]
  ]
\end{forest}
  \caption{DT for instantiated classifier in~\cref{tab:09:tt2}}
  \label{tab:09:dt2}
\end{figure}

\clearpage

\begin{figure*}[t]
  \begin{mdframed}[linewidth=1.5pt,linecolor=darkblue,roundcorner=5pt] %skipabove=10pt
    \begin{subfigure}[b]{0.29\textwidth}
      \begin{center}
        % Concocted example
%%
%\tikzset{every label/.style={xshift=-0.35ex,
%  yshift=-5.25ex,
%  text width=1ex,
%  align=right, inner sep=1pt, font=\tiny, text=midblue}}
%%
%\tikzset{tlabel/.style={xshift=0.25ex, yshift=1.75ex, text width=1ex,
%    align=right, inner sep=1pt, font=\tiny, text=midblue}}
%%%\tikzset{every node/.style={---rectangle---}}
%
\forestset{
  BDT/.style={
    for tree={
      l=1.5cm,s sep=1.15cm,
      if n children=0{}{circle}, %rectangle
      %if n children=0{}{draw},
      draw=midblue,%draw=black,%
      text=midblue,%text=black,%
      edge={
        my edge
      },
      %if n=1{
      %  edge+={0 my edge},
      %}{},
      edge=thick,
    }
  },
}
\begin{forest}
  BDT
  [{$x_1$}, label={[yshift=-6.875ex]{{\tiny1}}} 
    [{$x_3$}, label={[yshift=-6.875ex]{{\tiny2}}}, %edge={very thick}, %top-left=x
      edge label={node[midway,left,xshift=-0.5pt] {{\scriptsize$\in\{0\}$}}}
      [{$x_2$}, label={[yshift=-6.875ex]{{\tiny4}}}, %xshift=-3.075ex,yshift=-3.5ex
        edge label={node[midway,left,xshift=-1.5pt] {{\scriptsize$\in\{1\}$}}}
        [\dghlight{\textbf{4}}, label={[yshift=-5.25ex]{{\tiny6}}},
          edge label={node[midway,left,xshift=-0.5pt] {{\scriptsize$\in\{0\}$}}}, rectangle, fill={tblue2!25} ]
        [\dghlight{\textbf{7}}, label={[yshift=-5.25ex]{{\tiny7}}},
          edge label={node[midway,right,xshift=-0.575pt] {{\scriptsize$\in\{1\}$}}}, rectangle, fill={tblue2!25} ]
      ]
      [\dghlight{\textbf{0}}, label={[yshift=-5.25ex]{{\tiny5}}},
        edge label={node[midway,right,xshift=-0.5pt] {{\scriptsize$\in\{0,2\}$}}},
        rectangle, fill={tblue2!20} ]
    ]
    [\dghlight{\textbf{1}}, label={[yshift=-5.25ex]{{\tiny3}}},
      edge={very thick}, edge label={node[midway,right,xshift=0.5pt] {{\scriptsize$\in\{1\}$}}},
      rectangle, fill={tblue2!25} ]
  ]
\end{forest}
        %\begin{tabular}{c} \\[5pt] \\[5pt] \\[5pt] \end{tabular}
        %%%$\kappa(\cdot) = \ldots$
      \end{center}
      \caption{Decision tree (DT) for $\kappa_1$} \label{fig:cacm:dt1}
    \end{subfigure}
    \begin{subfigure}[b]{0.39\textwidth}
      \begin{center}
        \begin{tabular}{ccccccc} \toprule
          row \# & $x_1$ & $x_2$ & $x_3$ & $\kappa_1(\mbf{x})$ & $\kappa_2(\mbf{x})$ &
          \\ \toprule
          1 & 0 & 0 & 0 & 0 & 0 \\
          2 & 0 & 0 & 1 & 4 & 3 \\
          3 & 0 & 0 & 2 & 0 & 0 \\
          4 & 0 & 1 & 0 & 0 & 0 \\
          5 & 0 & 1 & 1 & 7 & 2 \\
          6 & 0 & 1 & 2 & 0 & 0 \\
          7 & 1 & 0 & 0 & 1 & 1 \\
          8 & 1 & 0 & 1 & 1 & 1 \\
          9 & 1 & 0 & 2 & 1 & 1 \\
          10 & 1 & 1 & 0 & 1 & 1 \\
          11 & 1 & 1 & 1 & 1 & 1 \\
          12 & 1 & 1 & 2 & 1 & 1 \\
          \bottomrule
        \end{tabular}
      \end{center}
      \medskip %%\begin{tabular}{c} \\ \end{tabular}
      \caption{Tabular representations (TRs) for $\kappa_1$ and $\kappa_2$} \label{fig:cacm:tt}
    \end{subfigure}
    \begin{subfigure}[b]{0.29\textwidth}
      \begin{center}
        % Concocted example
%%
%\tikzset{every label/.style={xshift=-0.35ex,
%  yshift=-5.25ex,
%  text width=1ex,
%  align=right, inner sep=1pt, font=\tiny, text=midblue}}
%%
%\tikzset{tlabel/.style={xshift=0.25ex, yshift=1.75ex, text width=1ex,
%    align=right, inner sep=1pt, font=\tiny, text=midblue}}
%%%\tikzset{every node/.style={---rectangle---}}
%
\forestset{
  BDT/.style={
    for tree={
      l=1.5cm,s sep=1.15cm,
      if n children=0{}{circle}, %rectangle
      %if n children=0{}{draw},
      draw=midblue,%draw=black,%
      text=midblue,%text=black,%
      edge={
        my edge
      },
      %if n=1{
      %  edge+={0 my edge},
      %}{},
      edge=thick,
    }
  },
}
\begin{forest}
  BDT
  [{$x_1$}, label={[yshift=-6.875ex]{{\tiny1}}} 
    [{$x_3$}, label={[yshift=-6.875ex]{{\tiny2}}}, %edge={very thick}, %top-left=x
      edge label={node[midway,left,xshift=-0.5pt] {{\scriptsize$\in\{0\}$}}}
      [{$x_2$}, label={[yshift=-6.875ex]{{\tiny4}}}, %xshift=-3.075ex,yshift=-3.5ex
        edge label={node[midway,left,xshift=-1.5pt] {{\scriptsize$\in\{1\}$}}}
        [\dghlight{\textbf{3}}, label={[yshift=-5.25ex]{{\tiny6}}},
          edge label={node[midway,left,xshift=-0.5pt] {{\scriptsize$\in\{0\}$}}}, rectangle, fill={tblue2!25} ]
        [\dghlight{\textbf{2}}, label={[yshift=-5.25ex]{{\tiny7}}},
          edge label={node[midway,right,xshift=-0.575pt] {{\scriptsize$\in\{1\}$}}}, rectangle, fill={tblue2!25} ]
      ]
      [\dghlight{\textbf{0}}, label={[yshift=-5.25ex]{{\tiny5}}},
        edge label={node[midway,right,xshift=-0.5pt] {{\scriptsize$\in\{0,2\}$}}},
        rectangle, fill={tblue2!20} ]
    ]
    [\dghlight{\textbf{1}}, label={[yshift=-5.25ex]{{\tiny3}}},
      edge={very thick}, edge label={node[midway,right,xshift=0.5pt] {{\scriptsize$\in\{1\}$}}},
      rectangle, fill={tblue2!25} ]
  ]
\end{forest}
        %\begin{tabular}{c} \\[5pt] \\[5pt] \\[5pt] \end{tabular}
        %%%$\kappa(\cdot) = \ldots$
      \end{center}
      \caption{Decision tree (DT) for $\kappa_2$} \label{fig:cacm:dt2}
    \end{subfigure}

    \smallskip
    \begin{subfigure}[b]{0.975\textwidth}
      \setlength{\fboxrule}{0.875pt}
      \setlength{\fboxsep}{2.5pt}
      \fbox{
        \begin{minipage}{\textwidth}
          Analysis of the two classifiers: \\
          %DT \& truth table (TT): \\ %Instance $((1,1,2),1)$: \\
          By inspection of the DTs/TRs for both $\kappa_1$ and
          $\kappa_2$, it is immediate that: (i) if $x_1=1$, then the
          prediction must be 1; (ii) if $x_1=0$, then the prediction
          must not be 1. Thus, the classifiers predict class 1 (or
          predict a class other than 1) \emph{independently} of the
          values assigned to $x_2$ and $x_3$.
          %clear that the prediction is 1 whenever $x_1=1$,
          %\emph{independently} of the values of $x_2$ and $x_3$. %\\
          Hence, for point $\mbf{v}=(1,1,2)$, the prediction will be 1
          as long as $x_1=1$, \emph{independently} of $x_2$ and
          $x_3$.
          %These conclusions apply to both $\kappa_1$ and  $\kappa_2$.
          To change the prediction from class, the value of $x_1$ must
          be changed. In this case, the prediction will change to a
          class other than 1 \emph{independently} of the values
          assigned to $x_2$ and $x_3$. %\\
          %Thus, given the instance $((1,1,2),1)$, the value of the
          %prediction is determined \emph{uniquely} by the value of
          %$x_1$, and it is independent of the values of $x_2$ and
          %$x_3$. A change of prediction is also \emph{uniquely}
          %determined by the value of $x_1$, and it is again
          %independent of $x_2$ and $x_3$. Once again,
          These observations apply to both $\kappa_1$ and
          $\kappa_2$.\\
          The influence of features can be related with adversarial
          examples for the prediction of class 1. To change the
          prediction, it must be the case that feature 1 changes its
          value. Furthermore, any $l_0$ minimal adversarial example
          will only change the value of feature 1.
        \end{minipage}
      }
      
      \caption{Influence of features on $\kappa_1$ and $\kappa_2$ for
        $((1,1,2),1)$}
      \label{fig:cacm:disc}
    \end{subfigure}

    \caption{Example classifiers -- decision trees and their tabular
      representations.
     For these two classifiers, we have $\fml{F}=\{1,2,3\}$,
     $\mbb{D}_1=\mbb{D}_2=\{0,1\}$, and $\mbb{D}_3=\{0,1,2\}$,
     $\mbb{F}=\{0,1\}^2\times\{0,1,2\}$, and
     $\fml{K}=\{0,1,2,3,4,5,6,7\}$, albeit the DTs and TRs only use a
     subset of the classes.
     Literals in the DTs are represented with set notation, as used
     in earlier work~\cite{iims-jair22}; this solution yields more
     compact DTs.
     The classification functions are given by the decision trees, or
     alternatively by the tabular representations.
     The instance considered is $((1,1,2),1)$, which is consistent
     with path $\langle1,3\rangle$ in both DTs; this path is
     highlighted in the two DTs. The prediction is 1, as indicated in
     terminal node 3.%\\
    }
    \label{fig:cacm}
  \end{mdframed}
\end{figure*}

\section{Example for CACM Journal Submission} \label{sec:ex:cacm}

This section studies the classifiers shown in~\cref{fig:cacm}.
The decision trees (DTs) for the classifiers are depicted
in~\cref{fig:cacm:dt1,fig:cacm:dt2}, and the corresponding tabular
representations (TRs) of the classifiers are presented
in~\cref{fig:cacm:tt}.
The instance considered is $(\mbf{v},c)=((1,1,2),1)$, i.e.\ our goal
is to explain the prediction of 1 for the point $(1,1,2)$ in feature
space.
A brief analysis of the influence of features on the classifier's
behavior is presented in~\cref{fig:cacm:disc}.
This analysis is based solely on inspecting the functions computed by 
the classifiers, as shown either in the DTs or the TR.

%%\jnote{Analyze example(s) in terms of adversarial examples!!!}

\paragraph{An hypothetical scenario.}
%~\\
To motivate the analysis of the classifier in~\cref{fig:cacm}, we
consider the following hypothetical scenario\footnote{%
It would be fairly straightforward to create two datasets, e.g.\ the
two TRs shown, from which the DTs shown
in~\cref{fig:cacm:dt1,fig:cacm:dt2} would be induced using existing tree
learning tools.}.
A small college aims to predict the number of extra-curricular
activities of each of its students, where this number can be between 0
and 7. Let feature 1 represent whether the student is a honors student
(0 for no, and 1 for yes). Let feature 2 represent where the student
originates from a urban/non-urban household (0 for non-urban, and 1
for urban). Finally, let feature 3 represent whether the student's
field of study is humanities, arts or sciences (0 for humanities, 1
for arts and 2 for sciences). Thus, the target instance $((1,1,2),1)$
denotes a honors student from an urban household, studying sciences,
for whom the predicted number of extra-curricular activities is 1.

\begin{table}[t]
  \begin{tabular}{ccccc} \toprule
    row \# & $x_1$ & $x_2$ & $x_3$ & $\kappa(\mbf{x})$ \\ \toprule
    1      & 0     & 0     & 0     & $\sigma_1$ \\
    2      & 0     & 0     & 1     & $\sigma_2$ \\
    3      & 0     & 0     & 2     & $\sigma_3$ \\
    4      & 0     & 1     & 0     & $\sigma_4$ \\
    5      & 0     & 1     & 1     & $\sigma_5$ \\
    6      & 0     & 1     & 2     & $\sigma_6$ \\
    7      & 1     & 0     & 0     & $\alpha$ \\
    8      & 1     & 0     & 1     & $\alpha$ \\
    9      & 1     & 0     & 2     & $\alpha$ \\
    10     & 1     & 1     & 0     & $\alpha$ \\
    11     & 1     & 1     & 1     & $\alpha$ \\
    12     & 1     & 1     & 2     & $\alpha$ \\
    \bottomrule
  \end{tabular}
  \caption{Tabular representation for parameterized classifier}
  %either $\kappa_1$ or $\kappa_2$
  \label{tab:cacm:ptt}
\end{table}

\paragraph{Parameterized example.}
%~\\
Throughout the remainder of this section, we will consider a more
general classifier, shown in~\cref{tab:cacm:ptt}, which encompasses
the two classifiers shown in~\cref{fig:cacm}. As clarified below,
we impose that,
\begin{equation} \label{eq:cacm:cond00}
  \alpha\not=\sigma_j, j=1\ldots,6
\end{equation}
For simplicity, we also require that
$\alpha,\sigma_j\in\mbb{Z}^{+}_0, j=1,\ldots,6$.
It is plain that the DT of~\cref{fig:cacm:dt1} is a concrete
instantiation of the parameterized classifier shown
in~\cref{tab:cacm:ptt}, by setting $\sigma_1=\sigma_3=\sigma_4=\sigma_6=0$,
$\sigma_2=4$, $\sigma_5=7$ and $\alpha=1$.
For the parameterized example, the instance that will be considered is
$((1,1,2),\alpha)$.

\paragraph{Feature selection with formal explanations.}
%~\\
For both the classifiers of~\cref{fig:cacm} or the parameterized
classifier of~\cref{tab:cacm:ptt} (under the stated assumptions), it
is simple to show that $\mbb{A} = \{ \{1\} \}$ and that
$\mbb{C} = \{ \{1\} \}$. The computation of AXp's/CXp's is shown
in~\cref{fig:cacm:xps}. (Observe that computation of AXp's/CXp's shown
holds as long as $\alpha$ \emph{differs} from each of the $\sigma_i$,
$i=1,\ldots,6$, i.e.~\eqref{eq:05:cond00} holds.)
Thus, for any of these classifiers, feature 1 is relevant (in fact it
is necessary), and features 2 and 3 are irrelevant.
These results agree with the analysis of the
classifier (see~\cref{fig:cacm}) in terms of feature influence, in
that feature 1 occurs in explanations, and features 2 and 3 do not.
More importantly, there is not difference in terms of the computed
explanations for either $\kappa_1$ or $\kappa_2$.

\begin{figure*}[t]
  \begin{mdframed}[linewidth=1.5pt,linecolor=darkblue,roundcorner=5pt]
    %skipabove=10pt
    %

    \medskip\smallskip
    \begin{center}
      \renewcommand{\tabcolsep}{0.5em}
      \begin{tabular}{cccc|cccc}
        \toprule[1pt]
        $\fml{S}$ &
        $\rows(\fml{S})$ &
        \makecell{$\waxp(\fml{S})$?\\$\fml{S}$ sufficient?} &
        \makecell{$\axp(\fml{S})$?\\$\fml{S}$ also minimal?} &
        $\fml{F}\setminus\fml{S}$ &
        $\rows(\fml{F}\setminus\fml{S})$ &
        \makecell{$\wcxp(\fml{S})$?\\$\fml{S}$ changes $\kappa$?} &
        \makecell{$\cxp(\fml{S})$?\\$\fml{S}$ also minimal?}
        \\
        \midrule[0.875pt]
        $\emptyset$ &
        1..12 & \nomark & &
        $\{1,2,3\}$ & 12 & \nomark & 
        \\
        $\{1\}$ &
        7,8,9,10,11,12 & \yesmark & \yesmark &
        $\{2,3\}$ & 6,12 & \yesmark & \yesmark
        \\
        $\{2\}$ &
        4,5,6,10,11,12 & \nomark & &
        $\{1,3\}$ & 9,12 & \nomark &
        \\
        $\{3\}$ &
        3,6,9,12 & \nomark & &
        $\{1,2\}$ & 10,11,12 & \nomark &
        \\
        $\{1,2\}$ & 10,11,12 & \yesmark & \nomark & 
        $\{3\}$ & 3,6,9,12 & \yesmark & \nomark
        \\
        $\{1,3\}$ & 9,12 & \yesmark & \nomark & 
        $\{2\}$ & 4,5,6,10,11,12 & \yesmark & \nomark
        \\
        $\{2,3\}$ & 6,12 & \nomark & & 
        $\{1\}$ & 7,8,9,10,11,12 & \nomark & 
        \\
        $\{1,2,3\}$ & 12 & \yesmark & \nomark & 
        $\emptyset$ & 1..12 & \yesmark & \nomark
        \\
        \bottomrule[1pt]
      \end{tabular}
    \end{center}
    \smallskip

    \caption{Computing AXp's/CXp's for the example parameterized
      classifier shown in~\cref{tab:05:tt} and instance
      $(\mbf{v},c)=((1,1,2),\alpha)$. All subsets of features are
      considered.
      For computing AXp's, and for some set $\fml{S}$, the features in
      $\fml{S}$ are fixed to their values as determined by $\mbf{v}$.
      The picked rows, i.e.\ $\rows(\fml{S})$, are the rows consistent
      with those fixed values.
      For example, if $\fml{S}=\{1,2\}$, then only rows 10, 11 and 12
      are consistent with having features 1 and 2 assigned value 1.
      Similarly, for computing CXp's, and for some set $\fml{S}$, the
      features in $\fml{F}\setminus\fml{S}$ are fixed to their values
      as determined by $\mbf{v}$. The picked rows are again the rows
      consistent with those fixed values. For example, if
      $\fml{S}=\{2\}$, then $\fml{F}\setminus\fml{S}=\{1,3\}$, and
      so only rows 9 and 12 are consistent with having feature 1
      assigned value 1 and feature 3 assigned value 2.
      An AXp is an irreducible set of features that is sufficient for
      the prediction. In this example, only $\{1\}$ respects the
      criteria.
      Moreover, a CXp is an irreducible set of features which, if
      allowed to take any value from their domain, the prediction
      changes value. For this example, $\{1\}$ respect the criteria,
      i.e.\ by only changing feature $\{1\}$, we are able to change
      the prediction.
    }
    \label{fig:cacm:xps}
  \end{mdframed}
\end{figure*}

As can be concluded, the computed abductive and contrastive
explanations agree with the analysis shown in~\cref{fig:cacm:disc} in
terms of feature influence. Indeed, features 2 and 3, which have no
influence in determining nor in changing the prediction, are not
included in the computed explanations. In contrast, feature 1, which
is solely responsible for the prediction, is included in the computed
explanations.
Also unsurprisingly~\cite{inms-nips19,inams-aiia20,hms-corr23c}, the
existence of adversarial examples is tightly related with CXps, and so
indirectly with AXps.
These observations should be expected, since abductive explanations
are derived in terms of logic-based abduction, and so have a precise
logic formalization.

\paragraph{Feature attribution with Shapley values.}
%~\\
We will consider one of the most popular XAI approaches, which
consists in attributing relative feature importance by exploiting
Shapley
values~\cite{conklin-asmbi01,kononenko-jmlr10,kononenko-kis14,lundberg-nips17,barcelo-aaai21,vandenbroeck-aaai21,vandenbroeck-jair22,barcelo-jmlr23}.
To compute the Shapley values, it will be convenient to evaluate the
value of $\phi$ (see~\eqref{eq:phi}), i.e.\ the average value of the
classifier when the features of a given set $\fml{S}$ are fixed, for
each of the possible sets $\fml{S}$ we will need.  This is shown
in~\cref{tab:cacm:phi}.

\begin{table}[t]
  \begin{tabular}{ccc} \toprule
    $\fml{S}$ & rows picked by $\fml{S}$ & $\phi(\fml{S})$  \\ \toprule
    $\emptyset$ & 1..12 & $\nfrac{(\sum_{j=1}^{6}\sigma_j)}{12}+\nfrac{\alpha}{2}$ \\
    $\{1\}$ & 7..12 & $\alpha$ \\
    $\{2\}$ & 4..6,10..12 & $\nfrac{(\sigma_4+\sigma_5+\sigma_6)}{6}+\nfrac{\alpha}{2}$ \\
    $\{3\}$ & 3,6,9,12 & $\nfrac{(\sigma_3+\sigma_6)}{4}+\nfrac{\alpha}{2}$ \\
    $\{1,2\}$ & 10..12 & $\alpha$ \\
    $\{1,3\}$ & 9,12 & $\alpha$ \\
    $\{2,3\}$ & 6,12 & $\nfrac{\sigma_6}{2}+\nfrac{\alpha}{2}$ \\
    $\{1,2,3\}$ & 12 & $\alpha$ \\
    \bottomrule
  \end{tabular}
  \caption{Computing $\phi(\fml{S})$, by inspecting the tabular
    representation}
  \label{tab:cacm:phi}
\end{table}

\begin{figure*}[t]
  \begin{mdframed}[linewidth=1.5pt,linecolor=darkblue,roundcorner=5pt]
    %skipabove=10pt
    %

    \medskip\smallskip
    %
    %% Feature 1:
    \begin{subfigure}{1.0\textwidth}
      \begin{center}
        \renewcommand{\tabcolsep}{0.45em}
        \begin{tabular}{cccccc}
          \toprule[1pt]
          $\fml{S}$ &
          $\phi(\fml{S})$ &
          $\phi(\fml{S}\cup\{1\})$ &
          $\Delta(\fml{S})$ &
          $\varsigma(\fml{S})$ &
          $\varsigma(\fml{S})\times\Delta(\fml{S})$
          \\
          \midrule[0.875pt]
          $\emptyset$ &
          $\nfrac{(\sum_{j=1}^{6}\sigma_j)}{12}+\nfrac{\alpha}{2}$ &
          $\alpha$ &
          $\nfrac{\alpha}{2}-\nfrac{(\sum_{j=1}^{6}\sigma_j)}{12}$ &
          $\sfrac{0!(3-0-1)!}{3!}=\sfrac{1}{3}$ &
          $\nfrac{\alpha}{6}-\nfrac{(\sum_{j=1}^{6}\sigma_j)}{36}$
          \\
          $\{2\}$ &
          $\nfrac{(\sigma_4+\sigma_5+\sigma_6)}{6}+\nfrac{\alpha}{2}$ &
          $\alpha$ &
          $\nfrac{\alpha}{2}-\nfrac{(\sigma_4+\sigma_5+\sigma_6)}{6}$ &
          $\sfrac{1!(3-1-1)!}{3!}=\sfrac{1}{6}$ &
          $\nfrac{\alpha}{12}-\nfrac{(\sigma_4+\sigma_5+\sigma_6)}{36}$
          \\
          $\{3\}$ &
          $\nfrac{(\sigma_3+\sigma_6)}{4}+\nfrac{\alpha}{2}$ &
          $\alpha$ &
          $\nfrac{\alpha}{2}-\nfrac{(\sigma_3+\sigma_6)}{4}$ &
          $\sfrac{1!(3-1-1)!}{3!}=\sfrac{1}{6}$ &
          $\nfrac{\alpha}{12}-\nfrac{(\sigma_3+\sigma_6)}{24}$
          \\
          $\{2,3\}$ &
          $\nfrac{\sigma_6}{2}+\nfrac{\alpha}{2}$ &
          $\alpha$ &
          $\nfrac{\alpha}{2}-\nfrac{\sigma_6}{2}$ &
          $\sfrac{2!(3-2-1)!}{3!}=\sfrac{1}{3}$ &
          $\nfrac{\alpha}{6}-\nfrac{\sigma_6}{6}$
          \\
          \midrule[0.75pt]
          \multicolumn{5}{r}{Shapley value for feature 1 \hfill
            $\sv(1)~~=$} &
          $\nfrac{\alpha}{2}-\nfrac{(2\sigma_1+2\sigma_2+5\sigma_3+4\sigma_4+4\sigma_5+19\sigma_6)}{72}$ \\
          \bottomrule[1pt]
        \end{tabular}
      \end{center}
    \end{subfigure}
    %%\captionof{table}{Shapley value for feature 1} \label{tab:sv1}

    \medskip\medskip\medskip
    %
    %% Feature 2:
    \begin{subfigure}{1.0\textwidth}
      \begin{center}
        \renewcommand{\tabcolsep}{0.5em}
        \begin{tabular}{cccccc}
          \toprule[1pt]
          $\fml{S}$ &
          $\phi(\fml{S})$ &
          $\phi(\fml{S}\cup\{2\})$ &
          $\Delta(\fml{S})$ &
          $\varsigma(\fml{S})$ &
          $\varsigma(\fml{S})\times\Delta(\fml{S})$
          \\
          \midrule[0.875pt]
          $\emptyset$ &
          $\nfrac{(\sum_{j=1}^{6}\sigma_j)}{12}+\nfrac{\alpha}{2}$ &
          $\nfrac{(\sigma_4+\sigma_5+\sigma_6)}{6}+\nfrac{\alpha}{2}$ &
          $-\nfrac{(\sigma_1+\sigma_2+\sigma_3)}{12}+\nfrac{(\sigma_4+\sigma_5+\sigma_6)}{12}$ &
          $\sfrac{0!(3-0-1)!}{3!}=\sfrac{1}{3}$ &
          $-\nfrac{(\sigma_1+\sigma_2+\sigma_3)}{36}+\nfrac{(\sigma_4+\sigma_5+\sigma_6)}{36}$
          \\
          $\{1\}$ &
          $\alpha$ &
          $\alpha$ &
          $0$ &
          $\sfrac{1!(3-1-1)!}{3!}=\sfrac{1}{6}$ &
          $0$
          \\
          $\{3\}$ &
          $\nfrac{(\sigma_3+\sigma_6)}{4}+\nfrac{\alpha}{2}$ &
          $\nfrac{\sigma_6}{2}+\nfrac{\alpha}{2}$ &
          $-\nfrac{\sigma_3}{4}+\nfrac{\sigma_6}{4}$ &
          $\sfrac{1!(3-1-1)!}{3!}=\sfrac{1}{6}$ &
          $-\nfrac{\sigma_3}{24}+\nfrac{\sigma_6}{24}$
          \\
          $\{1,3\}$ &
          $\alpha$ &
          $\alpha$ &
          $0$ &
          $\sfrac{2!(3-2-1)!}{3!}=\sfrac{1}{3}$ &
          $0$
          \\
          \midrule[0.75pt]
          \multicolumn{5}{r}{Shapley value for feature 2 \hfill
            $\sv(2)~~=$} &
          $\nfrac{(-2\sigma_1-2\sigma_2-5\sigma_3+2\sigma_4+2\sigma_5+5\sigma_6)}{72}$ \\
          \bottomrule[1pt]
        \end{tabular}
      \end{center}
    \end{subfigure}
    %%\captionof{table}{Shapley value for feature 2} \label{tab:sv2}

    \medskip\medskip\medskip
    %
    %% Feature 3:
    \begin{subfigure}{1.0\textwidth}
      \begin{center}
        \renewcommand{\tabcolsep}{0.5em}
        \begin{tabular}{cccccc}
          \toprule[1pt]
          $\fml{S}$ &
          $\phi(\fml{S})$ &
          $\phi(\fml{S}\cup\{3\})$ &
          $\Delta(\fml{S})$ &
          $\varsigma(\fml{S})$ &
          $\varsigma(\fml{S})\times\Delta(\fml{S})$
          \\
          \midrule[0.875pt]
          $\emptyset$ &
          $\nfrac{(\sum_{j=1}^{6}\sigma_j)}{12}+\nfrac{\alpha}{2}$ &
          $\nfrac{(\sigma_3+\sigma_6)}{4}+\nfrac{\alpha}{2}$ &
          $-\nfrac{(\sigma_1+\sigma_2+\sigma_4+\sigma_5)}{12}+\nfrac{(\sigma_3+\sigma_6)}{6}$ &
          $\sfrac{0!(3-0-1)!}{3!}=\sfrac{1}{3}$ &
          $-\nfrac{(\sigma_1+\sigma_2+\sigma_4+\sigma_5)}{36}+\nfrac{(\sigma_3+\sigma_6)}{18}$
          \\
          $\{1\}$ &
          $\alpha$ &
          $\alpha$ &
          $0$ &
          $\sfrac{1!(3-1-1)!}{3!}=\sfrac{1}{6}$ &
          $0$
          \\
          $\{2\}$ &
          $\nfrac{(\sigma_4+\sigma_5+\sigma_6)}{6}+\nfrac{\alpha}{2}$ &
          $\nfrac{\sigma_6}{2}+\nfrac{\alpha}{2}$ &
          $-\nfrac{(\sigma_4+\sigma_5)}{6}+\nfrac{\sigma_6}{3}$ &
          $\sfrac{1!(3-1-1)!}{3!}=\sfrac{1}{6}$ &
          $-\nfrac{(\sigma_4+\sigma_5)}{36}+\nfrac{\sigma_6}{18}$
          \\
          $\{1,2\}$ &
          $\alpha$ &
          $\alpha$ &
          $0$ &
          $\sfrac{2!(3-2-1)!}{3!}=\sfrac{1}{3}$ &
          $0$
          \\
          \midrule[0.75pt]
          \multicolumn{5}{r}{Shapley value for feature 3 \hfill
            $\sv(3)~~=$} &
          $\nfrac{(-\sigma_1-\sigma_2+2\sigma_3-2\sigma_4-2\sigma_5+4\sigma_6)}{36}$ \\
          \bottomrule[1pt]
        \end{tabular}
      \end{center}
    \end{subfigure}
    
    \smallskip%%\medskip

    \caption{Computation of Shapley values for the example DT and
      instance $((1,1,2),\alpha)$. For each feature $i$, the sets to
      consider are all the sets that do not include the feature.
      The average values are obtained by summing up the values of the
      classifier in the rows consistent with $\fml{S}$ and dividing by
      the total number of rows.
      %For $\fml{S}=\{2,4\}$, only row 3 in the truth table takes value
      %1, and so the average becomes $\sfrac{1}{4}$.
    }
    \label{fig:cacm:svs}
  \end{mdframed}
\end{figure*}

Next, given the values obtained in~\cref{tab:cacm:phi}, we can compute
the Shapley value of each feature as shown in~\cref{fig:cacm:svs}.
The computed Shapley values are symbolic, and depend on the values
given to the predicted classes.

\paragraph{Shapley values for XAI can be misleading.}
%~\\
We now demonstrate that there is surprising flexibility in choosing
the influence of each feature. Concretely, we want to show that
Shapley values can yield misleading information, and that that is easy
to attain.
To achieve this goal, we are going to create an evidently disturbing
scenario. We are going to parameterize the classifier such that
feature 1 will be deemed to have \emph{no} importance on the
prediction, and feature 2 and 3 will be deemed to have \emph{some}
importance on the prediction. (Such choice of Shapley values can only
be considered misleading; as argued in~\cref{fig:cacm:disc}, for
either $\kappa_1$ or $\kappa_2$, predicting class 1 or predicting a
class other than 1 depends uniquely on feature 1.)
To obtain such a choice of Shapley values, we must have,
\begin{align}
  \sv(1) =~&\nfrac{\alpha}{2}-\nfrac{(2\sigma_1+2\sigma_2+5\sigma_3+4\sigma_4+4\sigma_5+19\sigma_6)}{72}
  = 0 \label{eq:cacm:cond01} \\
  \sv(2) =~&\nfrac{(-2\sigma_1-2\sigma_2-5\sigma_3+2\sigma_4+2\sigma_5+5\sigma_6)}{72}
  \not= 0 \label{eq:cacm:cond02} \\
  \sv(3) =~&\nfrac{(-\sigma_1-\sigma_2+2\sigma_3-2\sigma_4-2\sigma_5+4\sigma_6)}{36}
  \not=0 \label{eq:cacm:cond03} %%\\
\end{align}
Clearly, there are arbitrary many assignments to the values of
$\alpha$ and $\sigma_j,j=1,\ldots,6$ such that 
constraints~\eqref{eq:cacm:cond01},~\eqref{eq:cacm:cond02},
and~\eqref{eq:cacm:cond03} are satisfied.
Indeed, we can express $\alpha$ in terms of the other parameters
$\sigma_1\ldots,\sigma_6$, as follows:
\begin{equation} \label{eq:cacm:def:alpha}
  \alpha=\nfrac{(2\sigma_1+2\sigma_2+5\sigma_3+4\sigma_4+4\sigma_5+19\sigma_6)}{36}
\end{equation}
In addition, we can identify values for $\sigma_j$, $j=1,\ldots,6$,
such that the two remaining conditions \eqref{eq:cacm:cond02}
and~\eqref{eq:cacm:cond03} are satisfied.
Moreover, we can analyze the values shown in~\cref{fig:cacm}.
Concretely, we pick
$\alpha=1\land\sigma_1=\sigma_3=\sigma_4=\sigma_6=0$ for both
$\kappa_1$ and $\kappa_2$.
Also, for $\kappa_1$, we choose $\sigma_2=4\land\sigma_5=7$.
Finally, for $\kappa_2$, we choose $\sigma_2=2\land\sigma_5=3$.
The resulting sets of Shapley values are shown
in~\cref{tab:cacm:svs:cmp} for the two classifiers of~\cref{fig:cacm}.

\begin{table}[t]
  \begin{tabular}{ccccc} \toprule
    DT & Classifier & $\sv(1)$ & $\sv(2)$ & $\sv(3)$
    \\ \toprule
    \cref{fig:00:dt}  & $\kappa_1$ & 0 & 0.167 & -0.5 \\[1pt]
    \cref{fig:00:dt2} & $\kappa_2$ & 0.306 & 0.028 & -0.194 \\
    \bottomrule
  \end{tabular}
  \caption{Comparison of Shapley values} \label{tab:cacm:svs:cmp}
\end{table}

In the case of classifier $\kappa_1$ (or by selecting $\alpha=1$,
$\sigma_1=\sigma_3=\sigma_4=\sigma_6=0$, $\sigma_2=4$ and
$\sigma_5=7$), and given the discussion in~\cref{fig:cacm:disc}, we
conclude that feature 1, which uniquely determines the prediction for
class 1, and which must be modified to change the prediction for class
1, is assigned a Shapley value of 0, which is supposed to
signify~\cite{kononenko-jmlr10} that it has \emph{no} influence on the
prediction.
Similarly, features 2 and 3, which were shown in~\cref{fig:cacm:disc}
to bear \emph{no} influence in determining or changing the prediction
for class 1, are assigned non-zero Shapley values, which are supposed
to signify~\cite{kononenko-jmlr10} \emph{some} influence on the
prediction.
Thus, as the example demonstrates, the computed Shapley values
rank \emph{incorrectly} the features in terms of their influence in
determining or changing the prediction.

\if0
\condaddval{
  \paragraph{Validation.}
  %~\\
  To validate the results shown above, we get:
  %
  \[\begin{array}{l}
  \sum_{j=1}^{6} \sigma_j = 2 + 0 + 0 + 4 + 4 + 0 = 10 \\
  \kappa(\mbf{v}) = \kappa(1,1,2) = \alpha = 1 \\
  \phi(\emptyset) = \nfrac{10}{12} + \nfrac{1}{2} = \nfrac{4}{3}\\
  \sum_i \sv(i) = 0 + \nfrac{1}{6} - \nfrac{1}{2} = -\nfrac{1}{3}\\
  \end{array}
  \]
  Thus, we should have,
  \[
  \begin{array}{rcc}
    -\nfrac{1}{3} + \nfrac{4}{3} = 1 & \Leftrightarrow & \\
    \nfrac{3}{3} = 1 & ~~ & \tn{\yesmark}\\
  \end{array}
  \]
}
\fi

\paragraph{Generalization.}
%~\\
Since the example is parameterized in $\alpha$, $\sigma_j$,
$j=1,\ldots,6$, we can define arbitrary many classifiers for which the
limitations of Shapley values for XAI are revealed.
For example, it is plain to conclude that by setting $\alpha=\psi$ and 
$\sigma_1=\sigma_3=\sigma_4=\sigma_6=0,\sigma_2=4\psi,\sigma_5=7\psi$, 
e.g.\ with $\psi\in\mbb{Z}^{+}$, we obtain a classifier where feature
1 is relevant (and necessary), features 2 and 3 are irrelevant, and
such that $\sv(1)=0$, $\sv(2)\not=0$ and $\sv(3)\not=0$.

Furthermore, other combinations of values can be selected as long as
conditions \eqref{eq:cacm:cond00}, \eqref{eq:cacm:cond01},
\eqref{eq:cacm:cond02} and \eqref{eq:cacm:cond03} hold true. Each such
combination of values captures an arbitrary number of classifiers, as
illustrated above.

\paragraph{Consequences for the hypothetical scenario.}
%~\\
Although from a formal perspective, the proposed example suffices to
demonstrate the inadequacy of Shapley values for XAI, it can be more
intuitive to discuss the hypothetical scenario outlined at the outset
of this section.
As argued earlier, the instance $((1,1,2),1)$ denotes a honors student
from an urban household that is studying sciences. Formal
explanations, either abductive or contrastive, consist of feature 1
(honors student).
In contrast, in terms of Shapley values for XAI, feature 1 would be
assigned no importance, albeit it determines the prediction, and
features 2  (type of household) and 3  (field of study) would be
assigned some importance, albeit none of these features has any
influence on the prediction.
Given the information provided by Shapley values, a decision maker
would have to deem features 2 and 3 as influencing the prediction, and
feature 1 and not influencing the prediction. However, as clearly
shown by the DT and TR in~\cref{fig:02:dt,fig:02:tt}, feature 1 is
actually the only feature bearing influence in the prediction.
Clearly, the inevitable conclusion is that the information provided by
Shapley values for XAI is misleading.

\paragraph{Discussion.}
%~\\
As observed above, for the family of DTs discussed in this section,
namely those respecting \eqref{eq:cacm:cond00},
\eqref{eq:cacm:cond01}, \eqref{eq:cacm:cond02} and
\eqref{eq:cacm:cond03}, and in the concrete case of point $(1,1,2)$,
the computed Shapley values will provide misleading information. 
Concretely, feature 1, which must be used to fix or to change the
prediction, is assigned a Shapley value of 0, thus signifying no
influence on the prediction~\cite{kononenko-jmlr10}. Moreover,
features 2 and 3, which are never used to fix or change the
prediction, are assigned a non-zero Shapley value, thus signifying
some influence on the prediction.
Therefore, for this family of classifiers, the computed Shapley values
produce information that bears \emph{no} correlation with actual
feature influence in either fixing or changing the prediction.
Similar results have been reported in a number of recent
reports~\cite{hms-corr23a,msh-corr23,hms-corr23b}.

\paragraph{Pinpointing the problem.}
%~\\
How come Shapley values can provide such misleading information?
A more detailed analysis of the example reveals that several sets of
fixed features, which play no role in abductively explaining the
prediction, serve to bias the computed Shapley values in a way such
that these bear no correlation with the relative importance of
features in for the given prediction. As an example, this is the case
with the sets $\{1,2\}$, $\{1,3\}$ and $\{1,2,3\}$, i.e.\ any proper
superset of the computed AXp.
In contrast, if we were to look at the sets of AXp's or CXp's we would
only account for the features that play some role either in
determining or in changing the prediction.

The results reported in this and earlier reports have already
motivated the proposal of alternatives to the use of Shapley
values~\cite{hms-corr23a,ignatiev-corr23,izza-corr23}, which aim at
computing relative orders of feature importance that respect the
actual logic operation of the ML model.

  \end{appendices}
}

\end{document}